# Grounding the Lexical Semantics of Verbs in Visual Perception using Force Dynamics and Event Logic

**Jeffrey Mark Siskind** QOBI@RESEARCH.NJ.NEC.COM
*NEC Research Institute, Inc.*
*4 Independence Way*
*Princeton, NJ 08540 USA*

## Abstract

This paper presents an implemented system for recognizing the occurrence of events described by simple spatial-motion verbs in short image sequences. The semantics of these verbs is specified with event-logic expressions that describe changes in the state of force-dynamic relations between the participants of the event. An efficient finite representation is introduced for the infinite sets of intervals that occur when describing liquid and semi-liquid events. Additionally, an efficient procedure using this representation is presented for inferring occurrences of compound events, described with event-logic expressions, from occurrences of primitive events. Using force dynamics and event logic to specify the lexical semantics of events allows the system to be more robust than prior systems based on motion profile.

## 1. Introduction

If one were to look at the image sequence in Figure 1(a), one could describe the event depicted in that sequence by saying *Someone picked the red block up off the green block*. Similarly, if one were to look at the image sequence in Figure 1(b), one could describe the event depicted in that sequence by saying *Someone put the red block down on the green block*. One way that one recognizes that the former is a *pick up*[1] event is that one notices a state change in the force-dynamic relations between the participant objects. Prior to Frame 13, the red block is supported by the green block by a substantiality constraint, the fact that solid objects cannot interpenetrate (Spelke, 1983; Baillargeon, Spelke, & Wasserman, 1985; Baillargeon, 1986, 1987; Spelke, 1987, 1988). From Frame 13 onward, it is supported by being attached to the hand. Similarly, one way that one recognizes that the latter is a *put down* event is that one notices the reverse state change in Frame 14. This paper describes an implemented computer system, called LEONARD, that can produce similar event descriptions from such image sequences. A novel aspect of this system is that it produces event descriptions by recognizing state changes in force-dynamic relations between participant objects. *Force dynamics* is a term introduced by Talmy (1988) to describe a variety of causal relations between participants in an event, such as allowing, preventing, and forcing. In this paper, I use force dynamics in a slightly different sense, namely, to describe the *support*, *contact*, and *attachment* relations between participant objects.

A number of systems have been reported that can produce event descriptions from video input. Examples of such systems include the work reported in Yamoto, Ohya, and Ishii (1992), Starner (1995), Siskind and Morris (1996), Brand (1996), and Bobick and Ivanov (1998). LEONARD differs

1. Throughout this paper, I treat verb-particle constructions, like *pick up*, as atomic verbs, despite the fact that the verb and its associated particle may be discontinuous. Methods for deriving the semantics of a verb-particle construction compositionally from its (discontinuous) constituents are beyond the scope of this paper.

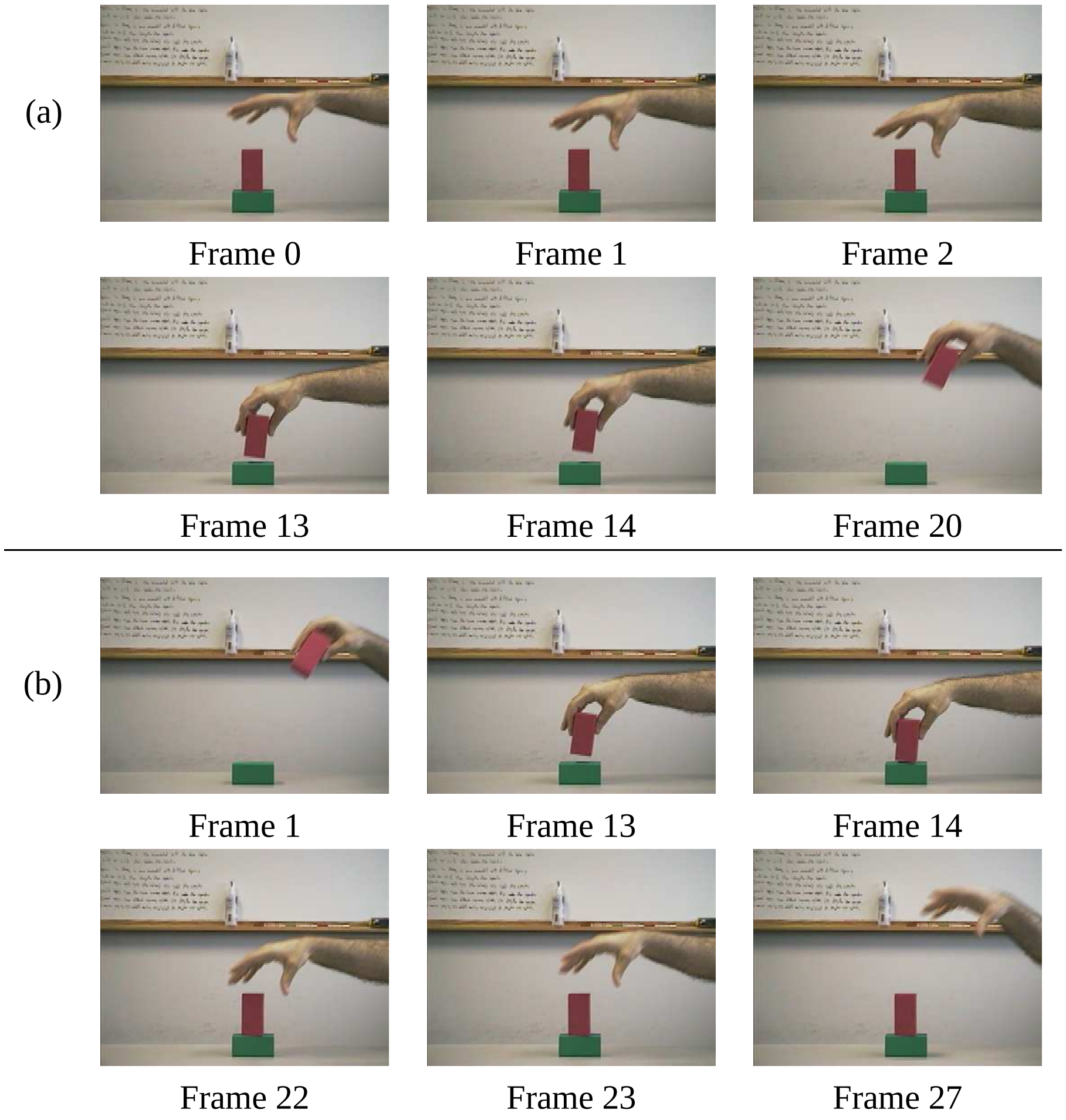

Figure 1: Image sequences depicting (a) a *pick up* event and (b) a *put down* event.

from these prior systems in two crucial ways. First, the prior systems classify events based on the motion profile of the participant objects. For example, Siskind and Morris (1996) characterize a *pick up* event as a sequence of two subevents: the agent moving towards the patient while the patient is at rest above the source, followed by the agent moving with the patient away from the source while the source remains at rest. Similarly, a *put down* event is characterized as the agent moving with the patient towards the destination while the destination is at rest, followed by the agent moving away from the patient while the patient is at rest above the destination. In contrast, LEONARD characterizes events as changes in the force-dynamic relations between the participant objects. For example, a *pick up* event is characterized as a change from a state where the patient is supported by a substantiality constraint with the source to a state where the patient is supported by being attached to the agent. Similarly, a *put down* event is characterized as the reverse state change. Irrespective of whether motion profile or force dynamics is used to recognize events, event recognition is a process of classifying time-series data. In the case of motion profile, this time-series data takes the form of relative-and-absolute positions, velocities, and accelerations of the participant objects as a function of time. In the case of force dynamics, this time-series data takes the form of the truth values of force-dynamic relations between the participant objects as a function of time. This leads to the second difference between LEONARD and prior systems. The prior systems use stochastic reasoning, in the form of hidden Markov models, to classify the time-series data into event types. In contrast, LEONARD uses logical reasoning, in the form of *event logic*, to do this classification.

Using force dynamics and event logic (henceforth the 'new approach') to recognize events offers several advantages over using motion profile and hidden Markov models (henceforth the 'prior approach'). First, the new approach will correctly recognize an event despite a wider variance in motion profile than the prior approach. For example, when recognizing, say, a *pick up* event, the prior approach is sensitive to aspects of event execution, like the approach angle and velocity of the hand, that are irrelevant to whether or not the event is actually a *pick up*. The new approach is not sensitive to such aspects of event execution. Second, the new approach will correctly recognize an event despite the presence of unrelated objects in the field of view. The prior approach computes the relative-and-absolute positions and motions of all objects and pairs of objects in the field of view. It then selects the subset of objects whose positions and motions best matched some model. This could produce incorrect descriptions when some unintended subset matched some unintended model better than the intended subset matched the intended model. The new approach does not exhibit such deficiencies. Extraneous objects typically do not exhibit the precise sequence of state changes in force-dynamic relations needed to trigger the event-classification process and thus will not generate spurious claims of event occurrences. Third, the new approach performs temporal and spatial segmentation of events. The prior approach matches an entire image sequence against an event model. It fails if that image sequence depicts multiple event executions, either in sequence or in parallel. In contrast, the new approach can segment a complex image sequence into a collection of sequential and/or overlapping events. In particular, it can handle *hierarchal* events, such as *move*, that consist of a *pick up* event followed by a *put down* event. It can recognize that all three events, and precisely those three events, occur in an appropriate image sequence whereas the prior approach would try to find the single best match. Finally, the new approach robustly detects the non-occurrence of events as well as the occurrence of events. The prior approach always selects the best match and reports some event occurrence for every image sequence. Thresholding the match cost does not work because an approach based on motion profile can be fooled into triggering recognition of an event occurrence by an event whose motion profile is similar to one or more target event classes even though that event is not actually in any of those target event classes. Consider, for example, the two image sequences in Figure 2. Suppose that an event-recognition system contained two target event classes, namely *pick up* and *put down*. Neither of the image sequences depict *pick up* or *put down* events. Nonetheless, the prior approach might mistakingly classify Figure 2(a) as a *pick up* event because the second half of this image sequence matches the second half of the motion profile of a *pick up* event. Alternatively, it might mistakingly classify this image sequence as a *put down* event because the first half of this image sequence matches the first half of the motion profile of a *put down* event. Similarly, the prior approach might mistakingly classify Figure 2(b) as a *pick up* event because the first half of this image sequence matches the first half of the motion profile of a *pick up* event. Alternatively, it might mistakingly classify this image sequence as a *put down* event because the second half of this image sequence matches the second half of the motion profile of a *put down* event. In contrast, the new approach correctly recognizes that neither of these image sequences exhibit the necessary state changes in force-dynamic relations to qualify as either *pick up* or *put down* events. All four of these advantages will be discussed in greater detail in Section 5.

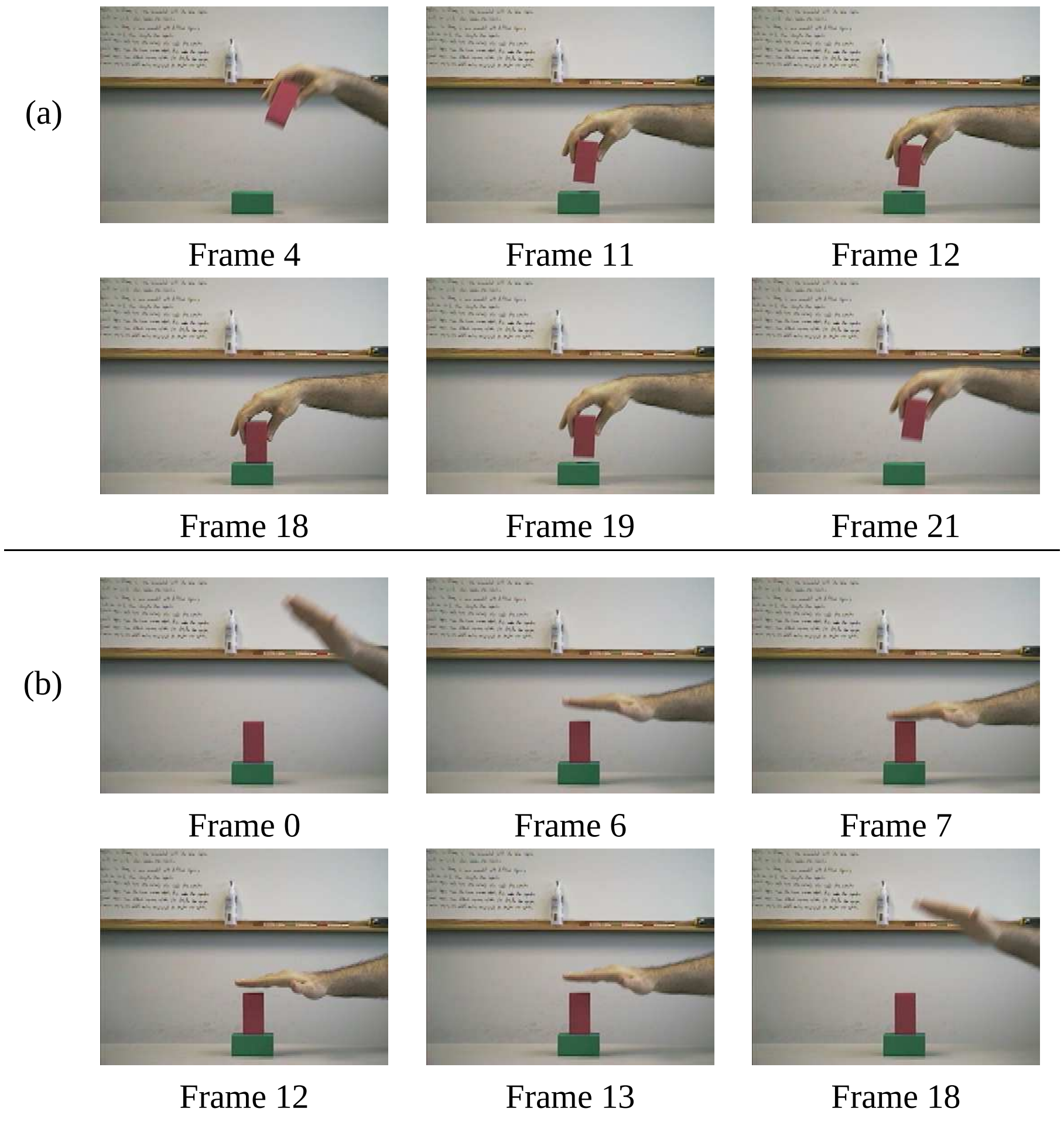

Figure 2: Image sequences depicting non-events.

The techniques described in this paper have been implemented in a system called LEONARD. LEONARD is a comprehensive system that takes image sequences as input and produces event descriptions as output. The overall architecture of LEONARD is shown in Figure 6. The input to LEONARD consists of a sequence of images taken by a Canon VC-C3 camera and Matrox Meteor frame grabber at 320×240 resolution at 30fps. This image sequence is first processed by a *segmentation-and-tracking* component. A real-time colour- and motion-based segmentation algorithm places a convex polygon around each coloured and moving object in each frame. A tracking algorithm then forms a correspondence between the polygons in each frame and those in temporally adjacent frames. The output of the segmentation-and-tracking component consists of a sequence of *scenes*, each scene being a set of polygons. Each polygon is represented as a sequence of image coordinates corresponding to a clockwise traversal of the polygon's vertices. The tracker guarantees that each scene contains the same number of polygons and that they are ordered so that the $i^{\text{th}}$ polygon in each scene corresponds to the same object. Figure 3 shows the output of the segmentation-and-tracking component on the image sequences from Figure 1. The polygons have been overlayed on the input images for ease of comprehension.

This scene sequence is passed to a *model-reconstruction* component. This component produces a force-dynamic model of each scene. This model specifies three types of information: which objects are *grounded*, i.e. are supported by an unseen mechanism that is not associated with any visible object, which objects are *attached* to other objects by rigid or revolute joints, and the qualitative depth of each object, i.e. a qualitative representation of the relative distance of different objects in the field of view from the observer, in the form of a *same layer* relation specifying which objects are at the same qualitative depth. Figure 4 shows the output of the model-reconstruction component on the scene sequences from Figure 3. The models are depicted graphically, overlayed on the input images, for ease of comprehension. The details of this depiction scheme will be described momentarily. For now, it suffices to point out that Figure 4(a) shows the red block on the same layer as the green block up through Frame 1 and attached to the hand from Frame 14 onward. Figure 4(b) shows the reverse sequence of relations, with the red block attached to the hand up through Frame 13 and on the same layer as the green block from Frame 23 onward.

This model sequence is passed to an *event-classification* component. This component first determines the intervals over which certain primitive event types are true. These primitive event types include SUPPORTED$(x)$, SUPPORTS$(x, y)$, CONTACTS$(x, y)$, and ATTACHED$(x, y)$. This component then uses an inference procedure to determine the intervals over which certain compound event types are true. These compound event types include PICKUP$(x, y, z)$, PUTDOWN$(x, y, z)$, STACK$(w, x, y, z)$, UNSTACK$(w, x, y, z)$, MOVE$(w, x, y, z)$, ASSEMBLE$(w, x, y, z)$, and DISASSEMBLE$(w, x, y, z)$ and are specified as expressions in event logic over the primitive event types. The output of the event-classification component consists of an indication of which compound event types occurred in the input movie as well as the subsequence(s) of frames during which those event types occurred. Figure 5 shows the output of the event-classification component on the model sequences from Figure 4. The subsequences of frames during which the events occurred are depicted as *spanning intervals*. Spanning intervals will be described in Section 4.1.

LEONARD is too complex to describe completely in one paper. This paper provides a detailed description of the event-classification component and, in particular, the event-logic inference procedure. The segmentation and tracking algorithms are extensions of the algorithms presented in Siskind and Morris (1996) and Siskind (1999), modified to place convex polygons around the participant objects instead of ellipses. The model-reconstruction techniques are extensions of those

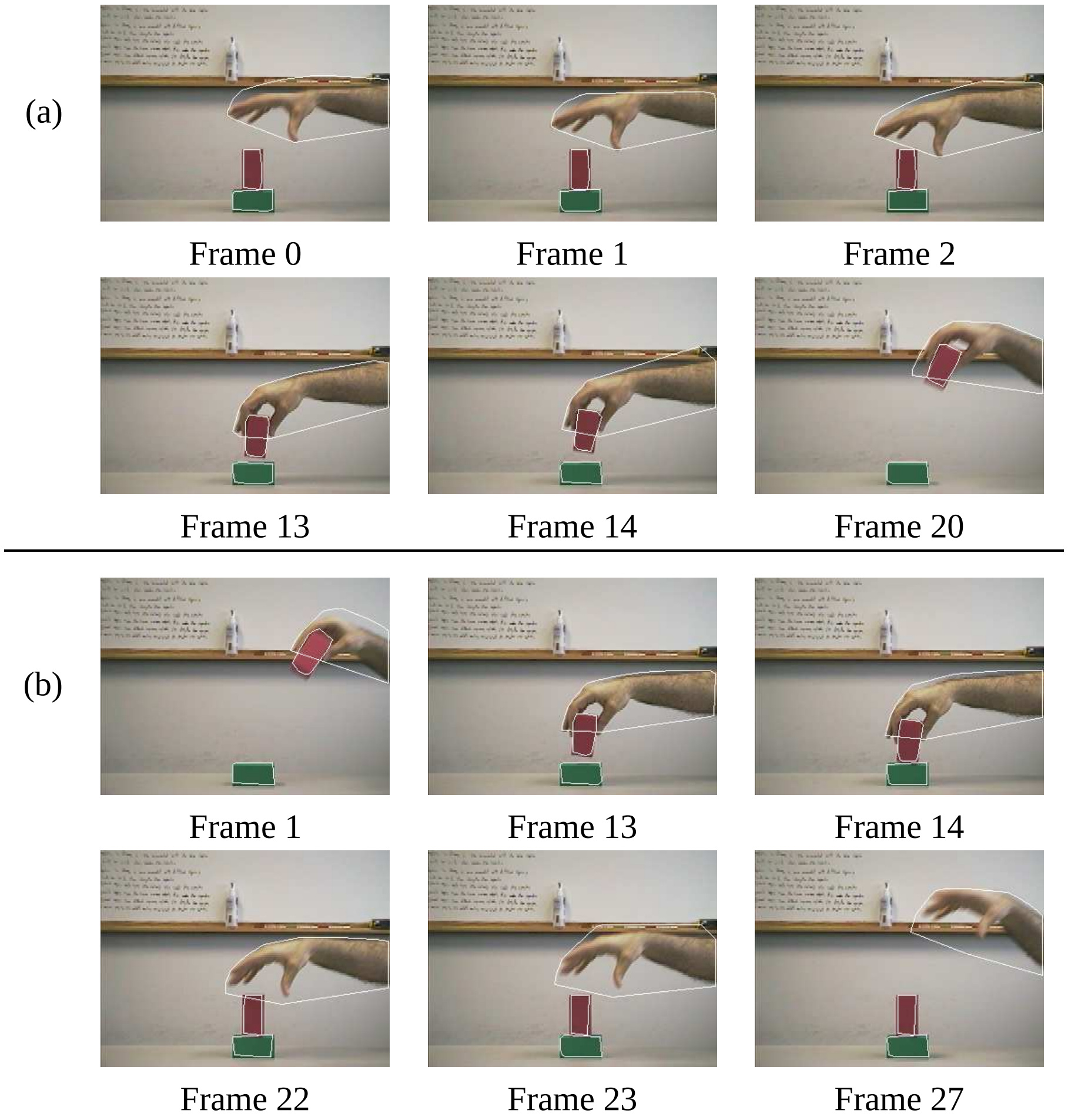

Figure 3: The output of the segmentation-and-tracking component applied to the image sequences from Figure 1. (a) depicts a *pick up* event. (b) depicts a *put down* event. The polygons have been overlayed on the input images for ease of comprehension.

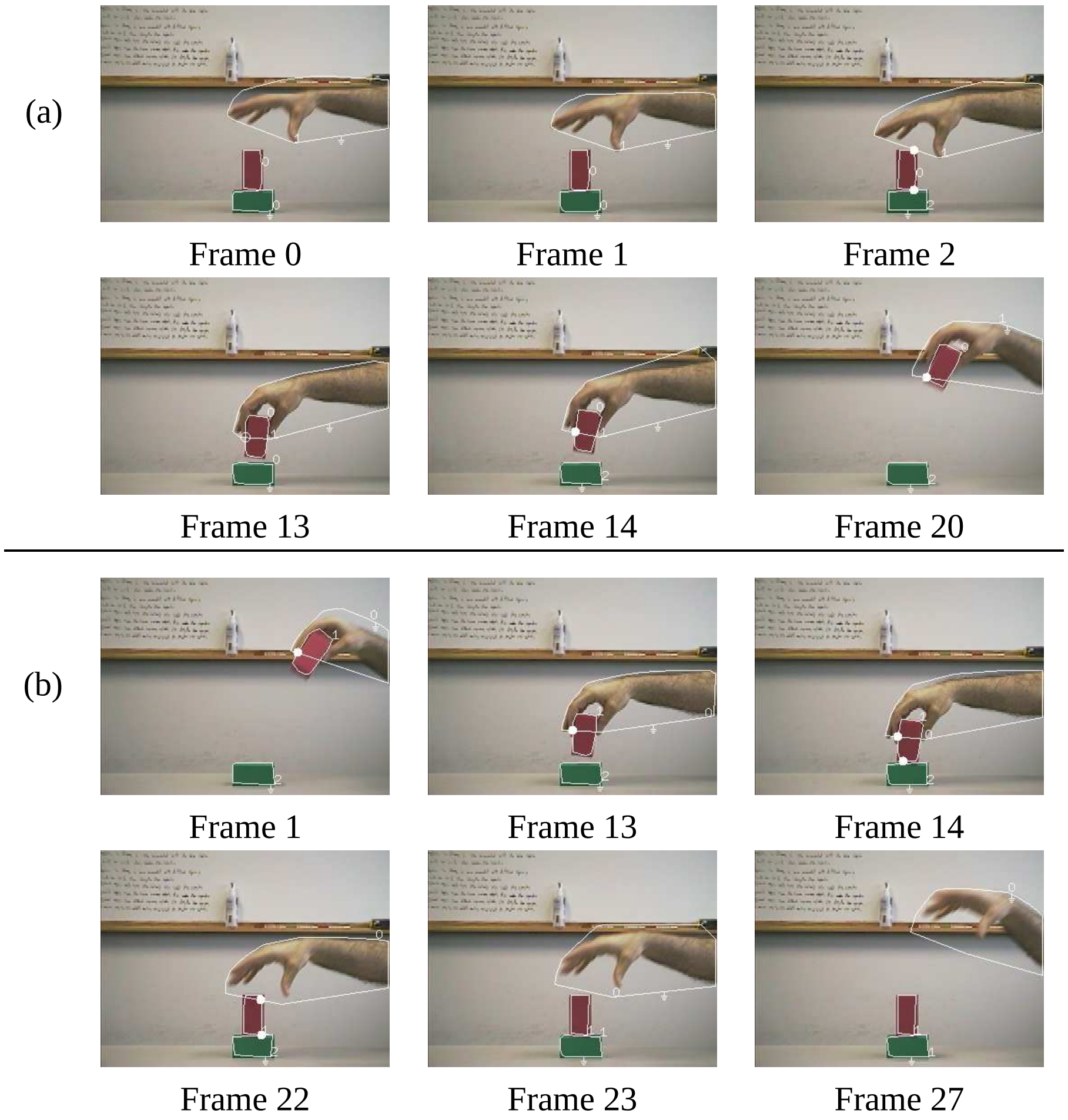

Figure 4: The output of the model-reconstruction component applied to the scene sequences from Figure 3. (a) depicts a *pick up* event. (b) depicts a *put down* event. The models have been overlayed on the input images for ease of comprehension. In (a), the red block is on the same layer as the green block up through Frame 1 and is attached to the hand from Frame 14 onward. In (b), the reverse sequence of relations holds, with the red block attached to the hand up through Frame 13 and on the same layer as the green block from Frame 23 onward.

```
(PICK-UP MOVING RED GREEN)@{[[0,1],[14,22])}

(SUPPORTED? RED)@{[[0:22])}
(SUPPORTED? MOVING)@{[[1:13]), [[24:26])}
(SUPPORTS? RED MOVING)@{[[1:13]), [[24:26])}
(SUPPORTS? MOVING RED)@{[[13:22])}
(SUPPORTS? GREEN RED)@{[[0:14])}
(SUPPORTS? GREEN MOVING)@{[[1:13])}
(CONTACTS? RED GREEN)@{[[0:2]), [[6:14])}
(ATTACHED? RED MOVING)@{[[1:26])}
(ATTACHED? RED GREEN)@{[[1:6])}
```

(a)

```
(PUT-DOWN MOVING RED GREEN)@{[[0,14],[23,32])}

(SUPPORTED? MOVING)@{[[14:23])}
(SUPPORTED? RED)@{[[0:32])}
(SUPPORTS? MOVING RED)@{[[0:14])}
(SUPPORTS? RED MOVING)@{[[14:23])}
(SUPPORTS? GREEN MOVING)@{[[14:23])}
(SUPPORTS? GREEN RED)@{[[14:32])}
(CONTACTS? RED GREEN)@{[[22:32])}
(ATTACHED? MOVING RED)@{[[0:23])}
(ATTACHED? RED GREEN)@{[[14:22])}
```

(b)

Figure 5: The output of the event-classification component applied to the model sequences from Figure 4. Note that the *pick up* event is correctly recognized in (a) and the *put down* event is correctly recognized in (b).

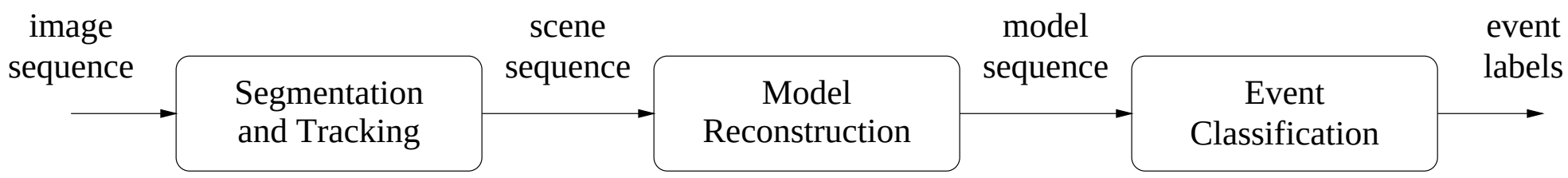

Figure 6: The overall architecture of LEONARD.

presented in Siskind (1997, 2000). The model-reconstruction techniques will be described briefly below to allow the reader to understand the event-classification techniques without reference to those papers.

## 2. Model Reconstruction

Certain properties of objects are visible. For example, position, orientation, shape, size, colour, texture, and so forth. Furthermore, relational variants of these properties are also visible, as well as changes in such properties and relations over time. In contrast, force-dynamic properties and relations are not visible. One cannot *see* the fact that the door knob is attached to, and supported by, the door. One must *infer* that fact using physical knowledge of the world. Such knowledge includes the fact that unsupported objects fall and attachment is one way of offering support. Using physical knowledge to infer force-dynamic properties and relations was first discussed by Siskind (1991, 1992, 1993). This later became known as the *perceiver framework* advanced by Jepson and Richards (1993). The perceiver framework states that perception involves four levels. First, one must specify the *observables*, what properties and relations can be discerned by direct observation. Second, one must specify an *ontology*, what properties and relations must be inferred from the observables. Descriptions of the observables in terms of such properties and relations are called *interpretations*. There may be multiple interpretations of a given observation. Third, one must specify a *theory*, a way of differentiating *consistent* interpretations from *inconsistent* ones. The consistent interpretations are the *models* of the observation. There may be multiple models of a given observation. Finally, one must specify a *preference relation*, a way of ordering the models. The most-preferred models of the observations are the *percepts*. One can instantiate the perceiver framework for different observables, ontologies, theories, and preference relations. Siskind (1991, 1992, 1993, 1994, 1995, 1997) instantiated this framework for a kinematic theory applied to simulated video. Mann, Jepson, and Siskind (1996, 1997) and Mann and Jepson (1998) instantiated this framework for a dynamics theory applied to real video. Siskind (2000) instantiated this framework for a kinematic theory applied to real video. This paper uses this later approach.

The input to the model-reconstruction process consists of a sequence of *scenes*, each scene being a set of convex polygons. Each polygon is represented as a sequence of points corresponding to a clockwise traversal of the polygon's vertices. The tracker guarantees that each scene contains the same number of polygons and that they are ordered so that the $i^{\text{th}}$ polygon in each scene corresponds to the same object. The output of the model-reconstruction process consists of a sequence of *interpretations*, one interpretation per scene. The interpretations are formulated out of the following primitive properties of, and relations between, the objects in each scene.

GROUNDED$(p)$ Polygon $p$ is *grounded*. It is constrained to occupy a fixed position and orientation by an unseen mechanism that is not associated with any visible object and thus cannot move either translationally or rotationally.

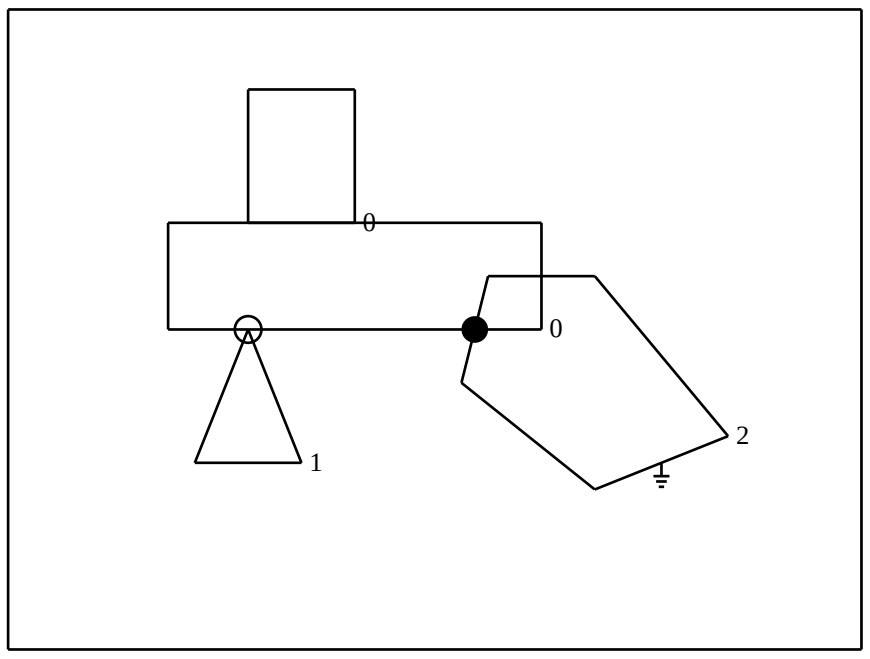

Figure 7: The graphical method for depicting interpretations that is used in this paper. The symbol '⏚' indicates that a polygon is grounded. A solid circle indicates a rigid joint. A hollow circle indicates a revolute joint. Two polygons with the same layer index are on the same layer.

RIGID$(p, q, r)$ Polygons $p$ and $q$ are *attached* by a *rigid joint* at point $r$. Both the relative position and orientation of $p$ and $q$ are constrained.

REVOLUTE$(p, q, r)$ Polygons $p$ and $q$ are *attached* by a *revolute joint* at point $r$. The relative position of $p$ and $q$ is constrained but the relative orientation is not.

SAMELAYER$(p, q)$ Polygons $p$ and $q$ are on the *same layer*. Layers are a qualitative representation of depth, or distance from the observer. This representation is impoverished. There is no notion of 'in-front-of' or 'behind' and there is no notion of adjacency in depth. The only representable notion is whether two objects are on the same or different layers. The same-layer relation is constrained to be an equivalence relation, i.e. it must be reflexive, symmetric, and transitive. Furthermore, two objects on the same layer must obey the *substantiality constraint*, the constraint that they not interpenetrate (Spelke, 1983; Baillargeon et al., 1985; Baillargeon, 1986, 1987; Spelke, 1987, 1988).

An interpretation $I$ is a 4-tuple: $\langle$GROUNDED, RIGID, REVOLUTE, SAMELAYER$\rangle$. Throughout this paper, interpretations will be depicted graphically, overlayed on scene images, for ease of comprehension. Figure 7 gives a sample interpretation depicted graphically. The symbol '⏚' attached to a polygon indicates that it is grounded. A solid circle indicates that two polygons are rigidly attached at the center of the circle. A hollow circle indicates that two polygons are attached by a revolute joint at the center of the circle. The same-layer relation is indicated by giving a *layer index*, a small nonnegative integer, to each polygon. Polygons with the same layer index are on the same layer, while those with different layer indices are on different layers.

Model reconstruction can be viewed as a generate-and-test process. Initially, all possible interpretations are generated for each scene. Then, *inadmissible* and *unstable* interpretations are filtered out. Admissibility and stability can be collectively viewed as a consistency requirement. The stable admissible interpretations are thus *models* of a scene. The nature of the theory guarantees that there will always be at least one model for each scene, namely the model where all objects are grounded. There may, however, be multiple models for a given scene. Therefore, a *preference re-* *lation* is then applied through a sequence of circumscription processes (McCarthy, 1980) to select the minimal, or preferred, models for each scene. While there will always be at least one minimal model for each scene, there may be several, since the preference relation may not induce a total order. If there are multiple minimal models for a given scene, one is chosen arbitrarily as the most-preferred model for that scene. The precise details of the admissibility criteria, the stability checking algorithm, the preference relations, and the circumscription process are beyond the scope of this paper. They are discussed in Siskind (2000). What is important, for the purpose of this paper, is that, given a scene sequence, model reconstruction produces a sequence of interpretations, one for each scene, and that these interpretations are $4$-tuples containing the predicates GROUNDED, RIGID, REVOLUTE, and SAMELAYER. Figure 4 shows sample interpretation sequences produced by the model-reconstruction component on the scene sequences from Figure 3.

## 3. Event Logic

Model reconstruction determines the truth values of the force-dynamic relations on a frame-by-frame basis in the input movie. Intervals of constant truth value for a given force-dynamic relation are taken to be primitive event occurrences. LEONARD uses *event logic* to infer compound event occurrences from primitive event occurrences. For example, for the image sequence in Figure 1(a), model reconstruction determines that the green block supports the red block from Frame 0 to Frame 13 and that the hand is attached to the red block from Frame 13 to Frame 20. This will be denoted as SUPPORTS(**green-block**, **red-block**)$@[0,13)$ and ATTACHED(**hand**, **red-block**)$@[13,20)$, i.e. that the primitive event types SUPPORTS(**green-block**, **red-block**) and ATTACHED(**hand**, **red-block**) occurred during the intervals $[0,13)$ and $[13,20)$ respectively. The compound event type PICKUP(**hand**, **red-block**, **green-block**) might be defined as

$$\textsc{Supports}(\textbf{green-block},\textbf{red-block});\textsc{Attached}(\textbf{hand},\textbf{red-block})$$

i.e. SUPPORTS(**green-block**, **red-block**) followed by ATTACHED(**hand**, **red-block**). (In the above, I use ';' informally as a sequence operator. The precise definition of ';' will be given momentarily.) The task of the event-logic inference procedure is to infer PICKUP(**hand**, **red-block**, **green-block**)$@[0,20)$, i.e. that the compound event type PICKUP(**hand**, **red-block**, **green-block**) occurred during the interval $[0,20)$.

Event logic provides a calculus for forming compound event types as expressions over primitive event types. The syntax and semantics of event logic will be described momentarily. Event-logic expressions denote event types, not event occurrences. As such, they do not have truth values. Rather, they are predicates that describe the truth conditions that must hold of an interval for an event to occur. In contrast, an event-occurrence formula does have a truth value. If $\Phi$ is an event-logic expression that denotes a primitive or compound event type, and $\mathbf{i}$ is an interval, then $\Phi@\mathbf{i}$ is an atomic event-occurrence formula that is true if and only if the truth conditions for the event type $\Phi$ hold of the interval $\mathbf{i}$.

$\Phi@\mathbf{i}$ denotes *coincidental* occurrence, the fact that an occurrence of $\Phi$ started at the beginning of $\mathbf{i}$ and finished at the end of $\mathbf{i}$. $\Phi@\mathbf{i}$ would not hold if an occurrence of $\Phi$ did not precisely coin-cide with **i**, but instead overlapped,[2] partially or totally, with **i**. Event types have internal temporal structure that render this distinction important. In the case of primitive event types, that structure is simple. Each primitive event type is derived from a predicate. If $\phi$ is a predicate, then $\overline{\phi}$ denotes the primitive event type derived from $\phi$. A primitive event type $\overline{\phi}$ holds of an interval if the corresponding predicate $\phi$ holds of every instant in that interval.[3] This means that $\neg(\overline{\phi}@\mathbf{i})$ and $\overline{\neg\phi}@\mathbf{i}$ might have different truth values. For example, if $\phi$ is true of every instant in $[0,2)$ and false of every other instant, then $\neg(\overline{\phi}@[1,3))$ is true while $\overline{\neg\phi}@[1,3)$ is false. Event logic takes coincidental occurrence to be a primitive notion. As will be demonstrated below, overlapping occurrence is a derived notion that can be expressed in terms of coincidental occurrence using compound event-logic expressions.

Two auxiliary notions are needed to define the syntax and semantics of event logic. First, there are thirteen possible relations between two intervals. Following Allen (1983), I denote these relations as $=$, $<$, $>$, $\mathsf{m}$, $\mathsf{mi}$, $\mathsf{o}$, $\mathsf{oi}$, $\mathsf{s}$, $\mathsf{si}$, $\mathsf{f}$, $\mathsf{fi}$, $\mathsf{d}$, and $\mathsf{di}$ and refer to them collectively as *Allen relations* throughout the paper. The names $\mathsf{m}$, $\mathsf{o}$, $\mathsf{s}$, $\mathsf{f}$, and $\mathsf{d}$ are mnemonics for *meet*, *overlap*, *start*, *finish*, and *during* respectively. The inverse relations, such as $\mathsf{mi}$, whose names end in $\mathsf{i}$ are the same as the corresponding relations, such as $\mathsf{m}$, whose names do not end in $\mathsf{i}$ except that the arguments are reversed. Figure 8 depicts all thirteen Allen relations graphically. Second, I define the *span* of two intervals **i** and **j**, denoted $\textsc{Span}(\mathbf{i},\mathbf{j})$, as the smallest super-interval of both **i** and **j**.

The syntax of event logic is defined as follows. We are given finite disjoint sets of constant symbols along with a finite set of primitive event-type symbols, each of a specified arity. Constant symbols, such as **red-block** and **hand**, denote objects in the input movie while primitive event-type symbols, such as SUPPORTS, denote parameterized primitive event types. An atomic event-logic expression is a primitive event-type symbol of arity $n$ applied to a sequence of $n$ constants. For example, $\textsc{Supports}(\textbf{hand},\textbf{red-block})$. An event-logic expression is either an atomic event-logic expression or one of the compound event-logic expressions $\neg\Phi$, $\Phi\vee\Psi$, $\Phi\wedge_R\Psi$, or $\Diamond_R\Phi$, where $\Phi$ and $\Psi$ are event-logic expressions and $R\subseteq\{=,<,>,\mathsf{m},\mathsf{mi},\mathsf{o},\mathsf{oi},\mathsf{s},\mathsf{si},\mathsf{f},\mathsf{fi},\mathsf{d},\mathsf{di}\}$.

Informally, the semantics of compound event-logic expressions is defined as follows:

- $\neg\Phi$ denotes the non-occurrence of $\Phi$. An occurrence of $\neg\Phi$ coincides with **i** if no occurrence of $\Phi$ coincides with **i**. Note that $(\neg\Phi)@\mathbf{i}$ could be true, even if an occurrence of $\Phi$ *overlapped* with **i**, so long as no occurrence of $\Phi$ *coincided* with **i**.

- $\Phi\vee\Psi$ denotes the occurrence of either $\Phi$ or $\Psi$.

- $\Phi\wedge_R\Psi$ denotes the occurrence of both $\Phi$ and $\Psi$. The occurrences of $\Phi$ and $\Psi$ need not be simultaneous. The subscript $R$ specifies a set of allowed Allen relations between the occurrences of $\Phi$ and $\Psi$. If occurrences of $\Phi$ and $\Psi$ coincide with **i** and **j** respectively, and $\mathbf{i}r\mathbf{j}$ for some $r\in R$, then an occurrence of $\Phi\wedge_R\Psi$ coincides with the span of **i** and **j**. I abbreviate the special case $\Phi\wedge_{\{=\}}\Psi$ simply as $\Phi\wedge\Psi$ without any subscript. $\Phi\wedge\Psi$ describes an aggregate event where both $\Phi$ and $\Psi$ occur simultaneously. I also abbreviate the special case $\Phi\wedge_{\{\mathsf{m}\}}\Psi$ as $\Phi;\Psi$. $\Phi;\Psi$ describes an aggregate event where an occurrence of $\Phi$ is immediately followed by an occurrence of $\Psi$.

2. I am using the term ‘overlap’ here in a different sense than the $\mathsf{o}$ relation from Allen (1983). Here, I am using the term overlap in the sense that two intervals overlap if they have a nonempty intersection. This corresponds to the union of the $\mathsf{o}$, $\mathsf{oi}$, $\mathsf{s}$, $\mathsf{si}$, $\mathsf{f}$, $\mathsf{fi}$, $\mathsf{d}$, and $\mathsf{di}$ relations from Allen (1983).

3. To deal with noise, in the actual implementation, the primitive event type $\overline{\phi}$ is derived from the predicate $\phi$ by first passing $\phi$ through a low-pass filter that takes $\phi(t)$ to be the majority vote of a five-frame window centered on $t$.

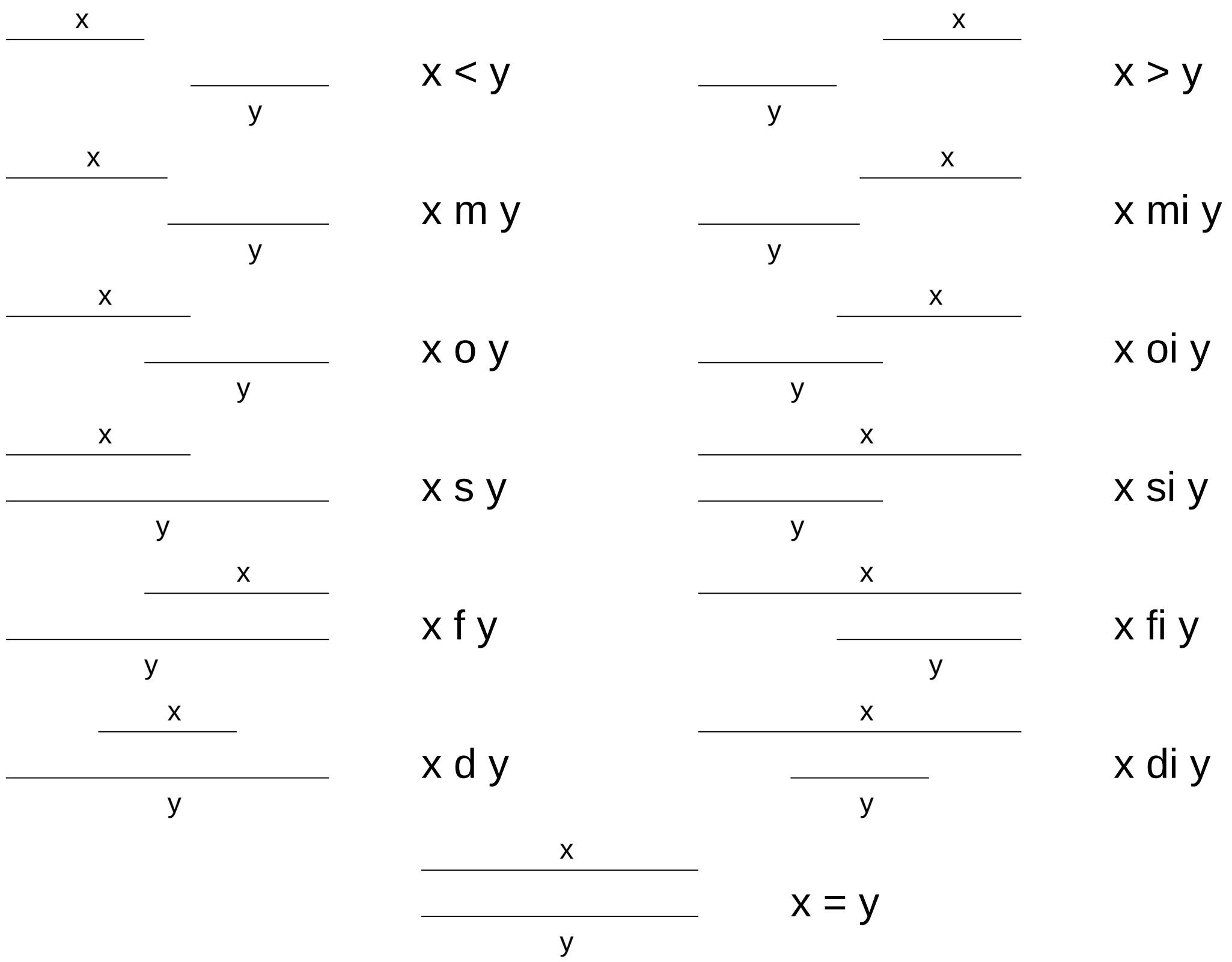

Figure 8: The Allen relations, the thirteen possible relations between two intervals.

- An occurrence of $\Diamond_R \Phi$ coinciding with $\mathbf{i}$ denotes an occurrence of $\Phi$ at some other interval $\mathbf{j}$ such that $\mathbf{j}r\mathbf{i}$ for some $r \in R$. $\Diamond_R$ can act as a tense operator. Expressions such as $\Diamond_{\{<\}}\Phi$, $\Diamond_{\{>\}}\Phi$, $\Diamond_{\{\mathsf{m}\}}\Phi$, and $\Diamond_{\{\mathsf{mi}\}}\Phi$ specify that $\Phi$ happened in the noncontiguous past, noncontiguous future, contiguous past, or contiguous future respectively. The $\Diamond_R$ operator can also be used to derive overlapped occurrence from coincidental occurrence. An occurrence of $\Diamond_{\{=,\mathsf{o},\mathsf{oi},\mathsf{s},\mathsf{si},\mathsf{f},\mathsf{fi},\mathsf{d},\mathsf{di}\}}\Phi$ coincides with $\mathbf{i}$ if an occurrence of $\Phi$ overlaps with $\mathbf{i}$. I abbreviate $\Diamond_{\{=,\mathsf{o},\mathsf{oi},\mathsf{s},\mathsf{si},\mathsf{f},\mathsf{fi},\mathsf{d},\mathsf{di}\}}\Phi$ simply as $\Diamond\Phi$ without any subscript. Note that while $(\neg\Phi)@\mathbf{i}$ indicates that no occurrence of $\Phi$ *coincided* with $\mathbf{i}$, $(\neg\Diamond\Phi)@\mathbf{i}$ indicates that no occurrence of $\Phi$ *overlapped* with $\mathbf{i}$.

Formally, the truth of an atomic event-occurrence formula $\Phi@\mathbf{i}$ is defined relative to a model. Let $I$ be the set of all intervals and $O$ be the set of all objects in the movie. A model $M$ is a map from primitive event-type symbols of arity $n$ to subsets of $I \times \underbrace{O \times \cdots \times O}_{n}$. ($M$ can be viewed as either a <u>m</u>odel or as a <u>m</u>ovie.) $M$ thus maps primitive event-type symbols to relations that take an interval as their first argument, in addition to the remaining object parameters. The semantics of event logic is formally defined by specifying an entailment relation $M \models \Phi@\mathbf{i}$ as follows:

- $M \models p(t_1, \ldots, t_n)@\mathbf{i}$ if and only if $\langle \mathbf{i}, t_1, \ldots, t_n \rangle \in M(p)$.
- $M \models (\neg\Phi)@\mathbf{i}$ if and only if $M \not\models \Phi@\mathbf{i}$.
- $M \models (\Phi \vee \Psi)@\mathbf{i}$ if and only if $M \models \Phi@\mathbf{i}$ or $M \models \Psi@\mathbf{i}$.
- $M \models (\Phi \wedge_R \Psi)@\mathbf{i}$ if and only if there exist two intervals $\mathbf{j}$ and $\mathbf{k}$ such that $\mathbf{i} = \textsc{Span}(\mathbf{j}, \mathbf{k})$, $\mathbf{j}r\mathbf{k}$ for some $r \in R$, $M \models \Phi@\mathbf{j}$, and $M \models \Psi@\mathbf{k}$.
- $M \models (\Diamond_R \Phi)@\mathbf{i}$ if and only if there exists some interval $\mathbf{j}$ such that $\mathbf{j}r\mathbf{i}$ for some $r \in R$ and $M \models \Phi@\mathbf{j}$.

Figure 9 shows the primitive event types currently used by LEONARD. The definitions in Figure 9 are formulated in terms of the predicates GROUNDED, RIGID, REVOLUTE, and SAMELAYER that are produced by model reconstruction as described in Section 2. An object is supported if it is not grounded. Two objects contact if their polygons touch and they are on the same layer. Two polygons $p$ and $q$ touch, denoted $\textsc{Touches}(p, q)$, if they intersect and their intersection has zero area. Two objects are attached if there is a rigid or revolute joint that joins them. Determining whether an object $x$ supports an object $y$ is a little more complex and requires counterfactual reasoning. Let $\mathcal{P}$ be the set of polygons in a scene, $I = \langle \textsc{Grounded}, \textsc{Rigid}, \textsc{Revolute}, \textsc{SameLayer} \rangle$ be the most-preferred model of the scene as produced by model reconstruction, and $\textsc{Stable}(\mathcal{P}, I)$ be true if the scene $\mathcal{P}$ is stable under an interpretation $I$. Stability analysis, i.e. the STABLE predicate, is a subroutine used by the model-reconstruction component. An object $x$ supports an object $y$ if $y$ is not grounded in the most-preferred model $I$ and a variant of $\mathcal{P}$ with $x$ removed is not stable under a variant of $I$ where all objects except for $y$ and those rigidly attached, directly or indirectly, to $y$ are grounded. In Figure 9, $\mathcal{P} \setminus \{x\}$ denotes the variant of $\mathcal{P}$ with $x$ removed, $\textsc{RigAtt}(x, y)$ denotes the fact that $x$ is rigidly attached to $y$, $\textsc{RigAtt}^*$ denotes the reflexive transitive closure of the RIGATT relation, $\{z | \neg\textsc{RigAtt}^*(z, y)\}$ denotes the set of objects that are rigidly attached, directly or indirectly, to $y$, and $\langle \textsc{Grounded} \cup \{z | \neg\textsc{RigAtt}^*(z, y)\}, \textsc{Rigid}, \textsc{Revolute}, \textsc{SameLayer} \rangle$ denotes

$$
\begin{array}{rcl}
x = y & \triangleq & \overline{x = y} \\
\textsc{Supported}(x) & \triangleq & \overline{\neg\textsc{Grounded}(x)} \\
\textsc{RigAtt}(x, y) & \triangleq & (\exists r)\textsc{Rigid}(x, y, r) \\
\textsc{Supports}(x, y) & \triangleq & \overline{\left[\begin{array}{l} \neg\textsc{Grounded}(y)\wedge \\ \neg\textsc{Stable}\left(\mathcal{P}\setminus\{x\}, \left\langle \begin{array}{l} \textsc{Grounded}\cup\{z|\neg\textsc{RigAtt}^*(z, y)\}, \\ \textsc{Rigid}, \\ \textsc{Revolute}, \\ \textsc{SameLayer} \end{array} \right\rangle\right) \end{array}\right]} \\
\textsc{Contacts}(x, y) & \triangleq & \overline{\textsc{Touches}(x, y) \wedge \textsc{SameLayer}(x, y)} \\
\textsc{Attached}(x, y) & \triangleq & \overline{(\exists r)\textsc{Rigid}(x, y, r) \vee \textsc{Revolute}(x, y, r)}
\end{array}
$$

Figure 9: Definition of the primitive event types used by LEONARD.

a variant of $I$ where all objects except for $y$ and those rigidly attached, directly or indirectly, to $y$ are grounded.

Figure 10 shows the compound event-type definitions currently used by LEONARD. PICKUP$(x, y, z)$ denotes an event type where $x$ picks $y$ up off of $z$. It is specified as a sequence of three intervals, where $x$ is not attached to and does not support $y$ in the first interval but is attached to and does support $y$ in the third interval. Additionally, $z$ supports $y$ in the first interval but does not support $y$ in the third interval. Furthermore, several conditions must hold in both the first and third intervals: $x$ must be unsupported, $y$ must not support either $x$ or $z$, $x$ and $z$ must not support each other, and $y$ must not be attached to $z$. During the second interval, intermediate between the first and third intervals, either $x$ is attached to $y$ or $y$ is attached to $z$.[4] Additionally, several conditions must hold throughout the entire event: $x$, $y$, and $z$ must be distinct and $y$ must be supported. PUTDOWN$(x, y, z)$ denotes an event type where $x$ puts $y$ down on $z$. It is specified in a fashion that is similar to PICKUP$(x, y, z)$ but where the three subevents occur in reverse order. STACK$(w, x, y, z)$ denotes an event type where $w$ puts $x$ down on $y$ which is resting on $z$. It is specified as PUTDOWN$(w, x, y)$, where $z$ supports but is not attached to $y$ and $z$ is distinct from $w$, $x$, and $y$. UNSTACK$(w, x, y, z)$ denotes an event type where $w$ picks $x$ up off of $y$ which is resting on $z$. It is specified as PICKUP$(w, x, y)$, where $z$ supports but is not attached to $y$ and $z$ is distinct from $w$, $x$, and $y$. MOVE$(w, x, y, z)$ denotes an event type where $w$ picks $x$ up off of $y$ and puts it down on $z$ which is distinct from $y$. ASSEMBLE$(w, x, y, z)$ denotes an event type where $w$ first puts $y$ down on $z$ then sometime later stacks $x$ on top of $y$. Finally, DISASSEMBLE$(w, x, y, z)$ denotes

4. Originally, a two-interval definition was used, consisting of only the first and third intervals. Such a definition better reflects human intuition. This requires that $x$ be unsupported, $y$ not support either $x$ or $z$, $x$ and $z$ not support each other, and $y$ not be attached to $z$ throughout the event. Unfortunately, however, the model-reconstruction process has some quirks. When the hand grasps the patient while the patient is still resting on the source it produces a most-preferred model where all three objects are attached and collectively supported by one of them being grounded. While such a model is consistent with the theory and is minimal, it does not match human intuition. Pending improvements in the model-reconstruction process to better reflect human intuition, the compound event-type definition for *pick up* was modified to reflect the force-dynamic interpretations produced by the current model-reconstruction process. Fortunately, the current model-reconstruction process is robust in reproducing this counterintuitive interpretation.

an event type where $w$ first unstacks $x$ from on top of $y$ (which is resting on $z$) and then sometime later picks $y$ up off of $z$. Figure 1 shows sample movies depicting occurrences of the event types PICKUP$(x, y, z)$ and PUTDOWN$(x, y, z)$. Figures 11 through 15 show sample movies depicting occurrences of the event types STACK$(w, x, y, z)$, UNSTACK$(w, x, y, z)$, MOVE$(w, x, y, z)$, ASSEMBLE$(w, x, y, z)$, and DISASSEMBLE$(w, x, y, z)$ respectively.

Nominally, all atomic event-logic expressions are primitive event types. However, we allow giving a name to a compound event-logic expression and using this name in another event-logic expression as short hand for the named expression with appropriate parameter substitution. This is simply a macro-expansion process and, as such, no recursion is allowed. This feature is used in Figure 10 to define UNSTACK, MOVE, and DISASSEMBLE in terms of PICKUP; STACK, MOVE, and ASSEMBLE in terms of PUTDOWN; ASSEMBLE in terms of STACK, which is itself defined in terms of PUTDOWN; and DISASSEMBLE in terms of UNSTACK, which is itself defined in terms of PICKUP.

The overall goal of the event-classification component is to infer all occurrences of a given set of compound event types from a given set of primitive event occurrences. The model-reconstruction component combined with the primitive event-type definitions given in Figure 9 produces a set of primitive event occurrences for a given scene sequence. Figure 10 lists parameterized compound event types. These are instantiated for all tuples of objects in the scene sequence to yield ground compound event-logic expressions. The event-classification component infers all occurrences of these compound event types that follow from the set of primitive event occurrences. Let us define $\mathcal{E}(M, \Phi)$ to be $\{\mathbf{i} | M \models \Phi @ \mathbf{i}\}$. The model-reconstruction component combined with the primitive event-type definitions given in Figure 9 produces $M$. Instantiating the parameterized compound event types from Figure 10 for all object tuples yields a set of event-logic expressions. The event-classification component computes $\mathcal{E}(M, \Phi)$ for every $\Phi$ in this set.

In principle, $\mathcal{E}(M, \Phi)$ could by implemented as a straightforward application of the formal semantics for event logic as specified above. There is a difficulty in doing so, however. The primitive event types have the property that they are *liquid*. Liquid events have the following two properties. First, if they are true during an interval $\mathbf{i}$, then they are also true during any subinterval of $\mathbf{i}$. Second, if they are true during two overlapping intervals $\mathbf{i}$ and $\mathbf{j}$, then they are also true during SPAN$(\mathbf{i}, \mathbf{j})$ and any subinterval of SPAN$(\mathbf{i}, \mathbf{j})$. For example, if an object is supported during $[1, 10]$, then it also is supported during $[2, 5]$, $[3, 8]$, and all other subintervals of $[1, 10]$. Similarly, if an object is supported during $[1, 5]$ and $[4, 10]$, then it also is supported during $[1, 10]$ and all of its subintervals. Shoham (1987) introduced the notion of liquidity and Vendler (1967), Dowty (1979), Verkuyl (1989), and Krifka (1992) have observed that many event types have this property. Because the primitive event types are liquid, they will hold over an infinite number of subintervals. This renders the formal semantics inappropriate for a computational implementation. Even if one limits oneself to intervals with integral endpoints, the primitive event types will hold over quadratically many subintervals of the scene sequence. Furthermore, a straightforward computational implementation of the formal semantics would be inefficient, because it requires quantifying over subintervals to implement $\Diamond_R$ and quantifying over pairs of subintervals to implement $\wedge_R$. The central result of this paper is a novel representation, called *spanning intervals*, that allows an efficient representation of the infinite sets of subintervals over which liquid event types hold along with an efficient inference procedure that operates on that representation. This representation, and the inference procedure that implements $\mathcal{E}(M, \Phi)$, are presented in the next section.

$$
\textsc{PickUp}(x,y,z) \triangleq \left(\begin{array}{l}
\neg\Diamond x=y \wedge \neg\Diamond z=x \wedge \neg\Diamond z=y\wedge \\
\textsc{Supported}(y) \wedge \neg\Diamond\textsc{Attached}(x,z)\wedge \\
\left\{\begin{array}{l}
\left[\begin{array}{l}
\neg\Diamond\textsc{Attached}(x,y) \wedge \neg\Diamond\textsc{Supports}(x,y)\wedge \\
\textsc{Supports}(z,y)\wedge \\
\neg\Diamond\textsc{Supported}(x) \wedge \neg\Diamond\textsc{Attached}(y,z)\wedge \\
\neg\Diamond\textsc{Supports}(y,x) \wedge \neg\Diamond\textsc{Supports}(y,z)\wedge \\
\neg\Diamond\textsc{Supports}(x,z) \wedge \neg\Diamond\textsc{Supports}(z,x)
\end{array}\right]; \\
{[\textsc{Attached}(x,y) \vee \textsc{Attached}(y,z)]}\,; \\
\left[\begin{array}{l}
\textsc{Attached}(x,y) \wedge \textsc{Supports}(x,y)\wedge \\
\neg\Diamond\textsc{Supports}(z,y)\wedge \\
\neg\Diamond\textsc{Supported}(x) \wedge \neg\Diamond\textsc{Attached}(y,z)\wedge \\
\neg\Diamond\textsc{Supports}(y,x) \wedge \neg\Diamond\textsc{Supports}(y,z)\wedge \\
\neg\Diamond\textsc{Supports}(x,z) \wedge \neg\Diamond\textsc{Supports}(z,x)
\end{array}\right]
\end{array}\right\}
\end{array}\right)
$$

$$
\textsc{PutDown}(x,y,z) \triangleq \left(\begin{array}{l}
\neg\Diamond x=y \wedge \neg\Diamond z=x \wedge \neg\Diamond z=y\wedge \\
\textsc{Supported}(y) \wedge \neg\Diamond\textsc{Attached}(x,z)\wedge \\
\left\{\begin{array}{l}
\left[\begin{array}{l}
\textsc{Attached}(x,y) \wedge \textsc{Supports}(x,y)\wedge \\
\neg\Diamond\textsc{Supports}(z,y)\wedge \\
\neg\Diamond\textsc{Supported}(x) \wedge \neg\Diamond\textsc{Attached}(y,z)\wedge \\
\neg\Diamond\textsc{Supports}(y,x) \wedge \neg\Diamond\textsc{Supports}(y,z)\wedge \\
\neg\Diamond\textsc{Supports}(x,z) \wedge \neg\Diamond\textsc{Supports}(z,x)
\end{array}\right]; \\
{[\textsc{Attached}(x,y) \vee \textsc{Attached}(y,z)]}\,; \\
\left[\begin{array}{l}
\neg\Diamond\textsc{Attached}(x,y) \wedge \neg\Diamond\textsc{Supports}(x,y)\wedge \\
\textsc{Supports}(z,y)\wedge \\
\neg\Diamond\textsc{Supported}(x) \wedge \neg\Diamond\textsc{Attached}(y,z)\wedge \\
\neg\Diamond\textsc{Supports}(y,x) \wedge \neg\Diamond\textsc{Supports}(y,z)\wedge \\
\neg\Diamond\textsc{Supports}(x,z) \wedge \neg\Diamond\textsc{Supports}(z,x)
\end{array}\right]
\end{array}\right\}
\end{array}\right)
$$

$$
\textsc{Stack}(w,x,y,z) \triangleq \left[\begin{array}{l}
\neg\Diamond z=w \wedge \neg\Diamond z=x \wedge \neg\Diamond z=y\wedge \\
\textsc{PutDown}(w,x,y) \wedge \textsc{Supports}(z,y)\wedge \\
\neg\textsc{Attached}(z,y)
\end{array}\right]
$$

$$
\textsc{Unstack}(w,x,y,z) \triangleq \left[\begin{array}{l}
\neg\Diamond z=w \wedge \neg\Diamond z=x \wedge \neg\Diamond z=y\wedge \\
\textsc{PickUp}(w,x,y) \wedge \textsc{Supports}(z,y) \wedge \neg\textsc{Attached}(z,y)
\end{array}\right]
$$

$$
\textsc{Move}(w,x,y,z) \triangleq \neg\Diamond y=z \wedge [\textsc{PickUp}(w,x,y); \textsc{PutDown}(w,x,z)]
$$

$$
\textsc{Assemble}(w,x,y,z) \triangleq \textsc{PutDown}(w,y,z) \wedge_{\{<\}} \textsc{Stack}(w,x,y,z)
$$

$$
\textsc{Disassemble}(w,x,y,z) \triangleq \textsc{Unstack}(w,x,y,z) \wedge_{\{<\}} \textsc{PickUp}(x,y,z)
$$

Figure 10: The lexicon of compound event types used by LEONARD.

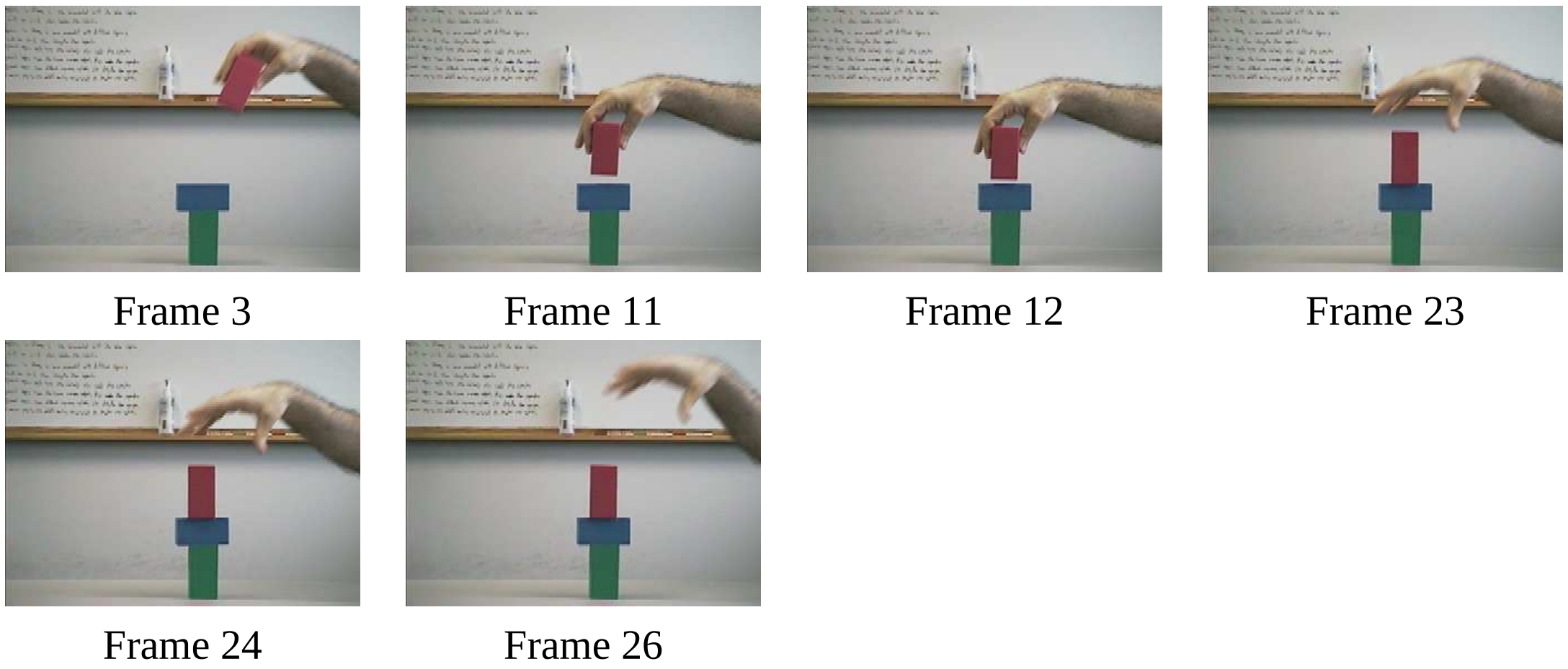

Figure 11: An image sequence depicting a *stack* event.

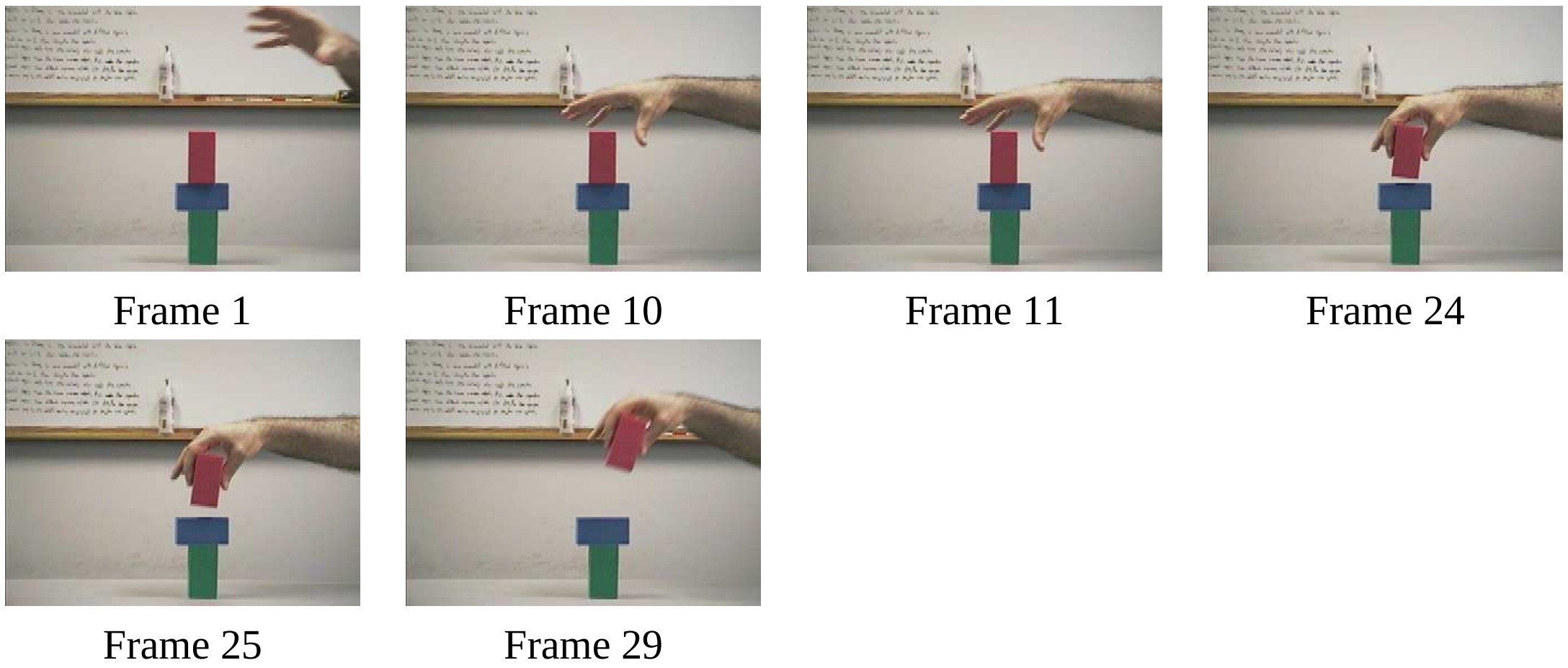

Figure 12: An image sequence depicting an *unstack* event.

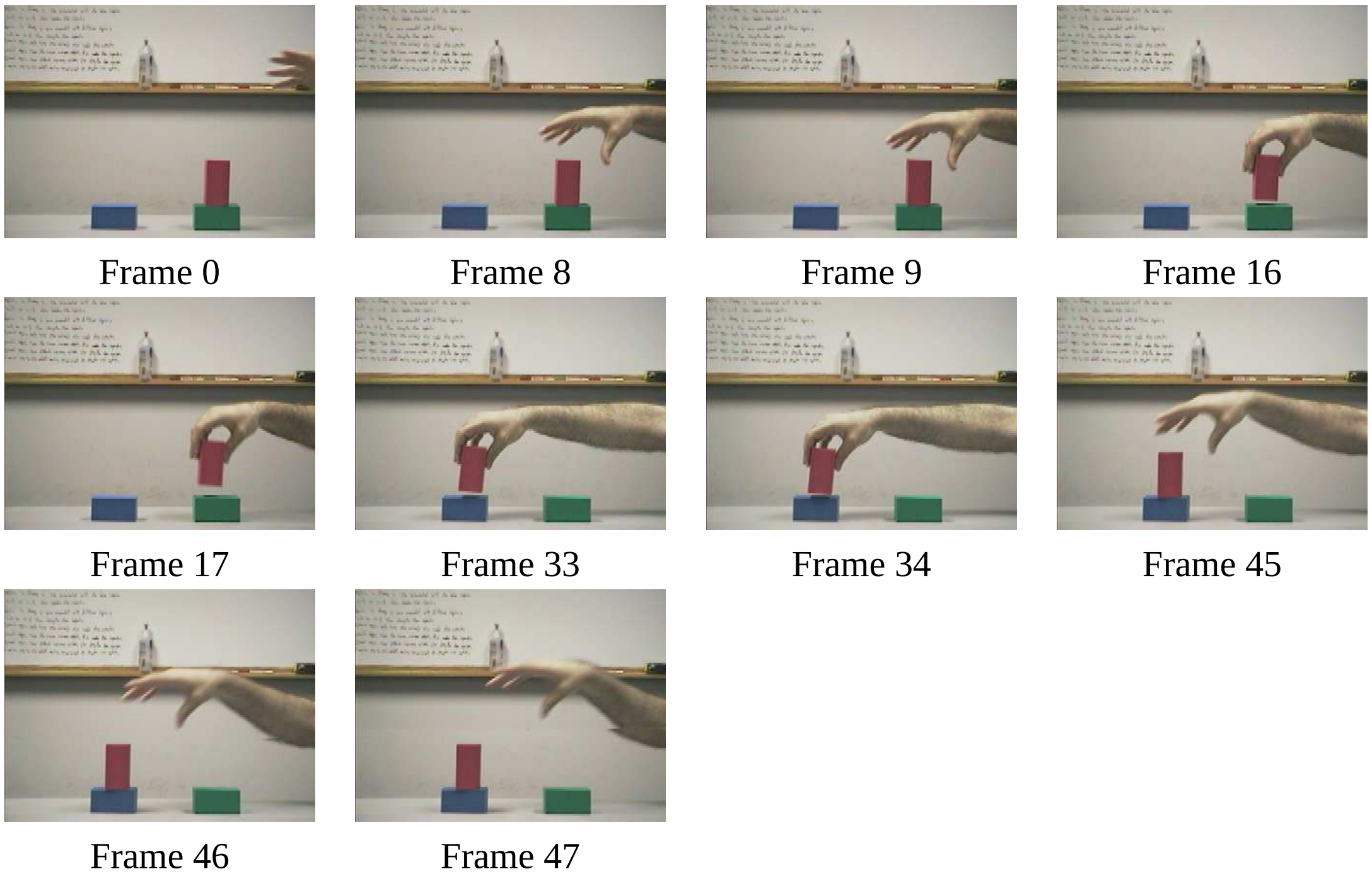

Figure 13: An image sequence depicting a *move* event.

## 4. An Efficient Representation and Inference Procedure for Event Logic

One might try to implement event logic using only closed intervals of the form $[q, r]$, where $q \leq r$. Such a closed interval would represent the set $\{p|q \leq p \leq r\}$ of real numbers. With such closed intervals, one would define Allen's relations as follows:

$$
\begin{aligned}
[q_1, r_1] = [q_2, r_2] &\triangleq (q_1 = q_2) \wedge (r_1 = r_2) \\
[q_1, r_1] < [q_2, r_2] &\triangleq r_1 < q_2 \\
[q_1, r_1] > [q_2, r_2] &\triangleq q_1 > r_2 \\
[q_1, r_1] \mathsf{\ m\ } [q_2, r_2] &\triangleq r_1 = q_2 \\
[q_1, r_1] \mathsf{\ mi\ } [q_2, r_2] &\triangleq q_1 = r_2 \\
[q_1, r_1] \mathsf{\ o\ } [q_2, r_2] &\triangleq q_1 < q_2 < r_1 < r_2 \\
[q_1, r_1] \mathsf{\ oi\ } [q_2, r_2] &\triangleq r_1 > r_2 > q_1 > q_2 \\
[q_1, r_1] \mathsf{\ s\ } [q_2, r_2] &\triangleq (q_1 = q_2) \wedge (r_1 < r_2) \\
[q_1, r_1] \mathsf{\ si\ } [q_2, r_2] &\triangleq (q_1 = q_2) \wedge (r_1 > r_2) \\
[q_1, r_1] \mathsf{\ f\ } [q_2, r_2] &\triangleq (q_1 > q_2) \wedge (r_1 = r_2) \\
[q_1, r_1] \mathsf{\ fi\ } [q_2, r_2] &\triangleq (q_1 < q_2) \wedge (r_1 = r_2) \\
[q_1, r_1] \mathsf{\ d\ } [q_2, r_2] &\triangleq (q_1 > q_2) \wedge (r_1 < r_2)
\end{aligned}
$$

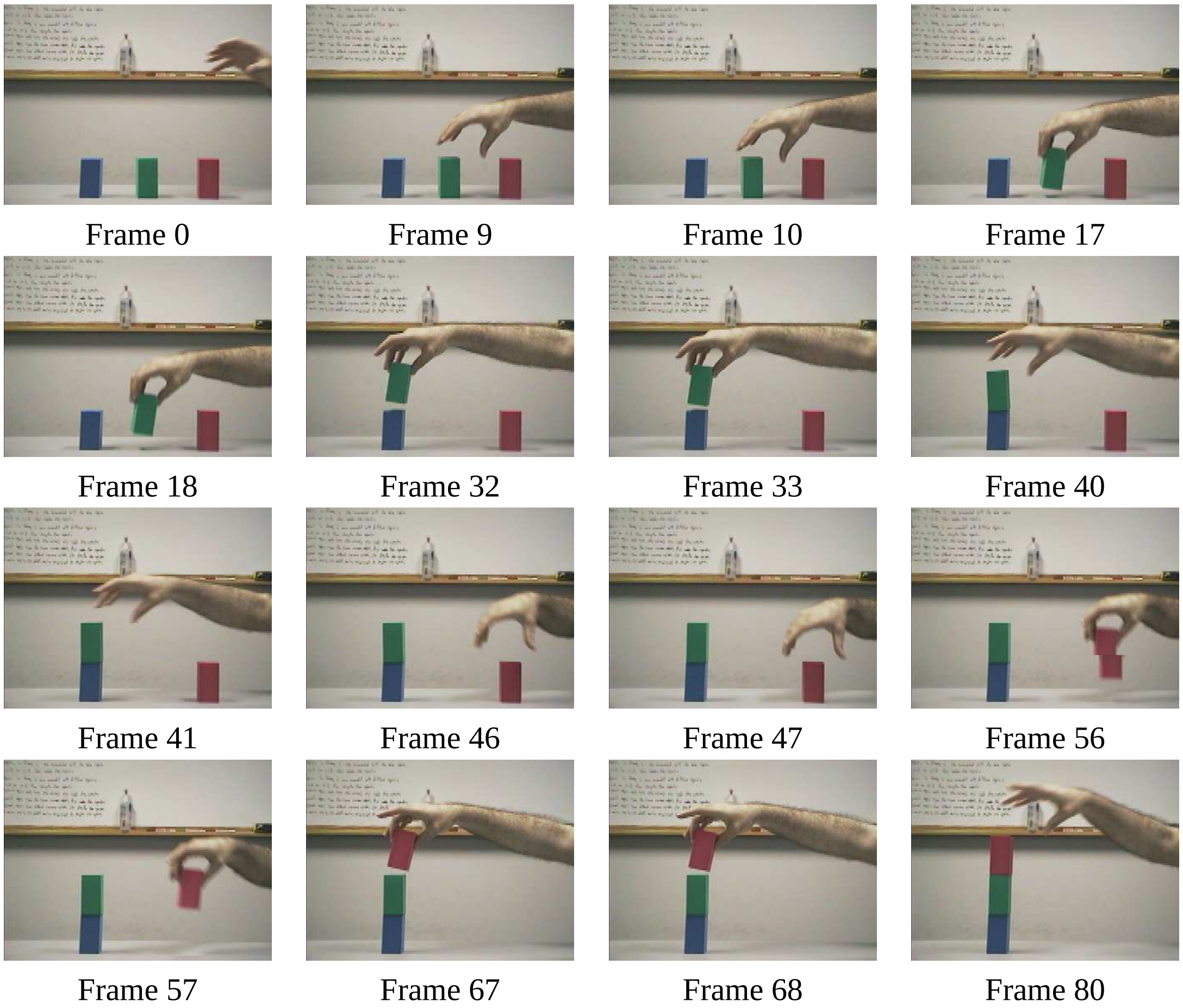

Figure 14: An image sequence depicting an *assemble* event.

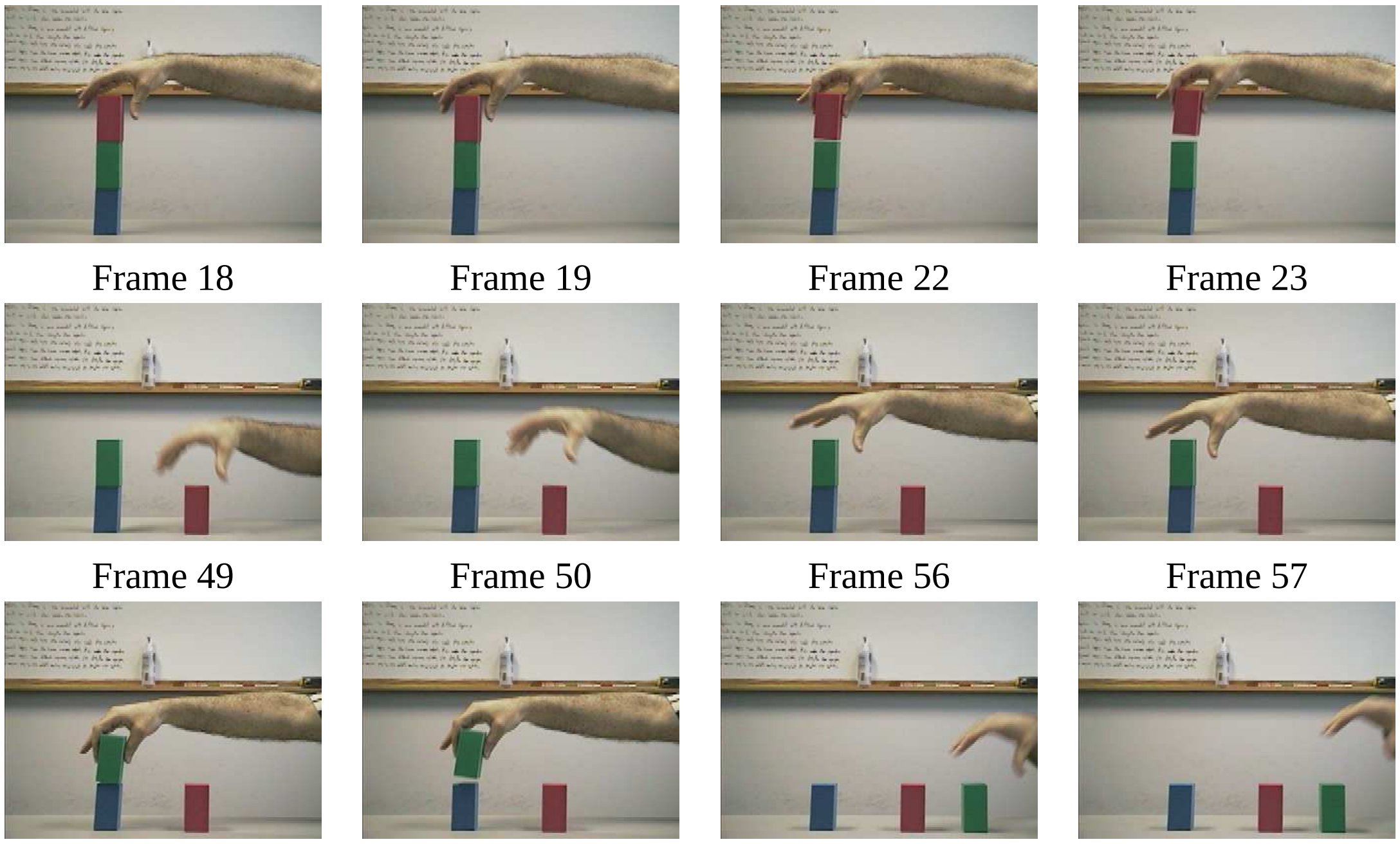

Frame 18 Frame 19 Frame 22 Frame 23
Frame 49 Frame 50 Frame 56 Frame 57
Frame 62 Frame 63 Frame 87 Frame 88

Figure 15: An image sequence depicting a *disassemble* event.

$$[q_1, r_1]\ \mathsf{di}\ [q_2, r_2] \quad \triangleq \quad (q_1 < q_2) \wedge (r_1 > r_2)$$

One difficulty with doing so is that it would be possible for more than one Allen relation to hold between two intervals when one or both of them are instantaneous intervals, such as $[q, q]$. Both m and s would hold between $[q, q]$ and $[q, r]$, both mi and si would hold between $[q, r]$ and $[q, q]$, both m and fi would hold between $[q, r]$ and $[r, r]$, both mi and f would hold between $[r, r]$ and $[q, r]$, and $=$, m, and mi would all hold between $[q, q]$ and itself. To create a domain where exactly one Allen relation holds between any pair of intervals, let us consider both open and closed intervals. Closed intervals contain their endpoints while open intervals do not. The intervals $(q, r]$, $[q, r)$, and $(q, r)$, where $q < r$, represent the sets $\{p|q < p \leq r\}$, $\{p|q \leq p < r\}$, and $\{p|q < p < r\}$ of real numbers respectively. The various kinds of open and closed intervals can be unified into a single representation ${}_\alpha[q, r]_\beta$, where $\alpha$ and $\beta$ are true or false to indicate the interval being closed or open on the left or right respectively.[5] More specifically, ${}_\mathbf{T}[q, r]_\mathbf{T}$ denotes $[q, r]$, ${}_\mathbf{F}[q, r]_\mathbf{T}$ denotes $(q, r]$, ${}_\mathbf{T}[q, r]_\mathbf{F}$ denotes $[q, r)$, and ${}_\mathbf{F}[q, r]_\mathbf{F}$ denotes $(q, r)$. To do this, let us use $q \leq_\alpha r$ to mean $q \leq r$ when $\alpha$ is true and $q < r$ when $\alpha$ is false. Similarly, let us use $q \geq_\alpha r$ to mean $q \geq r$ when $\alpha$ is true and $q > r$ when $\alpha$ is false. More precisely, $q \leq_\alpha r \triangleq (q < r) \vee [\alpha \wedge (q = r)]$ and $q \geq_\alpha r \triangleq (q > r) \vee [\alpha \wedge (q = r)]$. With these, ${}_\alpha[q, r]_\beta$ represents the set $\{p|q \leq_\alpha p \leq_\beta r\}$ of real numbers.

5. Throughout this paper, I use lowercase Greek letters, such as $\alpha$, $\beta$, $\gamma$, $\delta$, $\epsilon$, and $\zeta$, to denote Boolean values, lowercase Latin letters, such as $p$, $q$, $r$, $i$, $j$, $k$, and $l$, to denote real numbers, lowercase bold Latin letters, such as **i**, **j**, and **k**, to denote intervals or spanning intervals, and uppercase Latin letters, such as $I$, $J$, and $K$, to denote sets of intervals or sets of spanning intervals.

One can extend the definition of Allen's relations to both open and closed intervals as follows. The relation $\mathbf{i}_1 = \mathbf{i}_2$ holds if the corresponding endpoints of $\mathbf{i}_1$ and $\mathbf{i}_2$ are equal and have the same openness. The relation $\mathbf{i}_1 < \mathbf{i}_2$ holds if the right endpoint of $\mathbf{i}_1$ precedes the left endpoint of $\mathbf{i}_2$ or if they are equal and both open. For example, $[1,3] < [4,5]$ and $[1,3) < (3,5]$, but $[1,3] \not< (3,5]$, $[1,3) \not< [3,5]$, and $[1,3] \not< [3,5]$. The relation $\mathbf{i}_1 \;\mathsf{m}\; \mathbf{i}_2$ holds if the right endpoint of $\mathbf{i}_1$ equals the left endpoint of $\mathbf{i}_2$ and one of those endpoints is open while the other is closed. For example, $[1,3] \;\mathsf{m}\; (3,5]$ and $[1,3) \;\mathsf{m}\; [3,5]$ but $[1,3] \;\not\mathsf{m}\; [3,5]$ and $[1,3) \;\not\mathsf{m}\; (3,5]$. The relation $\mathbf{i}_1 \;\mathsf{o}\; \mathbf{i}_2$ holds if

- either the left endpoint of $\mathbf{i}_1$ precedes the left endpoint of $\mathbf{i}_2$ or they are equal while the former is closed and the latter is open,
- either the left endpoint of $\mathbf{i}_2$ precedes the right endpoint of $\mathbf{i}_1$ or they are equal while both endpoints are closed, and
- either the right endpoint of $\mathbf{i}_1$ precedes the right endpoint of $\mathbf{i}_2$ or they are equal while the former is open and the latter is closed.

For example, $[1,3] \;\mathsf{o}\; [2,4]$, $[1,3] \;\mathsf{o}\; (1,4]$, $[1,2] \;\mathsf{o}\; [2,4]$, and $[1,4) \;\mathsf{o}\; [2,4]$, but $[1,3] \;\not\mathsf{o}\; [1,4]$, $[1,2) \;\not\mathsf{o}\; [2,4]$, and $[1,4] \;\not\mathsf{o}\; [2,4]$. The relation $\mathbf{i}_1 \;\mathsf{s}\; \mathbf{i}_2$ holds if

- the left endpoints of $\mathbf{i}_1$ and $\mathbf{i}_2$ are equal and have the same openness and
- either the right endpoint of $\mathbf{i}_1$ precedes the right endpoint of $\mathbf{i}_2$ or they are equal while the former is open and the latter is closed.

For example, $[1,3] \;\mathsf{s}\; [1,4]$, $(1,3] \;\mathsf{s}\; (1,4]$, and $[1,3) \;\mathsf{s}\; [1,3]$, but $[1,3] \;\not\mathsf{s}\; (1,4]$, $[1,3] \;\not\mathsf{s}\; [1,3]$, $[1,3) \;\not\mathsf{s}\; [1,3)$, and $[1,3] \;\not\mathsf{s}\; [1,3)$. The relation $\mathbf{i}_1 \;\mathsf{f}\; \mathbf{i}_2$ holds if

- the right endpoints of $\mathbf{i}_1$ and $\mathbf{i}_2$ are equal and have the same openness and
- either the left endpoint of $\mathbf{i}_1$ follows the left endpoint of $\mathbf{i}_2$ or they are equal while the former is open and the latter is closed.

For example, $[2,4] \;\mathsf{f}\; [1,4]$, $[2,4) \;\mathsf{f}\; [1,4)$, and $(2,4] \;\mathsf{f}\; [2,4]$, but $[2,4) \;\not\mathsf{f}\; [1,4]$, $(2,4] \;\not\mathsf{f}\; (2,4]$, $[2,4] \;\not\mathsf{f}\; [2,4]$, and $[2,4] \;\not\mathsf{f}\; (2,4]$. The relation $\mathbf{i}_1 \;\mathsf{d}\; \mathbf{i}_2$ holds if

- either the left endpoint of $\mathbf{i}_1$ follows the left endpoint of $\mathbf{i}_2$ or they are equal while the former is open and the latter is closed and
- either the right endpoint of $\mathbf{i}_1$ precedes the right endpoint of $\mathbf{i}_2$ or they are equal while the former is open and the latter is closed.

For example, $[2,3] \;\mathsf{d}\; [1,4]$ and $(1,4) \;\mathsf{d}\; [1,4]$, but $[1,4) \;\not\mathsf{d}\; [1,4]$, $(1,4] \;\not\mathsf{d}\; [1,4]$, $(1,4) \;\not\mathsf{d}\; (1,4]$, and $(1,4) \;\not\mathsf{d}\; [1,4)$. The inverse Allen relations $>$, mi, oi, si, fi, and di are defined analogously to the $<$, m, o, s, f, and d relations respectively with the arguments reversed.

The above definitions can be stated more precisely as follows:

$$_{\alpha_1}[q_1, r_1]_{\beta_1} =_{\alpha_2} [q_2, r_2]_{\beta_2} \;\triangleq\; (q_1 = q_2) \wedge (\alpha_1 = \alpha_2) \wedge (r_1 = r_2) \wedge (\beta_1 = \beta_2) \quad (1)$$

$$_{\alpha_1}[q_1, r_1]_{\beta_1} <_{\alpha_2} [q_2, r_2]_{\beta_2} \;\triangleq\; r_1 \leq_{(\neg\beta_1 \wedge \neg\alpha_2)} q_2 \quad (2)$$

$$_{\alpha_1}[q_1, r_1]_{\beta_1} >_{\alpha_2} [q_2, r_2]_{\beta_2} \triangleq q_1 \geq_{(\neg\alpha_1 \wedge \neg\beta_2)} r_2 \tag{3}$$

$$_{\alpha_1}[q_1, r_1]_{\beta_1} \ \mathsf{m} \ _{\alpha_2}[q_2, r_2]_{\beta_2} \triangleq (r_1 = q_2) \wedge (\beta_1 \neq \alpha_2) \tag{4}$$

$$_{\alpha_1}[q_1, r_1]_{\beta_1} \ \mathsf{mi} \ _{\alpha_2}[q_2, r_2]_{\beta_2} \triangleq (q_1 = r_2) \wedge (\alpha_1 \neq \beta_2) \tag{5}$$

$$_{\alpha_1}[q_1, r_1]_{\beta_1} \ \mathsf{o} \ _{\alpha_2}[q_2, r_2]_{\beta_2} \triangleq q_1 \leq_{(\alpha_1 \wedge \neg\alpha_2)} q_2 \leq_{(\beta_1 \wedge \alpha_2)} r_1 \leq_{(\neg\beta_1 \wedge \beta_2)} r_2 \tag{6}$$

$$_{\alpha_1}[q_1, r_1]_{\beta_1} \ \mathsf{oi} \ _{\alpha_2}[q_2, r_2]_{\beta_2} \triangleq r_1 \geq_{(\beta_1 \wedge \neg\beta_2)} r_2 \geq_{(\alpha_1 \wedge \beta_2)} q_1 \geq_{(\neg\alpha_1 \wedge \alpha_2)} q_2 \tag{7}$$

$$_{\alpha_1}[q_1, r_1]_{\beta_1} \ \mathsf{s} \ _{\alpha_2}[q_2, r_2]_{\beta_2} \triangleq (q_1 = q_2) \wedge (\alpha_1 = \alpha_2) \wedge [r_1 \leq_{(\neg\beta_1 \wedge \beta_2)} r_2] \tag{8}$$

$$_{\alpha_1}[q_1, r_1]_{\beta_1} \ \mathsf{si} \ _{\alpha_2}[q_2, r_2]_{\beta_2} \triangleq (q_1 = q_2) \wedge (\alpha_1 = \alpha_2) \wedge [r_1 \geq_{(\beta_1 \wedge \neg\beta_2)} r_2] \tag{9}$$

$$_{\alpha_1}[q_1, r_1]_{\beta_1} \ \mathsf{f} \ _{\alpha_2}[q_2, r_2]_{\beta_2} \triangleq [q_1 \geq_{(\neg\alpha_1 \wedge \alpha_2)} q_2] \wedge (r_1 = r_2) \wedge (\beta_1 = \beta_2) \tag{10}$$

$$_{\alpha_1}[q_1, r_1]_{\beta_1} \ \mathsf{fi} \ _{\alpha_2}[q_2, r_2]_{\beta_2} \triangleq [q_1 \leq_{(\alpha_1 \wedge \neg\alpha_2)} q_2] \wedge (r_1 = r_2) \wedge (\beta_1 = \beta_2) \tag{11}$$

$$_{\alpha_1}[q_1, r_1]_{\beta_1} \ \mathsf{d} \ _{\alpha_2}[q_2, r_2]_{\beta_2} \triangleq [q_1 \geq_{(\neg\alpha_1 \wedge \alpha_2)} q_2] \wedge [r_1 \leq_{(\neg\beta_1 \wedge \beta_2)} r_2] \tag{12}$$

$$_{\alpha_1}[q_1, r_1]_{\beta_1} \ \mathsf{di} \ _{\alpha_2}[q_2, r_2]_{\beta_2} \triangleq [q_1 \leq_{(\alpha_1 \wedge \neg\alpha_2)} q_2] \wedge [r_1 \geq_{(\beta_1 \wedge \neg\beta_2)} r_2] \tag{13}$$

With the above definitions, exactly one Allen relation holds between any pair of intervals.

I refer to the set of real numbers represented by an interval as its *extension*. Given the above definition of interval, any interval, such as $[5, 4]$, $(5, 4]$, $[5, 4)$, or $(5, 4)$, where the upper endpoint is less than the lower endpoint represents the empty set. Furthermore, any open interval, such as $[5, 5)$, $(5, 5]$, or $(5, 5)$, where the upper endpoint equals the lower endpoint also represents the empty set. To create a situation where the extension of each interval has a unique representation, let us represent all such empty sets of real numbers as $\{\}$. Thus whenever we represent an interval $_\alpha[q, r]_\beta$ explicitly, it will have a nonempty extension and will satisfy the following normalization criterion: $q \leq_{(\alpha \wedge \beta)} r$.

## 4.1 Spanning Intervals

When using event logic, we wish to compute and represent the set $I$ of all intervals over which some event-logic expression $\Phi$ holds. Many event types, including all of the primitive event types used in LEONARD, are *liquid* (Shoham, 1987) in the sense that if some event holds of an interval then that event holds of every subinterval of that interval. With real-valued interval endpoints, this creates the need to compute and represent an infinite set of intervals for a liquid event. Even limiting ourselves to integer-valued interval endpoints, a liquid event will require the computation and representation of quadratically many intervals.

To address this problem, let us introduce the notion of *spanning interval*. A spanning interval $[i : j]$ represents the set of all subintervals of $[i, j]$, in other words $\{[q, r] | (i \leq q \leq j) \wedge (i \leq r \leq j)\}$. Similarly $(i : j]$, $[i : j)$, and $(i : j)$ represent $\{(q, r] | (i \leq q \leq j) \wedge (i \leq r \leq j)\}$, $\{[q, r) | (i \leq q \leq j) \wedge (i \leq r \leq j)\}$, and $\{(q, r) | (i \leq q \leq j) \wedge (i \leq r \leq j)\}$ respectively. We wish to use spanning intervals to represent the set of all intervals over which the primitive event types hold and to compute and represent the set of all intervals over which compound event types hold via structural induction over the compound event-logic expressions. A problem arises however. Given two liquid event types $\Phi$ and $\Psi$, the compound event type $\Phi; \Psi$ is not liquid. If $\Phi$ holds over $[i : j)$ and $\Psi$ holds over $[j : k)$, then $\Phi; \Psi$ might not hold over every subinterval of $[i, k)$. It holds over only those subintervals that include $j$. For example, if $\Phi$ holds over $[1:10)$ and $\Psi$ holds over $[8:20)$ then $\Phi;\Psi$ holds for every interval that starts between 1 and 10 and ends between 8 and 20. But it doesn't hold for every subinterval of $[1,20)$. For example, it doesn't hold of $[12,20)$. I refer to such event types as *semi liquid*. Since spanning intervals are not sufficient to efficiently represent semi-liquid events, let us extend the notion of spanning interval. A spanning interval $[[i,j],[k,l]]$ represents the set of intervals $\{[q,r]|(i \leq q \leq j) \wedge (k \leq r \leq l)\}$. Similarly the spanning intervals $([i,j],[k,l]]$, $[[i,j],[k,l])$, and $([i,j],[k,l])$ represent the sets $\{(q,r]|(i \leq q \leq j) \wedge (k \leq r \leq l)\}$, $\{[q,r)|(i \leq q \leq j) \wedge (k \leq r \leq l)\}$, and $\{(q,r)|(i \leq q \leq j) \wedge (k \leq r \leq l)\}$ respectively. This extended notion of spanning interval subsumes the original notion. The spanning intervals $[i:j]$, $(i:j]$, $[i:j)$, and $(i:j)$ can be represented as the spanning intervals $[[i,j],[i,j]]$, $([i,j],[i,j]]$, $[[i,j],[i,j])$, and $([i,j],[i,j])$ respectively. For reasons that will become apparent in Section 4.4, it is necessary to also allow for spanning intervals where the ranges of endpoint values are open. In other words, we will need to consider spanning intervals like $[(i,j],[k,l]]$ to represent sets like $\{[q,r]|(i < q \leq j) \wedge (i \leq r \leq j)\}$. All told, there are six endpoints that can independently be either open or closed, namely $q$, $r$, $i$, $j$, $k$, and $l$, yielding sixty four kinds of spanning intervals. These can all be unified into a single representation ${}_{\alpha}[{}_{\gamma}[i,j]_{\delta},{}_{\epsilon}[k,l]_{\zeta}]_{\beta}$, where $\alpha$, $\beta$, $\gamma$, $\delta$, $\epsilon$, and $\zeta$ are true or false if the endpoints $q$, $r$, $i$, $j$, $k$, and $l$ are closed or open respectively. More precisely, the spanning interval ${}_{\alpha}[{}_{\gamma}[i,j]_{\delta},{}_{\epsilon}[k,l]_{\zeta}]_{\beta}$ represents the set

$$\{{}_{\alpha}[q,r]_{\beta}|(i \leq_{\gamma} q \leq_{\delta} j) \wedge (k \leq_{\epsilon} r \leq_{\zeta} l)\} \tag{14}$$

of intervals. I refer to the set of intervals represented by a spanning interval as its *extension*. Moreover, a set of spanning intervals will represent the union of the extensions of its members. Additionally, the empty set of spanning intervals will represent the empty set of intervals. I further refer to the set of intervals represented by a set of spanning intervals as its *extension*. A key result of this paper is that if the set of all intervals over which some set of primitive event types hold can be represented as finite sets of spanning intervals then the set of all intervals over which all event types that are expressible as compound event-logic expressions over those primitives hold can also be represented as finite sets of spanning intervals.

While we require that all intervals have finite endpoints, for reasons that will also become apparent in Section 4.4, it is necessary to allow spanning intervals to have infinite endpoints, for example $[[-\infty,j],[k,l]]$. Such spanning intervals with infinite endpoints represent sets of intervals with finite endpoints but where the range of possible endpoints is unconstrained from above or below.

### 4.2 Normalizing Spanning Intervals

Just as we desire that the extension of every interval have a unique representation, we also desire that the extension of every spanning interval have a unique representation. There are a number of situations where two different spanning intervals will have the same extension. First, all spanning intervals ${}_{\alpha}[{}_{\gamma}[i,j]_{\delta},{}_{\epsilon}[k,l]_{\zeta}]_{\beta}$ where $i=\infty$, $j=-\infty$, $k=\infty$, or $l=-\infty$ represent the empty set of intervals, because there are no intervals with an endpoint that is less than or equal to minus infinity or greater than or equal to infinity. Second, if $i=-\infty$, $j=\infty$, $k=-\infty$, or $l=\infty$, the value of $\gamma$, $\delta$, $\epsilon$, or $\zeta$ does not affect the denotation respectively, because there are no intervals with infinite endpoints. Third, if $j>l$, $j$ can be decreased as far as $l$ without changing the denotation, because all intervals where the upper endpoint is less than the lower endpoint equivalently denote the empty interval. Similarly, if $k<i$, $k$ can be increased as far as $i$ without changing the denotation. Fourth, all spanning intervals where $i > j$ or $k > l$ represent the empty set of intervals, because the range of possible endpoints would be empty. Fifth, all spanning intervals where $i = j$ and either $\gamma$ or $\delta$ is false (indicating an open range for the lower endpoint) represent the empty set of intervals, because the range of possible endpoints would be empty. Similarly, all spanning intervals where $k = l$ and either $\epsilon$ or $\zeta$ is false (indicating an open range for the upper endpoint) also represent the empty set of intervals. Sixth, all spanning intervals where $i = l$ and either $\alpha$ or $\beta$ is false (indicating an open interval) also represent the empty set of intervals, because the endpoints of an open interval must be different. Seventh, if $j = l$ and $\zeta$ is false, the value of $\delta$ does not affect the denotation, because if $j = l$ and $\zeta$ is false, the upper endpoint must be less than $l$ and the lower endpoint must be less than or equal to $j$ which equals $l$, so the lower endpoint must be less than $j$. Similarly, if $k = i$ and $\gamma$ is false, the value of $\epsilon$ does not affect the denotation. Eighth, if $j = l$ and either $\alpha$ or $\beta$ is false, the value of $\delta$ does not affect the denotation, because the lower endpoint of an open interval must be less than its upper endpoint. Similarly, if $k = i$ and either $\alpha$ or $\beta$ is false, the value of $\epsilon$ does not affect the denotation.

To create a situation where the extension of every spanning interval has a unique representation, let us represent all empty sets of intervals as $\{\}$. When the values of $i$, $j$, $k$, $l$, $\alpha$, $\beta$, $\gamma$, $\delta$, $\epsilon$, or $\zeta$ can be changed without changing the denotation, we will select the tightest such values. In other words, false values for the Boolean parameters, maximal values for the lower bounds, and minimal values for the upper bounds. Thus whenever we represent a spanning interval ${}_\alpha[_\gamma[i,j]_\delta,_\epsilon[k,l]_\zeta]_\beta$ explicitly, it will have a nonempty extension and will satisfy the following normalization criterion:

$$
\begin{array}{ll}
(1) & (i \neq \infty) \wedge (j \neq -\infty) \wedge (k \neq \infty) \wedge (l \neq -\infty) \wedge \\
(2) & [(i = -\infty) \rightarrow \neg\gamma] \wedge [(j = \infty) \rightarrow \neg\delta] \wedge [(k = -\infty) \rightarrow \neg\epsilon] \wedge [(l = \infty) \rightarrow \neg\zeta] \wedge \\
(3) & (j \leq l) \wedge (k \geq i) \wedge \\
(4) & (i \leq j) \wedge (k \leq l) \wedge \\
(5) & [(i \neq j) \vee (\gamma \wedge \delta)] \wedge [(k \neq l) \vee (\epsilon \wedge \zeta)] \wedge \\
(6) & [(i \neq l) \vee (\alpha \wedge \beta)] \wedge \\
(7) & \{[(j = l) \wedge \neg\zeta] \rightarrow \neg\delta\} \wedge \{[(k = i) \wedge \neg\gamma] \rightarrow \neg\epsilon\} \wedge \\
(8) & \{[(j = l) \wedge (\neg\alpha \vee \neg\beta)] \rightarrow \neg\delta\} \wedge \{[(k = i) \wedge (\neg\alpha \vee \neg\beta)] \rightarrow \neg\epsilon\}
\end{array}
$$

Criteria (1) through (8) correspond to points one through eight above.

A spanning interval ${}_\alpha[_\gamma[i,j]_\delta,_\epsilon[k,l]_\zeta]_\beta$ is *normalized* if $i$, $j$, $k$, $l$, $\alpha$, $\beta$, $\gamma$, $\delta$, $\epsilon$, and $\zeta$ cannot be changed without changing its denotation. Given a (potentially non-normalized) spanning interval $\mathbf{i}$, its *normalization* $\langle\mathbf{i}\rangle$ is the smallest set of normalized spanning intervals that represents the extension of $\mathbf{i}$. One can compute $\langle\mathbf{i}\rangle$ as follows:

$$\langle {}_{\alpha}[{}_{\gamma}[i,j]_{\delta},{}_{\epsilon}[k,l]_{\zeta}]_{\beta}\rangle \triangleq \begin{cases} \{{}_{\alpha}[{}_{\gamma'}[i,j']_{\delta'},{}_{\epsilon'}[k',l]_{\zeta'}]_{\beta}\} \\ \quad \text{where} \begin{array}{rcl} j' & = & \min(j,l) \\ k' & = & \max(k,i) \\ \gamma' & = & \gamma \wedge (i \neq -\infty) \\ \delta' & = & \delta \wedge [\min(j,l) \neq \infty] \wedge [(j < l) \vee (\zeta \wedge \alpha \wedge \beta)] \\ \epsilon' & = & \epsilon \wedge [\max(k,i) \neq -\infty] \wedge [(k > i) \vee (\gamma \wedge \beta \wedge \alpha)] \\ \zeta' & = & \zeta \wedge (l \neq \infty) \end{array} \\ \quad \text{when} \; (i \leq j') \wedge (k' \leq l) \wedge \\ \qquad [(i = j') \rightarrow (\gamma' \wedge \delta')] \wedge [(k' = l) \rightarrow (\epsilon' \wedge \zeta')] \wedge \\ \qquad [(i = l) \rightarrow (\alpha \wedge \beta)] \wedge \\ \qquad (i \neq \infty) \wedge (j' \neq -\infty) \wedge (k' \neq \infty) \wedge (l \neq -\infty) \\ \{\} \quad \text{otherwise} \end{cases}$$

An important property of spanning intervals is that for any spanning interval $\mathbf{i}$, $\langle\mathbf{i}\rangle$ contains at most *one* normalized spanning interval.[6]

### 4.3 Computing the Intersection of Two Normalized Spanning Intervals

Given two normalized spanning intervals $\mathbf{i}_1$ and $\mathbf{i}_2$, their *intersection* $\mathbf{i}_1 \cap \mathbf{i}_2$ is a set of normalized spanning intervals whose extension is the intersection of the extensions of $\mathbf{i}_1$ and $\mathbf{i}_2$. One can compute $\mathbf{i}_1 \cap \mathbf{i}_2$ as follows:

$$\begin{array}{l} {}_{\alpha_1}[{}_{\gamma_1}[i_1,j_1]_{\delta_1},{}_{\epsilon_1}[k_1,l_1]_{\zeta_1}]_{\beta_1} \cap {}_{\alpha_2}[{}_{\gamma_2}[i_2,j_2]_{\delta_2},{}_{\epsilon_2}[k_2,l_2]_{\zeta_2}]_{\beta_2} \triangleq \\ \begin{cases} \langle {}_{\alpha_1}[{}_{\gamma}[\max(i_1,i_2),\min(j_1,j_2)]_{\delta},{}_{\epsilon}[\max(k_1,k_2),\min(l_1,l_2)]_{\zeta}]_{\beta_1}\rangle \\ \quad \text{where} \begin{array}{rcl} \gamma & = & \begin{cases} \gamma_1 & i_1 > i_2 \\ \gamma_1 \wedge \gamma_2 & i_1 = i_2 \\ \gamma_2 & i_1 < i_2 \end{cases} \\ \delta & = & \begin{cases} \delta_1 & j_1 < j_2 \\ \delta_1 \wedge \delta_2 & j_1 = j_2 \\ \delta_2 & j_1 > j_2 \end{cases} \\ \epsilon & = & \begin{cases} \epsilon_1 & k_1 > k_2 \\ \epsilon_1 \wedge \epsilon_2 & k_1 = k_2 \\ \epsilon_2 & k_1 < k_2 \end{cases} \\ \zeta & = & \begin{cases} \zeta_1 & l_1 < l_2 \\ \zeta_1 \wedge \zeta_2 & l_1 = l_2 \\ \zeta_2 & l_1 > l_2 \end{cases} \end{array} \\ \quad \text{when} \; (\alpha_1 = \alpha_2) \wedge (\beta_1 = \beta_2) \\ \{\} \quad \text{otherwise} \end{cases} \end{array}$$

6. The reason that $\langle\mathbf{i}\rangle$ contains *at most* one normalized spanning interval and not *exactly* one normalized spanning interval is that $\mathbf{i}$ may denote the empty set of intervals. For example, normalizing the (non-normalized) spanning interval $[[10,10],[1,1]]$ yields the empty set. Many of the definitions in the coming sections compute sets of normalized spanning intervals as unions of one or more applications of the normalization operator $\langle\cdot\rangle$. Each such application might yield either the empty set or a set containing a single normalized spanning interval. This leads to upper, but not lower, bounds on the size of the computed unions.

An important property of normalized spanning intervals is that for any two normalized spanning intervals $\mathbf{i}_1$ and $\mathbf{i}_2$, $\mathbf{i}_1 \cap \mathbf{i}_2$ contains at most *one* normalized spanning interval.

The intuition behind the above definition is as follows. All of the intervals in the extension of a spanning interval are of the same type, namely $[q,r]$, $(q,r]$, $[q,r)$, or $(q,r)$. The intersection of two spanning intervals has a nonempty extension only if the two spanning intervals contain the same type of intervals in their extension. If they do, and the sets contain intervals whose lower endpoint is bound from below by $i_1$ and $i_2$ respectively, then the intersection will contain intervals whose lower endpoint is bound from below by both $i_1$ and $i_2$. The resulting bound is open or closed depending on which of the input bounds is tighter. Similarly for the upper bound on the lower endpoint and the lower and upper bounds on the upper endpoint.

### 4.4 Computing the Complement of a Normalized Spanning Interval

Given a normalized spanning intervals $\mathbf{i}$, its *complement* $\neg\mathbf{i}$ is a set of normalized spanning intervals whose extension is the complement of the extension of $\mathbf{i}$. One can compute $\neg\mathbf{i}$ as follows:

$$\neg{}_{\alpha}[_{\gamma}[i,j]_{\delta},_{\epsilon}[k,l]_{\zeta}]_{\beta} \triangleq \left(\begin{array}{l} \langle_{\alpha}[_{\mathbf{T}}[-\infty,\infty]_{\mathbf{T}},_{\mathbf{T}}[-\infty,k]_{\neg\epsilon}]_{\beta}\rangle\cup \\ \langle_{\alpha}[_{\mathbf{T}}[-\infty,\infty]_{\mathbf{T}},_{\neg\zeta}[l,\infty]_{\mathbf{T}}]_{\beta}\rangle\cup \\ \langle_{\alpha}[_{\mathbf{T}}[-\infty,i]_{\neg\gamma},_{\mathbf{T}}[-\infty,\infty]_{\mathbf{T}}]_{\beta}\rangle\cup \\ \langle_{\alpha}[_{\neg\delta}[j,\infty]_{\mathbf{T}},_{\mathbf{T}}[-\infty,\infty]_{\mathbf{T}}]_{\beta}\rangle\cup \\ \langle_{\neg\alpha}[_{\mathbf{T}}[-\infty,\infty]_{\mathbf{T}},_{\mathbf{T}}[-\infty,\infty]_{\mathbf{T}}]_{\beta}\rangle\cup \\ \langle_{\alpha}[_{\mathbf{T}}[-\infty,\infty]_{\mathbf{T}},_{\mathbf{T}}[-\infty,\infty]_{\mathbf{T}}]_{\neg\beta}\rangle\cup \\ \langle_{\neg\alpha}[_{\mathbf{T}}[-\infty,\infty]_{\mathbf{T}},_{\mathbf{T}}[-\infty,\infty]_{\mathbf{T}}]_{\neg\beta}\rangle \end{array}\right)$$

An important property of normalized spanning intervals is that for any normalized spanning interval $\mathbf{i}$, $\neg\mathbf{i}$ contains at most *seven* normalized spanning intervals.

The intuition behind the above definition is as follows. First note that the negation of $q \leq_{\alpha} r$ is $q \geq_{\neg\alpha} r$. Next note that the extension of $\mathbf{i}$ contains intervals whose endpoints $q$ and $r$ satisfy $(q \geq_{\gamma} i) \wedge (q \leq_{\delta} j) \wedge (r \geq_{\epsilon} k) \wedge (r \leq_{\zeta} l)$. Thus the extension of $\neg\mathbf{i}$ contains intervals whose endpoints satisfy the negation of this, namely $(q \leq_{\neg\gamma} i) \vee (q \geq_{\neg\delta} j) \vee (r \leq_{\neg\epsilon} k) \vee (r \geq_{\neg\zeta} l)$. Such a disjunction requires four spanning intervals, the first four in the above definition. Additionally, if the extension of $\mathbf{i}$ contains intervals of the form $[q,r]$, the extension of $\neg\mathbf{i}$ will contain all intervals not of the form $[q,r]$, namely $(q,r]$, $[q,r)$, and $(q,r)$. Similarly for the cases where the extension of $\mathbf{i}$ contains intervals of the form $(q,r]$, $[q,r)$, or $(q,r)$. This accounts for the last three spanning intervals in the above definition.

We now see why it is necessary to allow spanning intervals to have open ranges of endpoint values as well as infinite endpoints. The complement of a spanning interval, such as $[[i,j],[k,l]]$, with closed endpoint ranges and finite endpoints includes spanning intervals, such as $[[-\infty,i),[-\infty,\infty]]$, with open endpoint ranges and infinite endpoints.

### 4.5 Computing the Span of two Normalized Spanning Intervals

The *span* of two intervals $\mathbf{i}_1$ and $\mathbf{i}_2$, denoted $\text{SPAN}(\mathbf{i}_1,\mathbf{i}_2)$, is the smallest interval whose extension contains the extensions of both $\mathbf{i}_1$ and $\mathbf{i}_2$. For example, the span of $(1,4)$ and $[2,6]$ is $(1,6]$. Similarly, the span of $[3,7)$ and $(3,7]$ is $[3,7]$. More generally, the lower endpoint of $\text{SPAN}(\mathbf{i}_1,\mathbf{i}_2)$ is the minimum of the lower endpoints of $\mathbf{i}_1$ and $\mathbf{i}_2$. The lower endpoint of $\text{SPAN}(\mathbf{i}_1,\mathbf{i}_2)$ is open or closed depending on whether the smaller of the lower endpoints of $\mathbf{i}_1$ and $\mathbf{i}_2$ is open or closed. Analogously, the upper endpoint of $\text{SPAN}(\mathbf{i}_1, \mathbf{i}_2)$ is the maximum of the upper endpoints of $\mathbf{i}_1$ and $\mathbf{i}_2$. The upper endpoint of $\text{SPAN}(\mathbf{i}_1, \mathbf{i}_2)$ is open or closed depending on whether the larger of the upper endpoints of $\mathbf{i}_1$ and $\mathbf{i}_2$ is open or closed. More precisely, $\text{SPAN}(\mathbf{i}_1, \mathbf{i}_2)$ can be computed as follows:

$$\begin{array}{l}\text{SPAN}({}_{\alpha_1}[q_1, r_1]_{\beta_1}, {}_{\alpha_2}[q_2, r_2]_{\beta_2}) \stackrel{\triangle}{=} \\ \quad {}_{\{[\alpha_1 \wedge (q_1 \le q_2)] \vee [\alpha_2 \wedge (q_1 \ge q_2)]\}}[\min(q_1, q_2), \max(r_1, r_2)]_{\{[\beta_1 \wedge (r_1 \ge r_2)] \vee [\beta_2 \wedge (r_1 \le r_2)]\}}\end{array}$$

The notion of span will be used in Section 4.7.

Let us extend the notion of span to two sets of intervals by the following definition:

$$\text{SPAN}(I_1, I_2) \stackrel{\triangle}{=} \bigcup_{\mathbf{i}_1 \in I_1} \bigcup_{\mathbf{i}_2 \in I_2} \text{SPAN}(\mathbf{i}_1, \mathbf{i}_2)$$

We will want to compute the span of two sets of intervals $I_1$ and $I_2$, when both $I_1$ and $I_2$ are represented as spanning intervals. Additionally, we will want the resulting span to be represented as a small set of spanning intervals.

Given two normalized spanning intervals $\mathbf{i}_1$ and $\mathbf{i}_2$, their *span* $\text{SPAN}(\mathbf{i}_1, \mathbf{i}_2)$ is a set of normalized spanning intervals whose extension is the span of the extensions of $\mathbf{i}_1$ and $\mathbf{i}_2$. One can compute $\text{SPAN}(\mathbf{i}_1, \mathbf{i}_2)$ as follows:

$$\begin{array}{l}\text{SPAN}({}_{\alpha_1}[{}_{\gamma_1}[i_1, j_1]_{\delta_1}, {}_{\epsilon_1}[k_1, l_1]_{\zeta_1}]_{\beta_1}, {}_{\alpha_2}[{}_{\gamma_2}[i_2, j_2]_{\delta_2}, {}_{\epsilon_2}[k_2, l_2]_{\zeta_2}]_{\beta_2}) \stackrel{\triangle}{=} \\ \quad \left(\begin{array}{l}\langle {}_{\alpha_1}[{}_{\gamma_1}[i_1, j]_{\delta}, {}_{\epsilon}[k, l_1]_{\zeta_1}]_{\beta_1} \rangle \cup \\ \langle {}_{\alpha_1}[{}_{\gamma_1}[i_1, j]_{\delta}, {}_{\epsilon}[k, l_2]_{\zeta_2}]_{\beta_2} \rangle \cup \\ \langle {}_{\alpha_2}[{}_{\gamma_2}[i_2, j]_{\delta}, {}_{\epsilon}[k, l_1]_{\zeta_1}]_{\beta_1} \rangle \cup \\ \langle {}_{\alpha_2}[{}_{\gamma_2}[i_2, j]_{\delta}, {}_{\epsilon}[k, l_2]_{\zeta_2}]_{\beta_2} \rangle \end{array}\right) \\ \quad \begin{array}{rrcl} \text{where} & j & = & \min(j_1, j_2) \\ & k & = & \max(k_1, k_2) \\ & \delta & = & [\delta_1 \wedge (j_1 \le j_2)] \vee [\delta_2 \wedge (j_1 \ge j_2)] \\ & \epsilon & = & [\epsilon_1 \wedge (k_1 \ge k_2)] \vee [\epsilon_2 \wedge (k_1 \le k_2)] \end{array}\end{array}$$

An important property of normalized spanning intervals is that for any two normalized spanning intervals $\mathbf{i}_1$ and $\mathbf{i}_2$, $\text{SPAN}(\mathbf{i}_1, \mathbf{i}_2)$ contains at most *four* normalized spanning intervals. In practice, however, fewer normalized spanning intervals are needed, often only one.

The intuition behind the above definition is as follows. Consider, first, the lower endpoint. Suppose that the lower endpoints $q_1$ and $q_2$ of $\mathbf{i}_1$ and $\mathbf{i}_2$ are in ${}_{\gamma_1}[i_1, j_1]_{\delta_1}$ and ${}_{\gamma_2}[i_2, j_2]_{\delta_2}$ respectively. That means that $i_1 \le_{\gamma_1} q_1 \le_{\delta_1} j_1$ and $i_2 \le_{\gamma_2} q_2 \le_{\delta_2} j_2$. The lower endpoint of $\text{SPAN}(\mathbf{i}_1, \mathbf{i}_2)$ will be $q_1$, when $q_1 \le q_2$, and $q_2$, when $q_1 \ge q_2$. Thus it will be $q_1$, for all $i_1 \le_{\gamma_1} q_1 \le_{\delta} \min(j_1, j_2)$, and will be $q_2$, for all $i_2 \le_{\gamma_2} q_2 \le_{\delta} \min(j_1, j_2)$, where $\delta = \delta_1$, when $j_1 \le j_2$, and $\delta = \delta_2$, when $j_1 \ge j_2$. Thus there will be two potential ranges for the lower endpoint of $\text{SPAN}(\mathbf{i}_1, \mathbf{i}_2)$: ${}_{\gamma_1}[i_1, \min(j_1, j_2)]_{\delta}$ and ${}_{\gamma_2}[i_2, \min(j_1, j_2)]_{\delta}$. When the lower endpoint of $\text{SPAN}(\mathbf{i}_1, \mathbf{i}_2)$ is taken from the former, it will be open or closed depending on whether the lower endpoint of $\mathbf{i}_1$ is open or closed. When it is taken from the later, it will be open or closed depending on whether the lower endpoint of $\mathbf{i}_2$ is open or closed. Thus the lower endpoint of $\text{SPAN}(\mathbf{i}_1, \mathbf{i}_2)$ can be either ${}_{\alpha_1}[{}_{\gamma_1}[i_1, \min(j_1, j_2)]_{\delta}$ or ${}_{\alpha_1}[{}_{\gamma_2}[i_2, \min(j_1, j_2)]_{\delta}$. Analogous reasoning can be applied to the upper endpoints. If the upper endpoints of $\mathbf{i}_1$ and $\mathbf{i}_2$ are ${}_{\epsilon_1}[k_1, l_1]_{\zeta_1}]_{\beta_1}$ and ${}_{\epsilon_2}[k_2, l_2]_{\zeta_2}]_{\beta_2}$ respectively, then there are two possibilities for the upper endpoint of $\text{SPAN}(\mathbf{i}_1, \mathbf{i}_2)$, namely ${}_{\epsilon}[\max(k_1, k_2), l_1]_{\zeta_1}]_{\beta_1}$ and ${}_{\epsilon}[\max(k_1, k_2), l_2]_{\zeta_2}]_{\beta_2}$, where $\epsilon = \epsilon_1$, when $k_1 \ge k_2$, and $\epsilon = \epsilon_2$, when $k_1 \le k_2$.

### 4.6 Computing the $\mathcal{D}$ of a Normalized Spanning Interval

Given an Allen relation $r$ and a set $I$ of intervals, let $\mathcal{D}(r, I)$ denote the set $J$ of all intervals $\mathbf{j}$ such that $\mathbf{i}r\mathbf{j}$ for some $\mathbf{i} \in I$. Given an Allen relation $r$ and a normalized spanning interval $\mathbf{i}$, let $\mathcal{D}(r, \mathbf{i})$ denote a set of normalized spanning intervals whose extension is $\mathcal{D}(r, I)$, where $I$ is the extension of $\mathbf{i}$. One can compute $\mathcal{D}(r, \mathbf{i})$ as follows:

$$\mathcal{D}(=, \mathbf{i}) \triangleq \{\mathbf{i}\}$$

$$\mathcal{D}(<, {}_{\alpha_1}[_{\gamma_1}[i_1, j_1]_{\delta_1}, {}_{\epsilon_1}[k_1, l_1]_{\zeta_1}]_{\beta_1}) \triangleq \bigcup_{\alpha_2, \beta_2 \in \{\mathbf{T}, \mathbf{F}\}} \langle {}_{\alpha_2}[_{(\neg\beta_1 \wedge \neg\alpha_2 \wedge \epsilon_1)}[k_1, \infty]_{\mathbf{T}}, {}_{\mathbf{T}}[-\infty, \infty]_{\mathbf{T}}]_{\beta_2} \rangle$$

$$\mathcal{D}(>, {}_{\alpha_1}[_{\gamma_1}[i_1, j_1]_{\delta_1}, {}_{\epsilon_1}[k_1, l_1]_{\zeta_1}]_{\beta_1}) \triangleq \bigcup_{\alpha_2, \beta_2 \in \{\mathbf{T}, \mathbf{F}\}} \langle {}_{\alpha_2}[_{\mathbf{T}}[-\infty, \infty]_{\mathbf{T}}, {}_{\mathbf{T}}[-\infty, j_1]_{(\neg\alpha_1 \wedge \neg\beta_2 \wedge \delta_1)}]_{\beta_2} \rangle$$

$$\mathcal{D}(\mathsf{m}, {}_{\alpha_1}[_{\gamma_1}[i_1, j_1]_{\delta_1}, {}_{\epsilon_1}[k_1, l_1]_{\zeta_1}]_{\beta_1}) \triangleq \bigcup_{\beta_2 \in \{\mathbf{T}, \mathbf{F}\}} \langle {}_{\neg\beta_1}[_{\epsilon_1}[k_1, l_1]_{\zeta_1}, {}_{\mathbf{T}}[-\infty, \infty]_{\mathbf{T}}]_{\beta_2} \rangle$$

$$\mathcal{D}(\mathsf{mi}, {}_{\alpha_1}[_{\gamma_1}[i_1, j_1]_{\delta_1}, {}_{\epsilon_1}[k_1, l_1]_{\zeta_1}]_{\beta_1}) \triangleq \bigcup_{\alpha_2 \in \{\mathbf{T}, \mathbf{F}\}} \langle {}_{\alpha_2}[_{\mathbf{T}}[-\infty, \infty]_{\mathbf{T}}, {}_{\gamma_1}[i_1, j_1]_{\delta_1}]_{\neg\alpha_1} \rangle$$

$$\mathcal{D}(\mathsf{o}, {}_{\alpha_1}[_{\gamma_1}[i_1, j_1]_{\delta_1}, {}_{\epsilon_1}[k_1, l_1]_{\zeta_1}]_{\beta_1}) \triangleq \bigcup_{\alpha_2, \beta_2 \in \{\mathbf{T}, \mathbf{F}\}} \langle {}_{\alpha_2}[_{(\alpha_1 \wedge \neg\alpha_2 \wedge \gamma_1)}[i_1, l_1]_{(\beta_1 \wedge \alpha_2 \wedge \zeta_1)}, {}_{(\neg\beta_1 \wedge \beta_2 \wedge \epsilon_1)}[k_1, \infty]_{\mathbf{T}}]_{\beta_2} \rangle$$

$$\mathcal{D}(\mathsf{oi}, {}_{\alpha_1}[_{\gamma_1}[i_1, j_1]_{\delta_1}, {}_{\epsilon_1}[k_1, l_1]_{\zeta_1}]_{\beta_1}) \triangleq \bigcup_{\alpha_2, \beta_2 \in \{\mathbf{T}, \mathbf{F}\}} \langle {}_{\alpha_2}[_{\mathbf{T}}[-\infty, j_1]_{(\neg\alpha_1 \wedge \alpha_2 \wedge \delta_1)}, {}_{(\alpha_1 \wedge \beta_2 \wedge \gamma_1)}[i_1, l_1]_{(\beta_1 \wedge \neg\beta_2 \wedge \zeta_1)}]_{\beta_2} \rangle$$

$$\mathcal{D}(\mathsf{s}, {}_{\alpha_1}[_{\gamma_1}[i_1, j_1]_{\delta_1}, {}_{\epsilon_1}[k_1, l_1]_{\zeta_1}]_{\beta_1}) \triangleq \bigcup_{\beta_2 \in \{\mathbf{T}, \mathbf{F}\}} \langle {}_{\alpha_1}[_{\gamma_1}[i_1, j_1]_{\delta_1}, {}_{(\neg\beta_1 \wedge \beta_2 \wedge \epsilon_1)}[k_1, \infty]_{\mathbf{T}}]_{\beta_2} \rangle$$

$$\mathcal{D}(\mathsf{si}, {}_{\alpha_1}[_{\gamma_1}[i_1, j_1]_{\delta_1}, {}_{\epsilon_1}[k_1, l_1]_{\zeta_1}]_{\beta_1}) \triangleq \bigcup_{\beta_2 \in \{\mathbf{T}, \mathbf{F}\}} \langle {}_{\alpha_1}[_{\gamma_1}[i_1, j_1]_{\delta_1}, {}_{\mathbf{T}}[-\infty, l_1]_{(\beta_1 \wedge \neg\beta_2 \wedge \zeta_1)}]_{\beta_2} \rangle$$

$$\mathcal{D}(\mathsf{f}, {}_{\alpha_1}[_{\gamma_1}[i_1, j_1]_{\delta_1}, {}_{\epsilon_1}[k_1, l_1]_{\zeta_1}]_{\beta_1}) \triangleq \bigcup_{\alpha_2 \in \{\mathbf{T}, \mathbf{F}\}} \langle {}_{\alpha_2}[_{\mathbf{T}}[-\infty, j_1]_{(\neg\alpha_1 \wedge \alpha_2 \wedge \delta_1)}, {}_{\epsilon_1}[k_1, l_1]_{\zeta_1}]_{\beta_1} \rangle$$

$$\mathcal{D}(\mathsf{fi}, {}_{\alpha_1}[_{\gamma_1}[i_1, j_1]_{\delta_1}, {}_{\epsilon_1}[k_1, l_1]_{\zeta_1}]_{\beta_1}) \triangleq \bigcup_{\alpha_2 \in \{\mathbf{T}, \mathbf{F}\}} \langle {}_{\alpha_2}[_{(\alpha_1 \wedge \neg\alpha_2 \wedge \gamma_1)}[i_1, \infty]_{\mathbf{T}}, {}_{\epsilon_1}[k_1, l_1]_{\zeta_1}]_{\beta_1} \rangle$$

$$\mathcal{D}(\mathsf{d}, {}_{\alpha_1}[_{\gamma_1}[i_1, j_1]_{\delta_1}, {}_{\epsilon_1}[k_1, l_1]_{\zeta_1}]_{\beta_1}) \triangleq \bigcup_{\alpha_2, \beta_2 \in \{\mathbf{T}, \mathbf{F}\}} \langle {}_{\alpha_2}[_{\mathbf{T}}[-\infty, j_1]_{(\neg\alpha_1 \wedge \alpha_2 \wedge \delta_1)}, {}_{(\neg\beta_1 \wedge \beta_2 \wedge \epsilon_1)}[k_1, \infty]_{\mathbf{T}}]_{\beta_2} \rangle$$

$$\mathcal{D}(\mathsf{di}, {}_{\alpha_1}[_{\gamma_1}[i_1, j_1]_{\delta_1}, {}_{\epsilon_1}[k_1, l_1]_{\zeta_1}]_{\beta_1}) \triangleq \bigcup_{\alpha_2, \beta_2 \in \{\mathbf{T}, \mathbf{F}\}} \langle {}_{\alpha_2}[_{(\alpha_1 \wedge \neg\alpha_2 \wedge \gamma_1)}[i_1, \infty]_{\mathbf{T}}, {}_{\mathbf{T}}[-\infty, l_1]_{(\beta_1 \wedge \neg\beta_2 \wedge \zeta_1)}]_{\beta_2} \rangle$$

An important property of normalized spanning intervals is that for any normalized spanning interval $\mathbf{i}$, $\mathcal{D}(r, \mathbf{i})$ contains at most 1, 4, 4, 2, 2, 4, 4, 2, 2, 2, 2, 4, or 4 normalized spanning intervals when $r$ is $=$, $<$, $>$, m, mi, o, oi, s, si, f, fi, d, or di respectively. In practice, however, fewer normalized spanning intervals are needed, often only one.

The intuition behind the above definition is as follows. Let us handle each of the cases separately.

$r =<$ For any intervals $\mathbf{i}'_1$ and $\mathbf{i}'_2$ in the extensions of $\mathbf{i}_1$ and $\mathbf{i}_2$ respectively we want $\mathbf{i}'_1 < \mathbf{i}'_2$. From (2) we get $r_1 \leq_{(\neg\beta_1\wedge\neg\alpha_2)} q_2$. Furthermore, from (14) we get $k_1 \leq_{\epsilon_1} r_1$. Combining these we get $k_1 \leq_{(\neg\beta_1\wedge\neg\alpha_2\wedge\epsilon_1)} q_2$. In this case, both $\alpha_2$ and $\beta_2$ are free indicating that either endpoint of $\mathbf{i}'_2$ can be open or closed.

$r =>$ For any intervals $\mathbf{i}'_1$ and $\mathbf{i}'_2$ in the extensions of $\mathbf{i}_1$ and $\mathbf{i}_2$ respectively we want $\mathbf{i}'_1 > \mathbf{i}'_2$. From (3) we get $q_1 \geq_{(\neg\alpha_1\wedge\neg\beta_2)} r_2$. Furthermore, from (14) we get $q_1 \leq_{\delta_1} j_1$. Combining these we get $r_2 \leq_{(\neg\alpha_1\wedge\neg\beta_2\wedge\delta_1)} j_1$. In this case, both $\alpha_2$ and $\beta_2$ are free indicating that either endpoint of $\mathbf{i}'_2$ can be open or closed.

$r = \mathsf{m}$ For any intervals $\mathbf{i}'_1$ and $\mathbf{i}'_2$ in the extensions of $\mathbf{i}_1$ and $\mathbf{i}_2$ respectively we want $\mathbf{i}'_1$ m $\mathbf{i}'_2$. From (4) we get $r_1 = q_2$ and $\beta_1 \neq \alpha_2$. Furthermore, from (14) we get $k_1 \leq_{\epsilon_1} r_1 \leq_{\zeta_1} l_1$. Combining these we get $k_1 \leq_{\epsilon_1} q_2 \leq_{\zeta_1} l_1$ and $\beta_1 \neq \alpha_2$. In this case, only $\beta_2$ is free indicating that the upper endpoint of $\mathbf{i}'_2$ can be open or closed.

$r = \mathsf{mi}$ For any intervals $\mathbf{i}'_1$ and $\mathbf{i}'_2$ in the extensions of $\mathbf{i}_1$ and $\mathbf{i}_2$ respectively we want $\mathbf{i}'_1$ mi $\mathbf{i}'_2$. From (5) we get $q_1 = r_2$ and $\alpha_1 \neq \beta_2$. Furthermore, from (14) we get $i_1 \leq_{\gamma_1} q_1 \leq_{\delta_1} j_1$. Combining these we get $i_1 \leq_{\gamma_1} r_2 \leq_{\delta_1} j_1$ and $\alpha_1 \neq \beta_2$. In this case, only $\alpha_2$ is free indicating that the lower endpoint of $\mathbf{i}'_2$ can be open or closed.

$r = \mathsf{o}$ For any intervals $\mathbf{i}'_1$ and $\mathbf{i}'_2$ in the extensions of $\mathbf{i}_1$ and $\mathbf{i}_2$ respectively we want $\mathbf{i}'_1$ o $\mathbf{i}'_2$. From (6) we get $q_1 \leq_{(\alpha_1\wedge\neg\alpha_2)} q_2 \leq_{(\beta_1\wedge\alpha_2)} r_1 \leq_{(\neg\beta_1\wedge\beta_2)} r_2$. Furthermore, from (14) we get $i_1 \leq_{\gamma_1} q_1$ and $k_1 \leq_{\epsilon_1} r_1 \leq_{\zeta_1} l_1$. Combining these we get $i_1 \leq_{(\alpha_1\wedge\neg\alpha_2\wedge\gamma_1)} q_2 \leq_{(\beta_1\wedge\alpha_2\wedge\zeta_1)} l_1$ and $k_1 \leq_{(\neg\beta_1\wedge\beta_2\wedge\epsilon_1)} r_2$. In this case, both $\alpha_2$ and $\beta_2$ are free indicating that either endpoint of $\mathbf{i}'_2$ can be open or closed.

$r = \mathsf{oi}$ For any intervals $\mathbf{i}'_1$ and $\mathbf{i}'_2$ in the extensions of $\mathbf{i}_1$ and $\mathbf{i}_2$ respectively we want $\mathbf{i}'_1$ oi $\mathbf{i}'_2$. From (7) we get $q_2 \leq_{(\neg\alpha_1\wedge\alpha_2)} q_1 \leq_{(\alpha_1\wedge\beta_2)} r_2 \leq_{(\beta_1\wedge\neg\beta_2)} r_1$. Furthermore, from (14) we get $r_1 \leq_{\zeta_1} l_1$ and $i_1 \leq_{\gamma_1} q_1 \leq_{\delta_1} j_1$. Combining these we get $i_1 \leq_{(\alpha_1\wedge\beta_2\wedge\gamma_1)} r_2 \leq_{(\beta_1\wedge\neg\beta_2\wedge\zeta_1)} l_1$ and $q_2 \leq_{(\neg\alpha_1\wedge\alpha_2\wedge\delta_1)} j_1$. In this case, both $\alpha_2$ and $\beta_2$ are free indicating that either endpoint of $\mathbf{i}'_2$ can be open or closed.

$r = \mathsf{s}$ For any intervals $\mathbf{i}'_1$ and $\mathbf{i}'_2$ in the extensions of $\mathbf{i}_1$ and $\mathbf{i}_2$ respectively we want $\mathbf{i}'_1$ s $\mathbf{i}'_2$. From (8) we get $q_1 = q_2$, $\alpha_1 = \alpha_2$, and $r_1 \leq_{(\neg\beta_1\wedge\beta_2)} r_2$. Furthermore, from (14) we get $i_1 \leq_{\gamma_1} q_1 \leq_{\delta_1} j_1$ and $k_1 \leq_{\epsilon_1} r_1$. Combining these we get $\alpha_1 = \alpha_2$, $i_1 \leq_{\gamma_1} q_2 \leq_{\delta_1} j_1$, and $k_1 \leq_{(\neg\beta_1\wedge\beta_2\wedge\epsilon_1)} r_2$. In this case, only $\beta_2$ is free indicating that the upper endpoint of $\mathbf{i}'_2$ can be open or closed.

$r = \mathsf{si}$ For any intervals $\mathbf{i}'_1$ and $\mathbf{i}'_2$ in the extensions of $\mathbf{i}_1$ and $\mathbf{i}_2$ respectively we want $\mathbf{i}'_1$ si $\mathbf{i}'_2$. From (9) we get $q_1 = q_2$, $\alpha_1 = \alpha_2$, and $r_1 \geq_{(\beta_1\wedge\neg\beta_2)} r_2$. Furthermore, from (14) we get $i_1 \leq_{\gamma_1} q_1 \leq_{\delta_1} j_1$ and $r_1 \leq_{\zeta_1} l_1$. Combining these we get $\alpha_1 = \alpha_2$, $i_1 \leq_{\gamma_1} q_2 \leq_{\delta_1} j_1$, and $r_2 \leq_{(\beta_1\wedge\neg\beta_2\wedge\zeta_1)} l_1$. In this case, only $\beta_2$ is free indicating that the upper endpoint of $\mathbf{i}'_2$ can be open or closed.

$r = \mathsf{f}$ For any intervals $\mathbf{i}_1'$ and $\mathbf{i}_2'$ in the extensions of $\mathbf{i}_1$ and $\mathbf{i}_2$ respectively we want $\mathbf{i}_1'\ \mathsf{f}\ \mathbf{i}_2'$. From (10) we get $q_1 \geq_{(\neg\alpha_1\wedge\alpha_2)} q_2$, $r_1 = r_2$, and $\beta_1 = \beta_2$. Furthermore, from (14) we get $k_1 \leq_{\epsilon_1} r_1 \leq_{\zeta_1} l_1$ and $q_1 \leq_{\delta_1} j_1$. Combining these we get $\beta_1 = \beta_2$, $k_1 \leq_{\epsilon_1} r_2 \leq_{\zeta_1} l_1$, and $q_2 \leq_{(\neg\alpha_1\wedge\alpha_2\wedge\delta_1)} j_1$. In this case, only $\alpha_2$ is free indicating that the lower endpoint of $\mathbf{i}_2'$ can be open or closed.

$r = \mathsf{fi}$ For any intervals $\mathbf{i}_1'$ and $\mathbf{i}_2'$ in the extensions of $\mathbf{i}_1$ and $\mathbf{i}_2$ respectively we want $\mathbf{i}_1'\ \mathsf{fi}\ \mathbf{i}_2'$. From (11) we get $q_1 \geq_{(\alpha_1\wedge\neg\alpha_2)} q_2$, $r_1 = r_2$, and $\beta_1 = \beta_2$. Furthermore, from (14) we get $k_1 \leq_{\epsilon_1} r_1 \leq_{\zeta_1} l_1$ and $i_1 \leq_{\gamma_1} q_1$. Combining these we get $\beta_1 = \beta_2$, $k_1 \leq_{\epsilon_1} r_2 \leq_{\zeta_1} l_1$, and $i_1 \leq_{(\alpha_1\wedge\neg\alpha_2\wedge\gamma_1)} q_2$. In this case, only $\alpha_2$ is free indicating that the lower endpoint of $\mathbf{i}_2'$ can be open or closed.

$r = \mathsf{d}$ For any intervals $\mathbf{i}_1'$ and $\mathbf{i}_2'$ in the extensions of $\mathbf{i}_1$ and $\mathbf{i}_2$ respectively we want $\mathbf{i}_1'\ \mathsf{d}\ \mathbf{i}_2'$. From (12) we get $q_1 \geq_{(\neg\alpha_1\wedge\alpha_2)} q_2$ and $r_1 \leq_{(\neg\beta_1\wedge\beta_2)} r_2$. Furthermore, from (14) we get $q_1 \leq_{\delta_1} j_1$ and $k_1 \leq_{\epsilon_1} r_1$. Combining these we get $q_2 \leq_{(\neg\alpha_1\wedge\alpha_2\wedge\delta_1)} j_1$ and $k_1 \leq_{(\neg\beta_1\wedge\beta_2\wedge\epsilon_1)} r_2$. In this case, both $\alpha_2$ and $\beta_2$ are free indicating that either endpoint of $\mathbf{i}_2'$ can be open or closed.

$r = \mathsf{di}$ For any intervals $\mathbf{i}_1'$ and $\mathbf{i}_2'$ in the extensions of $\mathbf{i}_1$ and $\mathbf{i}_2$ respectively we want $\mathbf{i}_1'\ \mathsf{di}\ \mathbf{i}_2'$. From (13) we get $q_1 \leq_{(\alpha_1\wedge\neg\alpha_2)} q_2$ and $r_1 \geq_{(\beta_1\wedge\neg\beta_2)} r_2$. Furthermore, from (14) we get $i_1 \leq_{\gamma_1} q_1$ and $r_1 \leq_{\zeta_1} l_1$. Combining these we get $i_1 \leq_{(\alpha_1\wedge\neg\alpha_2\wedge\gamma_1)} q_2$ and $r_2 \leq_{(\beta_1\wedge\neg\beta_2\wedge\zeta_1)} l_1$. In this case, both $\alpha_2$ and $\beta_2$ are free indicating that either endpoint of $\mathbf{i}_2'$ can be open or closed.

### 4.7 Computing the $\mathcal{I}$ of two Normalized Spanning Intervals

Given an Allen relation $r$ and two sets $I$ and $J$ of intervals, let $\mathcal{I}(I, r, J)$ denote the set $K$ of all intervals $\mathbf{k}$ such that $\mathbf{k} = \text{SPAN}(\mathbf{i}, \mathbf{j})$ for some $\mathbf{i} \in I$ and $\mathbf{j} \in J$, where $\mathbf{i}r\mathbf{j}$. Given an Allen relation $r$ and two normalized spanning intervals $\mathbf{i}$ and $\mathbf{j}$, let $\mathcal{I}(\mathbf{i}, r, \mathbf{j})$ denote a set of normalized spanning intervals whose extension is $\mathcal{I}(I, r, J)$, where $I$ and $J$ are the extensions of $\mathbf{i}$ and $\mathbf{j}$ respectively. One can compute $\mathcal{I}(\mathbf{i}, r, \mathbf{j})$ as follows:

$$\mathcal{I}(\mathbf{i}, r, \mathbf{j}) \triangleq \bigcup_{\mathbf{i}' \in \mathcal{D}(r^{-1}, \mathbf{j})} \bigcup_{\mathbf{i}'' \in \mathbf{i}' \cap \mathbf{i}} \bigcup_{\mathbf{j}' \in \mathcal{D}(r, \mathbf{i})} \bigcup_{\mathbf{j}'' \in \mathbf{j}' \cap \mathbf{j}} \text{SPAN}(\mathbf{i}'', \mathbf{j}'')$$

Here, $r^{-1}$ denotes the inverse relation corresponding to $r$, i.e. the same relation as $r$ but with the arguments reversed. It is easy to see that $|\mathcal{I}(\cdot, r, \cdot)| \leq 4|\mathcal{D}(r, \cdot)|^2$. Thus an important property of normalized spanning intervals is that for any two normalized spanning intervals $\mathbf{i}$ and $\mathbf{j}$, $\mathcal{I}(\mathbf{i}, r, \mathbf{j})$ contains at most 4, 64, 64, 16, 16, 64, 64, 16, 16, 16, 16, 64, or 64 normalized spanning intervals, when $r$ is $=$, $<$, $>$, $\mathsf{m}$, $\mathsf{mi}$, $\mathsf{o}$, $\mathsf{oi}$, $\mathsf{s}$, $\mathsf{si}$, $\mathsf{f}$, $\mathsf{fi}$, $\mathsf{d}$, or $\mathsf{di}$ respectively. While simple combinatorial enumeration yields the above weak bounds on the number of normalized spanning intervals needed to represent $\mathcal{I}(\mathbf{i}, r, \mathbf{j})$, in practice, far fewer normalized spanning intervals are needed, in most cases only one.

The intuition behind the above definition is as follows. Let $I$ and $J$ be the extensions of $\mathbf{i}$ and $\mathbf{j}$ respectively. The extension of the set of all $\mathbf{i}'$ is the set of all intervals $\mathbf{i}$ such that $\mathbf{i}r\mathbf{j}$ for some $\mathbf{j}$ in $J$. Furthermore, the extension of the set of all $\mathbf{i}''$ is the set of all intervals $\mathbf{i}$ in $I$ such that $\mathbf{i}r\mathbf{j}$ for some $\mathbf{j}$ in $J$. Similarly, the extension of the set of all $\mathbf{j}'$ is the set of all intervals $\mathbf{j}$ such that $\mathbf{i}r\mathbf{j}$ for some $\mathbf{i}$ in $I$. Analogously, the extension of the set of all $\mathbf{j}''$ is the set of all intervals $\mathbf{j}$ in $J$ such that $\mathbf{i}r\mathbf{j}$ for some $\mathbf{i}$ in $I$. Thus the extension of the set of all $\text{SPAN}(\mathbf{i}'', \mathbf{j}'')$ is the set of all intervals $\mathbf{k}$ such that $\mathbf{k} = \text{SPAN}(\mathbf{i}, \mathbf{j})$ where $\mathbf{i}$ is in $I$, $\mathbf{j}$ is in $J$, and $\mathbf{i}r\mathbf{j}$.

### 4.8 An Efficient Inference Procedure for Event Logic

Given the above procedures for computing $\langle \mathbf{i} \rangle$, $\mathbf{i}_1 \cap \mathbf{i}_2$, $\neg\mathbf{i}$, SPAN$(\mathbf{i}_1, \mathbf{i}_2)$, $\mathcal{D}(r, \mathbf{i})$, and $\mathcal{I}(\mathbf{i}, r, \mathbf{j})$, one can now define a procedure for computing $\mathcal{E}(M, \Phi)$. This procedure takes a model $M$ along with an event-logic expression $\Phi$ and computes a set of normalized spanning intervals that represents the set $I$ of intervals $\mathbf{i}$ for which $\Phi @ \mathbf{i}$ is true. The model $M$ is a set of atomic event-occurrence formulae of the form $p(c_1, \ldots, c_n) @ \mathbf{i}$, where $p(c_1, \ldots, c_n)$ is a ground primitive event-logic expression and $\mathbf{i}$ is a normalized spanning interval. A model entry $p(c_1, \ldots, c_n) @ \mathbf{i}$ indicates that the primitive event $p(c_1, \ldots, c_n)$ occurred during all intervals in the extension of $\mathbf{i}$.

$$
\begin{array}{rcl}
\mathcal{E}(M, p(c_1, \ldots, c_n)) & \triangleq & \{\mathbf{i} | p(c_1, \ldots, c_n) @ \mathbf{i} \in M\} \\
\mathcal{E}(M, \Phi \vee \Psi) & \triangleq & \mathcal{E}(M, \Phi) \cup \mathcal{E}(M, \Psi) \\
\mathcal{E}(M, \neg\Phi) & \triangleq & \displaystyle\bigcup_{\mathbf{i}'_1 \in \neg\mathbf{i}_1} \cdots \bigcup_{\mathbf{i}'_n \in \neg\mathbf{i}_n} \mathbf{i}'_1 \cap \cdots \cap \mathbf{i}'_n \\
& & \text{where } \mathcal{E}(M, \Phi) = \{\mathbf{i}_1, \ldots, \mathbf{i}_n\} \\
\mathcal{E}(M, \Phi \wedge_R \Psi) & \triangleq & \displaystyle\bigcup_{\mathbf{i} \in \mathcal{E}(M, \Phi)} \bigcup_{\mathbf{j} \in \mathcal{E}(M, \Psi)} \bigcup_{r \in R} \mathcal{I}(\mathbf{i}, r, \mathbf{j}) \\
\mathcal{E}(M, \Diamond_R \Phi) & \triangleq & \displaystyle\bigcup_{\mathbf{i} \in \mathcal{E}(M, \Phi)} \bigcup_{r \in R} \mathcal{D}(r, \mathbf{i})
\end{array}
$$

The procedure performs structural induction on $\Phi$. It computes a set of normalized spanning intervals to the represent the occurrence of each atomic event-logic expression in $\Phi$ and recursively combines the sets so computed for each child subexpression to yield the sets for each parent subexpression. An important property of this inference procedure is that for any finite model $M$, $\mathcal{E}(M, \Phi)$, the set $I$ of intervals $\mathbf{i}$ for which $\Phi @ \mathbf{i}$ is true, can be represented by a finite set of normalized spanning intervals. Nominally, the number of normalized spanning intervals in $\mathcal{E}(M, \Phi)$ can be exponential in the subexpression depth of $\Phi$ because each step in the structural induction can introduce a constant factor growth in the size of the set. However, in practice, such exponential growth does not occur. Computing $\mathcal{E}(M, \Phi)$ for all of the event types given in Figure 10 for all of the movies that have been tried so far have yielded sets of fewer than a dozen normalized spanning intervals.

## 5. Experimental Results

The techniques described in this paper have been implemented as a system called LEONARD and tested on a number of video sequences.[7] LEONARD successfully recognizes the events *pick up*, *put down*, *stack*, *unstack*, *move*, *assemble*, and *disassemble* using the definitions given in Figure 10. Figures 1 and 11 through 15 show the key frames from movies that depict these seven event types. These movies were filmed using a Canon VC-C3 camera and a Matrox Meteor frame grabber at $320 \times 240$ resolution at 30fps. Figures 4 and 16 through 20 show the results of segmentation, tracking, and model reconstruction for those key frames superimposed on the original images. Figures 5 and 21 through 25 show the results of event classification for these movies. These figures show LEONARD correctly recognizing the intended event classes for each movie.

7. The code for LEONARD, the video input sequences discussed in this paper, and the full frame-by-frame output of LEONARD on those sequences is available as Online Appendix 1, as well as from `ftp://ftp.nj.nec.com/pub/qobi/leonard.tar.Z`.

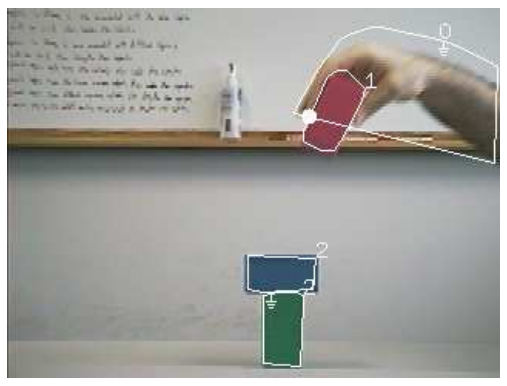
Frame 3

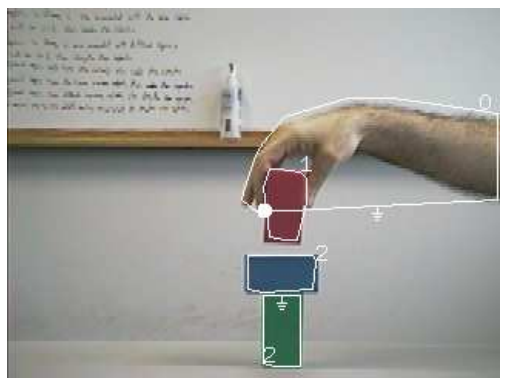
Frame 11

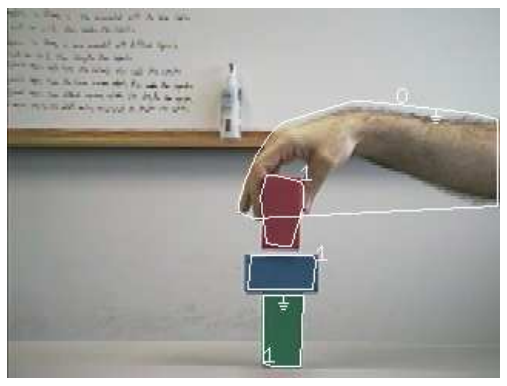
Frame 12

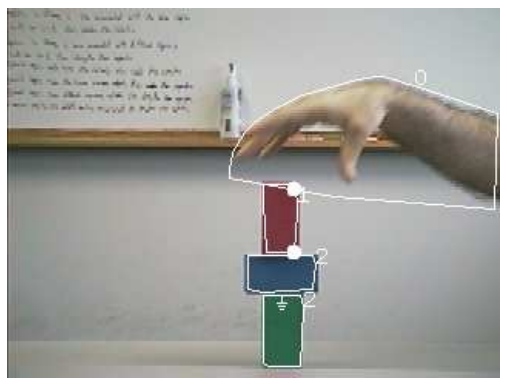
Frame 23

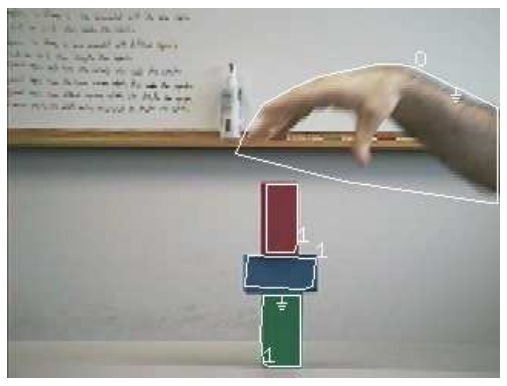
Frame 24

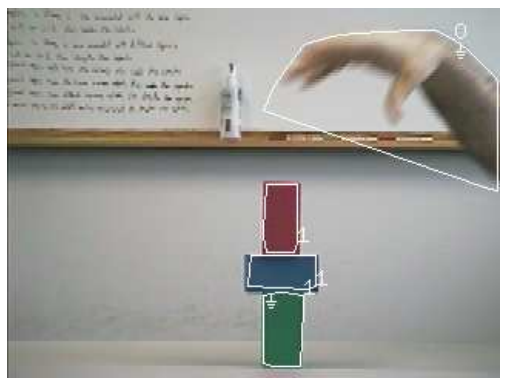
Frame 26

Figure 16: The output of the segmentation-and-tracking and model-reconstruction components applied to the image sequence from Figure 11, an image sequence that depicts a *stack* event.

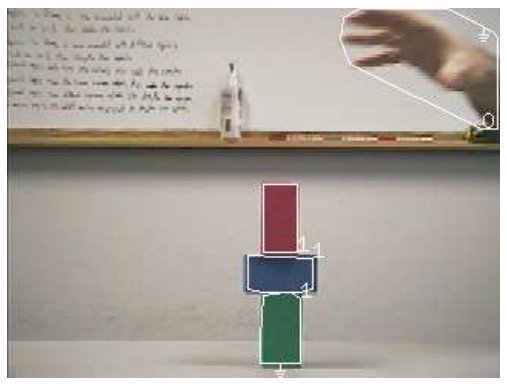
Frame 1

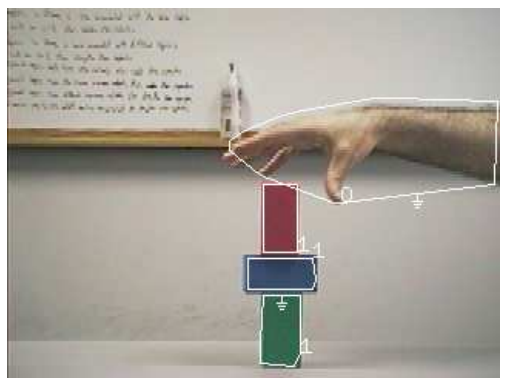
Frame 10

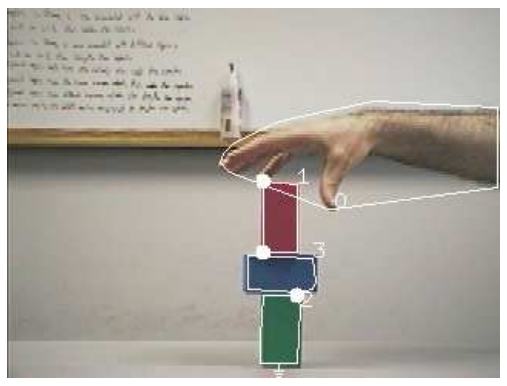
Frame 11

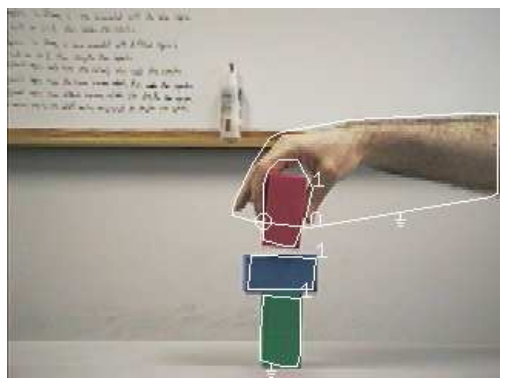
Frame 24

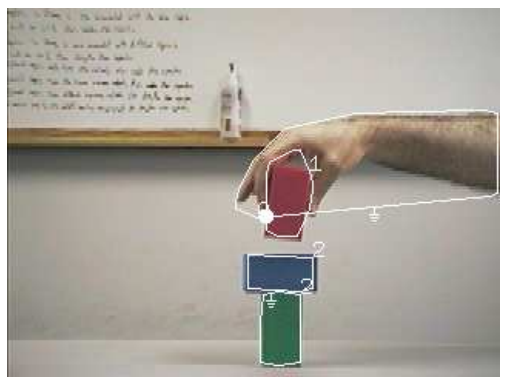
Frame 25

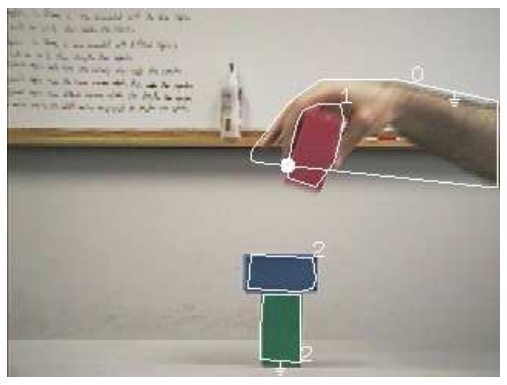
Frame 29

Figure 17: The output of the segmentation-and-tracking and model-reconstruction components applied to the image sequence from Figure 12, an image sequence that depicts an *unstack* event.

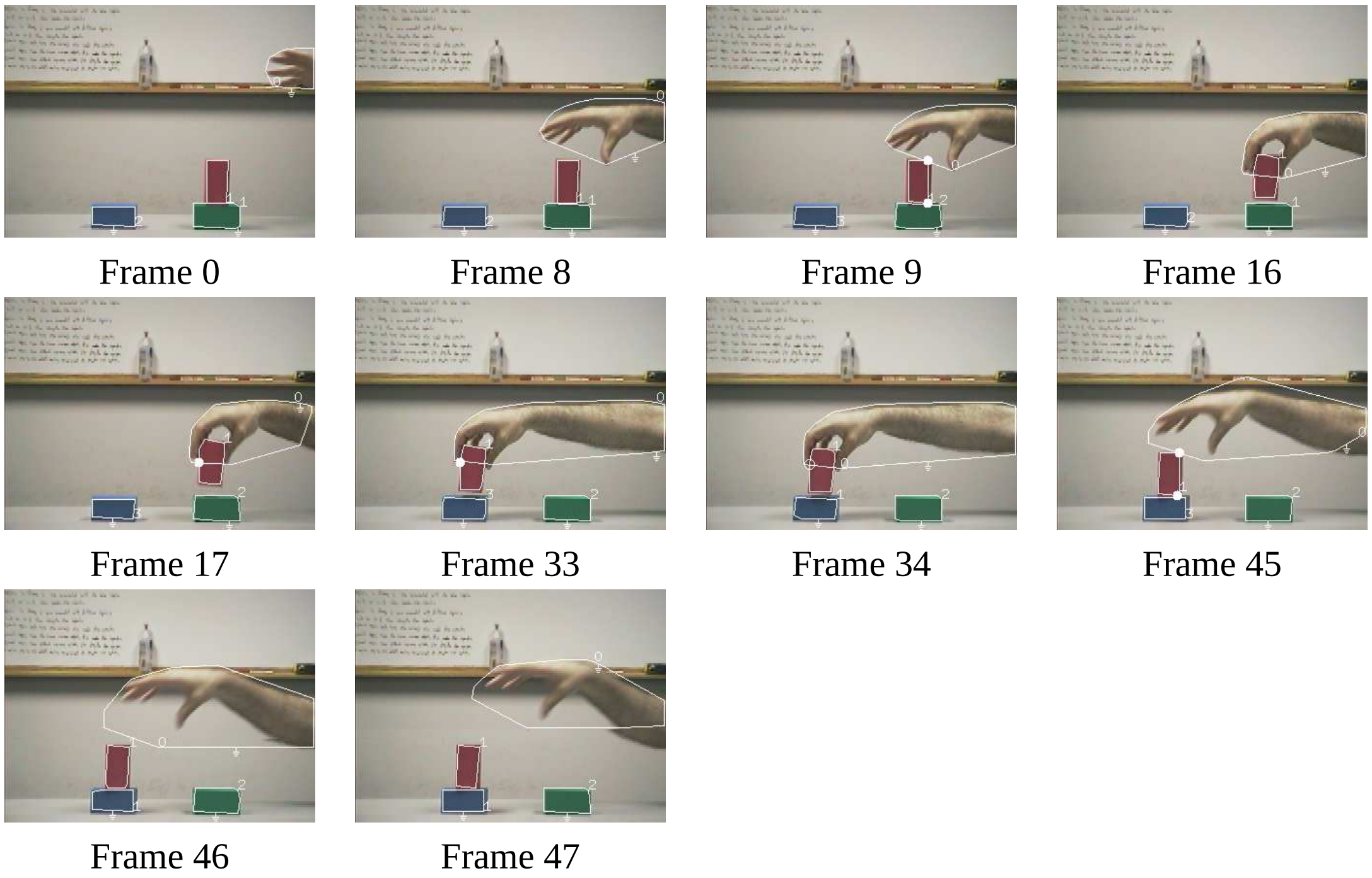

Figure 18: The output of the segmentation-and-tracking and model-reconstruction components applied to the image sequence from Figure 13, an image sequence that depicts a *move* event.

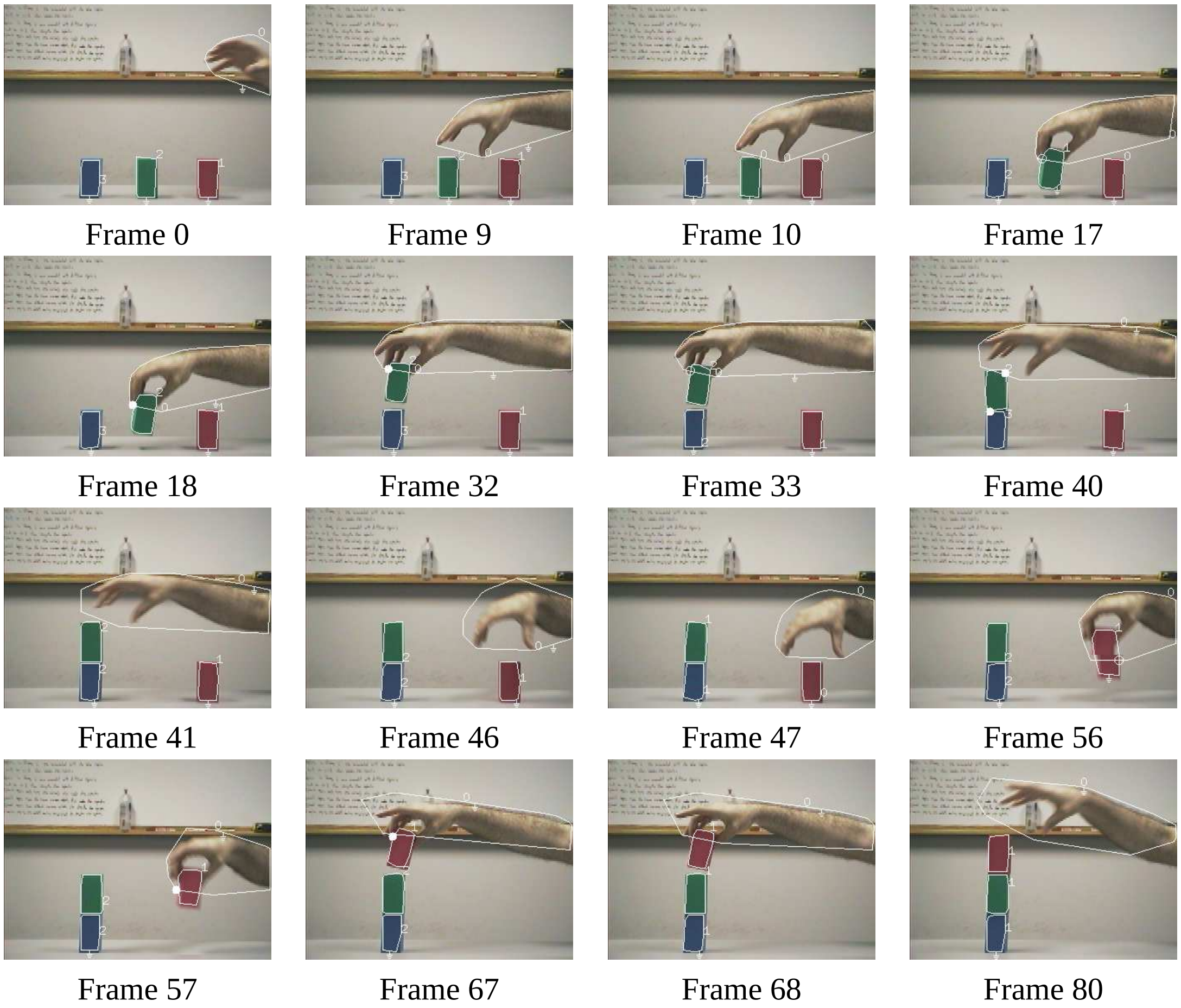

Figure 19: The output of the segmentation-and-tracking and model-reconstruction components applied to the image sequence from Figure 14, an image sequence that depicts an *assemble* event.

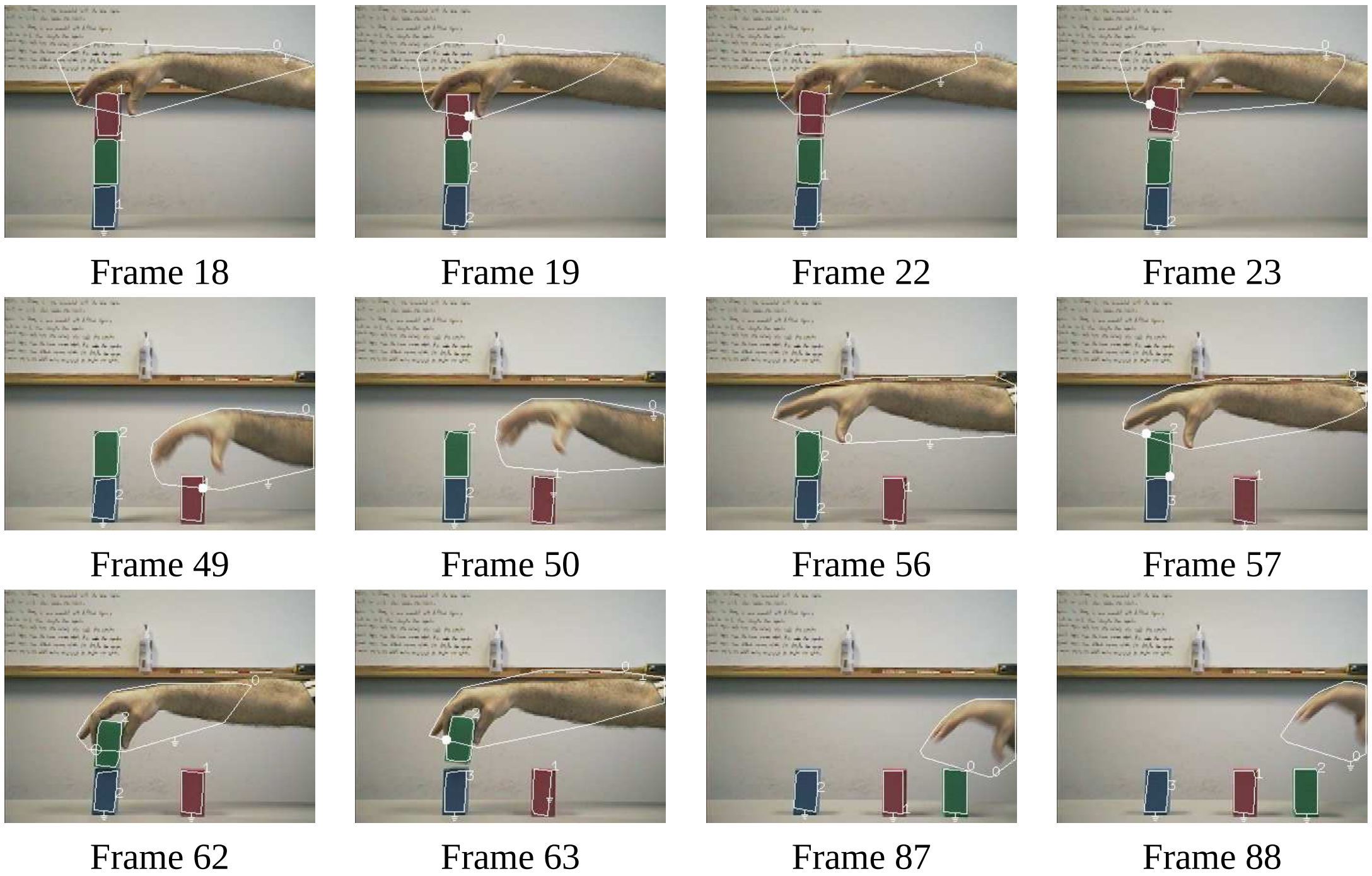

Figure 20: The output of the segmentation-and-tracking and model-reconstruction components applied to the image sequence from Figure 15, an image sequence that depicts a *disassemble* event.

```
(PUT-DOWN MOVING RED BLUE)@{[[0,12],[24,30])}
(STACK MOVING RED BLUE GREEN)@{[[0,12],[24,30])}

(SUPPORTED? MOVING)@{[[13:24])}
(SUPPORTED? RED)@{[[0:30])}
(SUPPORTED? BLUE)@{[[0:30])}
(SUPPORTS? MOVING RED)@{[[0:12])}
(SUPPORTS? RED MOVING)@{[[13:24])}
(SUPPORTS? RED BLUE)@{[[19:20]), [[21:22])}
(SUPPORTS? GREEN MOVING)@{[[19:20]), [[21:22])}
(SUPPORTS? GREEN RED)@{[[19:20]), [[21:22])}
(SUPPORTS? GREEN BLUE)@{[[0:30])}
(SUPPORTS? BLUE MOVING)@{[[13:24])}
(SUPPORTS? BLUE RED)@{[[12:30])}
(CONTACTS? RED BLUE)@{[[12:19]), [[20:21]), [[22:30])}
(CONTACTS? GREEN BLUE)@{[[0:30])}
(ATTACHED? MOVING RED)@{[[0:24])}
(ATTACHED? RED BLUE)@{[[19:20]), [[21:22])}
```

Figure 21: The output of the event-classification component applied to the model sequence from Figure 16. Note that the *stack* event is correctly recognized, as well as the constituent *put down* event.

```
(PICK-UP MOVING RED BLUE)@{[[0,11],[25,33])}
(UNSTACK MOVING RED BLUE GREEN)@{[[0,11],[25,33])}

(SUPPORTED? MOVING)@{[[11:23])}
(SUPPORTED? RED)@{[[0:36])}
(SUPPORTED? BLUE)@{[[0:36])}
(SUPPORTS? MOVING RED)@{[[23:36])}
(SUPPORTS? RED MOVING)@{[[11:23])}
(SUPPORTS? RED BLUE)@{[[13:14])}
(SUPPORTS? GREEN MOVING)@{[[13:14])}
(SUPPORTS? GREEN RED)@{[[13:14])}
(SUPPORTS? GREEN BLUE)@{[[0:36])}
(SUPPORTS? BLUE MOVING)@{[[11:23])}
(SUPPORTS? BLUE RED)@{[[0:25])}
(CONTACTS? MOVING RED)@{[[34:36])}
(CONTACTS? RED BLUE)@{[[0:13]), [[14:24])}
(CONTACTS? GREEN BLUE)@{[[0:13]), [[14:36])}
(ATTACHED? MOVING RED)@{[[11:33])}
(ATTACHED? RED BLUE)@{[[13:14])}
(ATTACHED? GREEN BLUE)@{[[13:14])}
```

Figure 22: The output of the event-classification component applied to the model sequence from Figure 17. Note that the *unstack* event is correctly recognized, as well as the constituent *pick up* event.

```
(PICK-UP MOVING RED GREEN)@{[[0,9],[17,46])}
(PUT-DOWN MOVING RED BLUE)@{[[17,35],[46,52])}
(MOVE MOVING RED GREEN BLUE)@{[[0,9],[46,52])}

(SUPPORTED? MOVING)@{[[9:15])}
(SUPPORTED? RED)@{[[0:52])}
(SUPPORTED? BLUE)@{[[35:46])}
(SUPPORTS? MOVING RED)@{[[17:46])}
(SUPPORTS? MOVING BLUE)@{[[35:46])}
(SUPPORTS? RED MOVING)@{[[9:15])}
(SUPPORTS? RED BLUE)@{[[35:46])}
(SUPPORTS? GREEN MOVING)@{[[9:15])}
(SUPPORTS? GREEN RED)@{[[0:17])}
(SUPPORTS? BLUE RED)@{[[46:52])}
(CONTACTS? RED GREEN)@{[[0:17])}
(CONTACTS? RED BLUE)@{[[46:52])}
(ATTACHED? MOVING RED)@{[[9:46])}
(ATTACHED? RED BLUE)@{[[35:46])}
```

Figure 23: The output of the event-classification component applied to the model sequence from Figure 18. Note that the *move* event is correctly recognized, as well as the constituent *pick up* and *put down* subevents.

```
(PUT-DOWN MOVING RED GREEN)@{[[57,68],[68,87])}
(PUT-DOWN MOVING GREEN BLUE)@{[[18,35],[41,47])}
(STACK MOVING RED GREEN BLUE)@{[[57,68],[68,87])}
(ASSEMBLE MOVING RED GREEN BLUE)@{[[18,35],[68,87])}

(SUPPORTED? MOVING)@{[[10:18]), [[47:57])}
(SUPPORTED? RED)@{[[57:87])}
(SUPPORTED? GREEN)@{[[11:87])}
(SUPPORTED? BLUE)@{[[35:41])}
(SUPPORTS? MOVING RED)@{[[57:68])}
(SUPPORTS? MOVING GREEN)@{[[11:41])}
(SUPPORTS? MOVING BLUE)@{[[35:41])}
(SUPPORTS? RED MOVING)@{[[10:18]), [[47:57])}
(SUPPORTS? RED GREEN)@{[[11:16])}
(SUPPORTS? GREEN RED)@{[[68:87])}
(SUPPORTS? GREEN BLUE)@{[[35:41])}
(SUPPORTS? BLUE GREEN)@{[[41:87])}
(CONTACTS? RED GREEN)@{[[68:87])}
(CONTACTS? GREEN BLUE)@{[[41:87])}
(ATTACHED? MOVING RED)@{[[11:16]), [[49:68])}
(ATTACHED? MOVING GREEN)@{[[11:41])}
(ATTACHED? GREEN BLUE)@{[[35:41])}
```

Figure 24: The output of the event-classification component applied to the model sequence from Figure 19. Note that the *assemble* event is correctly recognized, as well as the constituent *put down* and *stack* subevents.

```
(PICK-UP MOVING RED GREEN)@{[[0,19],[23,50])}
(PICK-UP MOVING GREEN BLUE)@{[[22,58],[62,87])}
(UNSTACK MOVING RED GREEN BLUE)@{[[0,19],[23,50])}
(DISASSEMBLE MOVING RED GREEN BLUE)@{[[0,19],[62,87])}

(SUPPORTED? MOVING)@{[[19:22])}
(SUPPORTED? RED)@{[[0:50])}
(SUPPORTED? GREEN)@{[[0:87])}
(SUPPORTED? BLUE)@{[[58:62])}
(SUPPORTS? MOVING RED)@{[[23:50])}
(SUPPORTS? MOVING GREEN)@{[[58:87])}
(SUPPORTS? MOVING BLUE)@{[[58:62])}
(SUPPORTS? RED MOVING)@{[[19:22])}
(SUPPORTS? GREEN MOVING)@{[[19:22])}
(SUPPORTS? GREEN RED)@{[[0:23])}
(SUPPORTS? GREEN BLUE)@{[[58:62])}
(SUPPORTS? BLUE GREEN)@{[[0:58])}
(CONTACTS? RED GREEN)@{[[0:23])}
(CONTACTS? GREEN BLUE)@{[[0:58])}
(ATTACHED? MOVING RED)@{[[19:50])}
(ATTACHED? MOVING GREEN)@{[[58:87])}
(ATTACHED? GREEN BLUE)@{[[58:62])}
```

Figure 25: The output of the event-classification component applied to the model sequence from Figure 20. Note that the *disassemble* event is correctly recognized, as well as the constituent *pick up* and *unstack* subevents.

In Figure 4(a), Frames 0 through 1 correspond to the first subevent of a *pick up* event, Frames 2 through 13 correspond to the second subevent, and Frames 14 through 22 correspond to the third subevent. In Figure 4(b), Frames 0 through 13 correspond to the first subevent of a *put down* event, Frames 14 through 22 correspond to the second subevent, and Frames 23 through 32 correspond to the third subevent. LEONARD correctly recognizes these as instances of *pick up* and *put down* respectively. In Figure 16, Frames 0 through 11, 12 through 23, and 24 through 30 correspond to the three subevents of a *put down* event. LEONARD correctly recognizes this as a *put down* event and also as a *stack* event. In Figure 17, Frames 0 through 10, 11 through 24, and 25 through 33 correspond to the three subevents of a *pick up* event. LEONARD correctly recognizes this as a *pick up* event and also as an *unstack* event. In Figure 18, Frames 0 through 8, 9 through 16, and 17 through 45 correspond to the three subevents of a *pick up* event and Frames 17 through 33, 34 through 45, and 46 through 52 correspond to the three subevents of a *put down* event. LEONARD correctly recognizes the combination of these two events as a *move* event. In Figure 19, Frames 18 through 32, 33 through 40, and 41 through 46 correspond to the three subevents of a *put down* event and Frames 57 through 67 and 68 through 87 correspond to the first and third subevents of a second *put down* event, with the second subevent being empty. The second *put down* event is also correctly recognized as a *stack* event and the combination of these two events is correctly recognized as an *assemble* event. In Figure 20, Frames 0 through 18, 19 through 22, and 23 through 50 correspond to the three subevents of a *pick up* event and Frames 23 through 56, 57 through 62, and 63 through 87 correspond to the three subevents of a second *pick up* event. The first *pick up* event is also correctly recognized as an *unstack* event and the combination of these two events is correctly recognized as a *disassemble* event. These examples show that LEONARD correctly recognizes each of the seven event types with no false positives.

As discussed in the introduction, using force dynamics and event logic to recognize events offers several advantages over the prior approach of using motion profile and hidden Markov models.

- robustness against variance in motion profile
- robustness against presence of extraneous objects in the field of view
- ability to perform temporal and spatial segmentation of events
- ability to detect non-occurrence of events

Figures 26 through 35 illustrate these advantages. Figure 26 shows a *pick up* event from the left in contrast to Figure 4(a) which is from the right. Even though these have different motion profiles, Figure 31 shows that LEONARD correctly recognizes that these exhibit the same sequence of changes in force-dynamic relations and constitute the same event type, namely *pick up*. Figure 27 shows a *pick up* event with two extraneous blocks in the field of view. Figure 32 shows that LEONARD correctly recognizes that these extraneous blocks do not participate in any events and, despite their presence, the truth conditions for a *pick up* event still hold between the other objects. Figure 28 shows a *pick up* event, followed by a *put down* event, followed by another *pick up* event, followed by another *put down* event. Figure 33 shows that LEONARD correctly recognizes this sequence of four event occurrences. Figure 29 shows two simultaneous *pick up* events. Figure 34 shows that LEONARD correctly recognizes these two simultaneous event occurrences. Finally, Figure 30 shows two non-events. Figure 35 shows that LEONARD is not fooled into thinking that these constitute *pick up* or *put down* events, even though portions of these events have similar motion

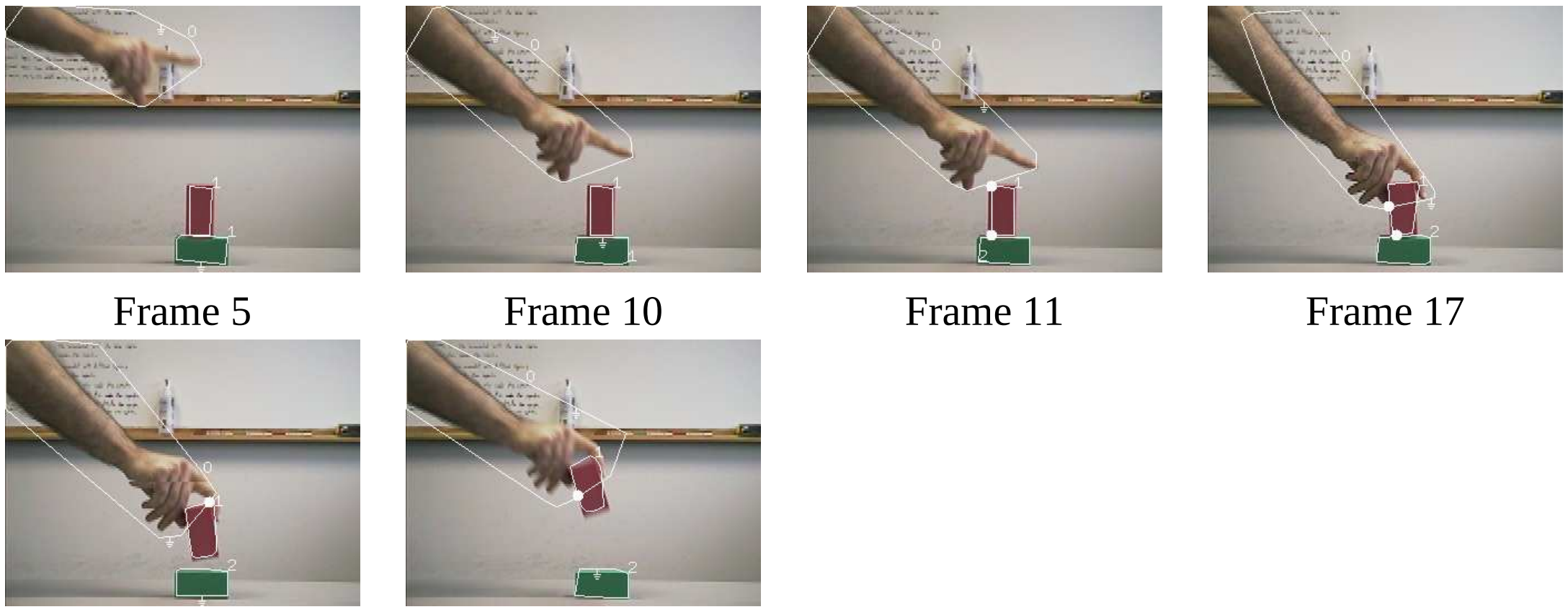

Frame 5 Frame 10 Frame 11 Frame 17

Frame 18 Frame 22

Figure 26: The output of the segmentation-and-tracking and model-reconstruction components on an image sequence depicting a *pick up* event from the left instead of from the right.

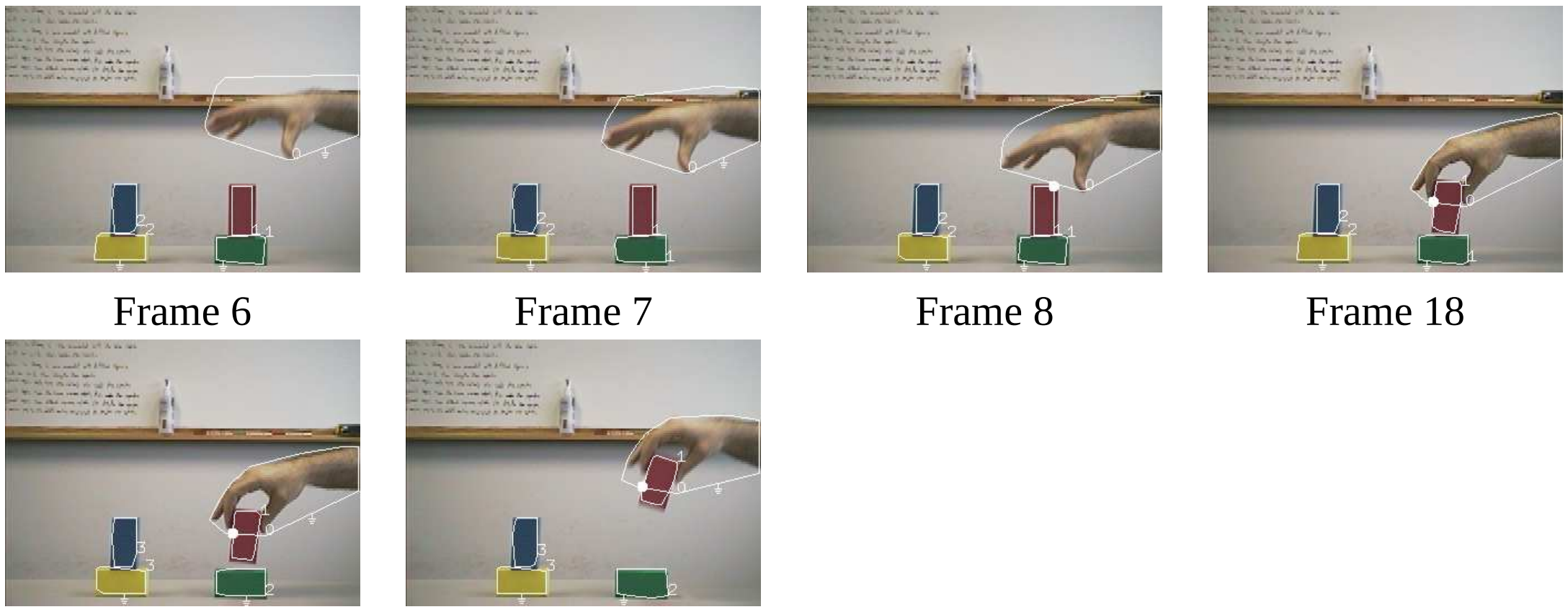

Frame 6 Frame 7 Frame 8 Frame 18

Frame 19 Frame 24

Figure 27: The output of the segmentation-and-tracking and model-reconstruction components on an image sequence depicting a *pick up* event with extraneous objects in the field of view.

profile to *pick up* and *put down* events. LEONARD correctly recognizes that these movies do not match any known event types.

An approach to even classification is valid and useful only if it is robust. A preliminary evaluation of the robustness of LEONARD was conducted. Thirty five movies were filmed, five instances of each of the seven event types *pick up*, *put down*, *stack*, *unstack*, *move*, *assemble*, and *disassemble*. These movies resemble those in Figures 1 and 11 through 15. The same subject performed all thirty five events. These movies were processed by LEONARD.[8] The results of this preliminary evaluation are summarized in Table 1. A more extensive evaluation of LEONARD will be conducted in the future.

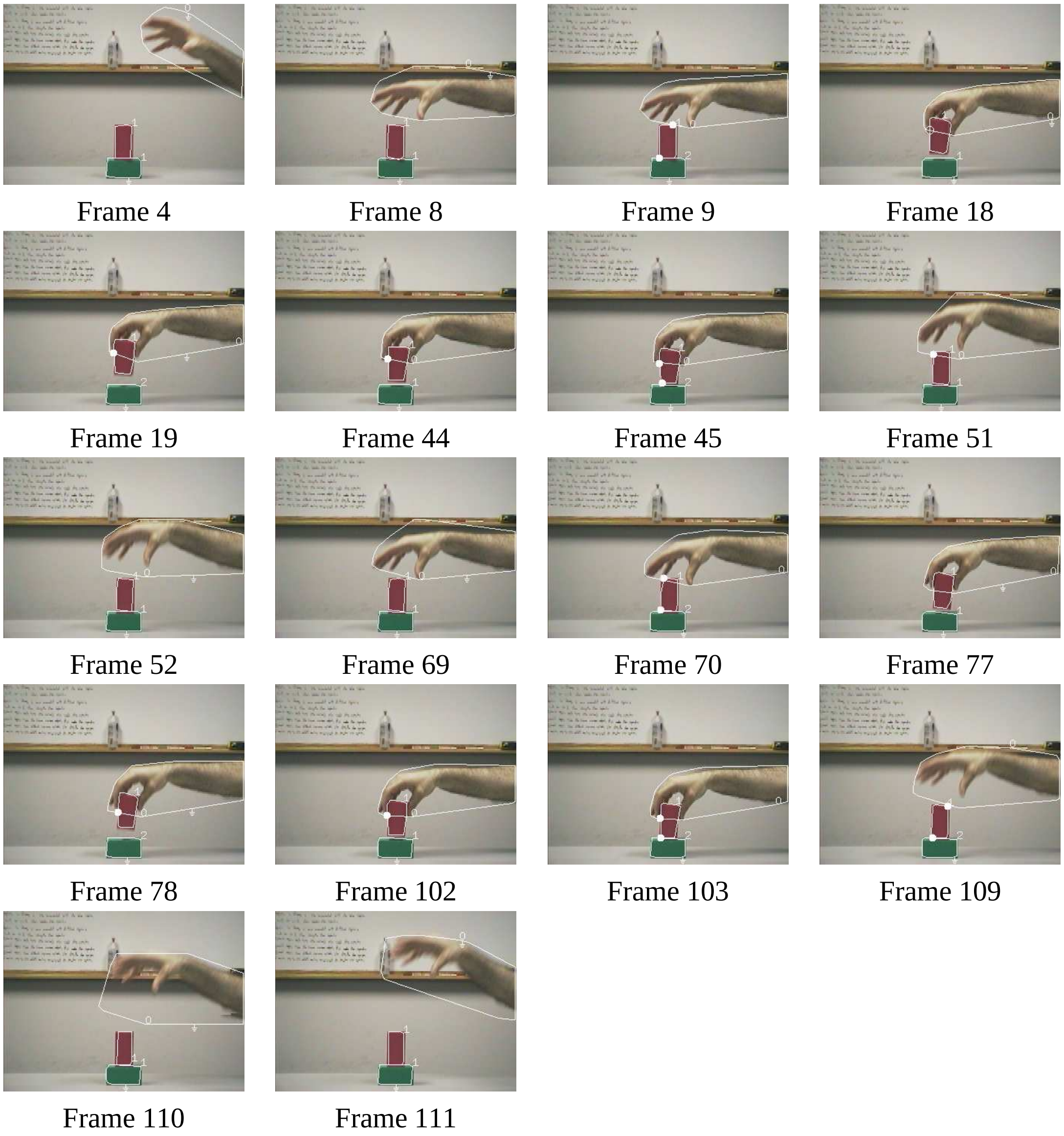

Figure 28: The output of the segmentation-and-tracking and model-reconstruction components on an image sequence depicting a sequence of a *pick up* event, followed by a *put down* event, followed by another *pick up* event, followed by another *put down* event.

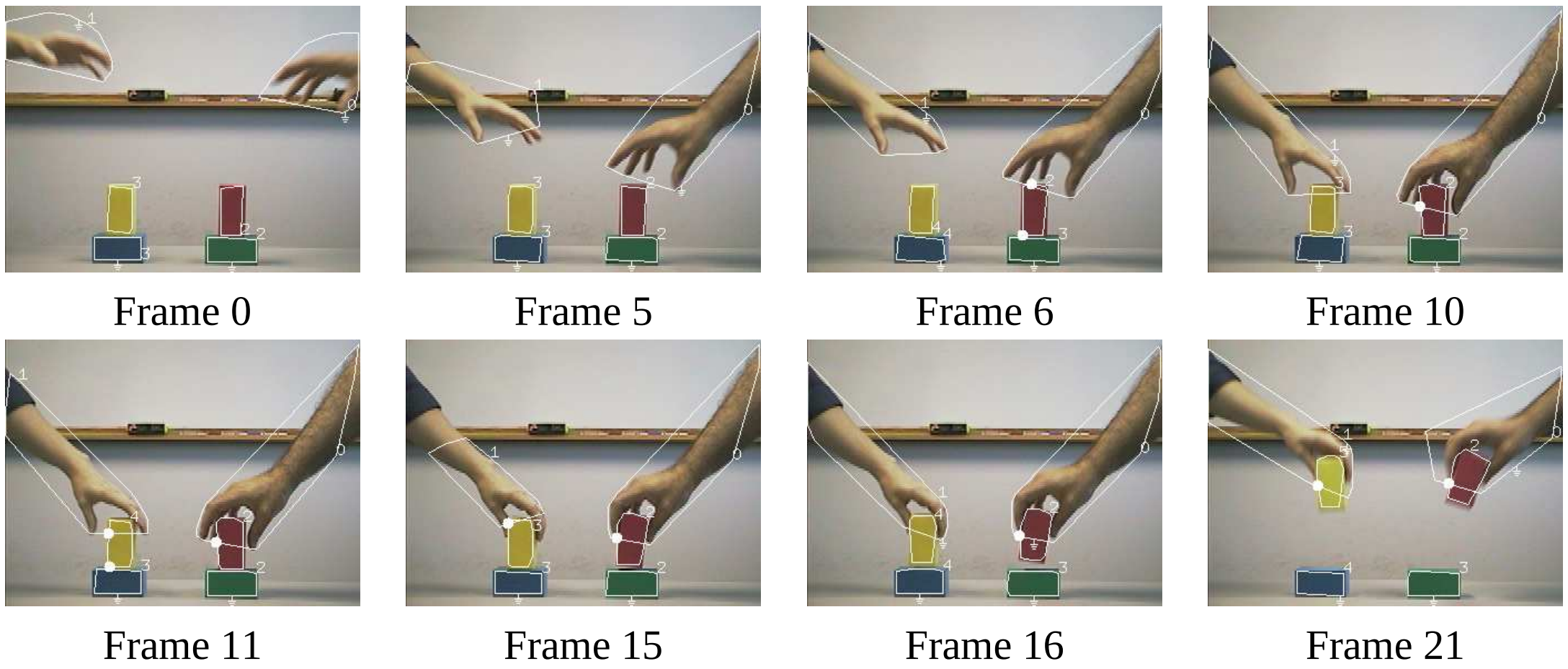

Figure 29: The output of the segmentation-and-tracking and model-reconstruction components on an image sequence depicting two simultaneous *pick up* events.

## 6. Discussion

This paper presents a new approach to event recognition that differs from the prior approach in two ways. First, it uses force dynamics instead of motion profile as the feature set to differentiate between event types. Second, it uses event logic instead of hidden Markov models as the computational framework for classifying time-series data containing these features. Nominally, these two differences are independent. One can imagine using hidden Markov models to classify time series of force-dynamic features or using event logic to classify time series of motion-profile features. While such combinations are feasible in principle, they are unwieldy in practice.

Consider using event logic to classify time series of motion-profile features. Motion-profile features, such as position, velocity, and acceleration, are typically continuous. A given event usually corresponds to a vague range of possible feature values. This vagueness is well modeled by continuous-output hidden Markov models. Event logic, which is discrete in nature, requires quantizing precise feature-value ranges. Such quantization can lead to a high misclassification rate. Furthermore, continuous distributions allow partitioning a multidimensional feature space into different classes where the boundaries between classes are more complex than lines along the feature axes. Emulating this in event logic would require complex disjunctive expressions.

Similarly, consider using hidden Markov models to classify time series of force-dynamic features. Suppose that a feature vector contains $n$ features. Since both force-dynamic and motion-profile features typically relate pairs of objects, $n$ is often quadratic in the number of event par-

8. The movies, as well as the results produced by LEONARD when processing the movies, are available as Online Appendix 1, as well as from `ftp://ftp.nj.nec.com/pub/qobi/leonard.tar.Z`.

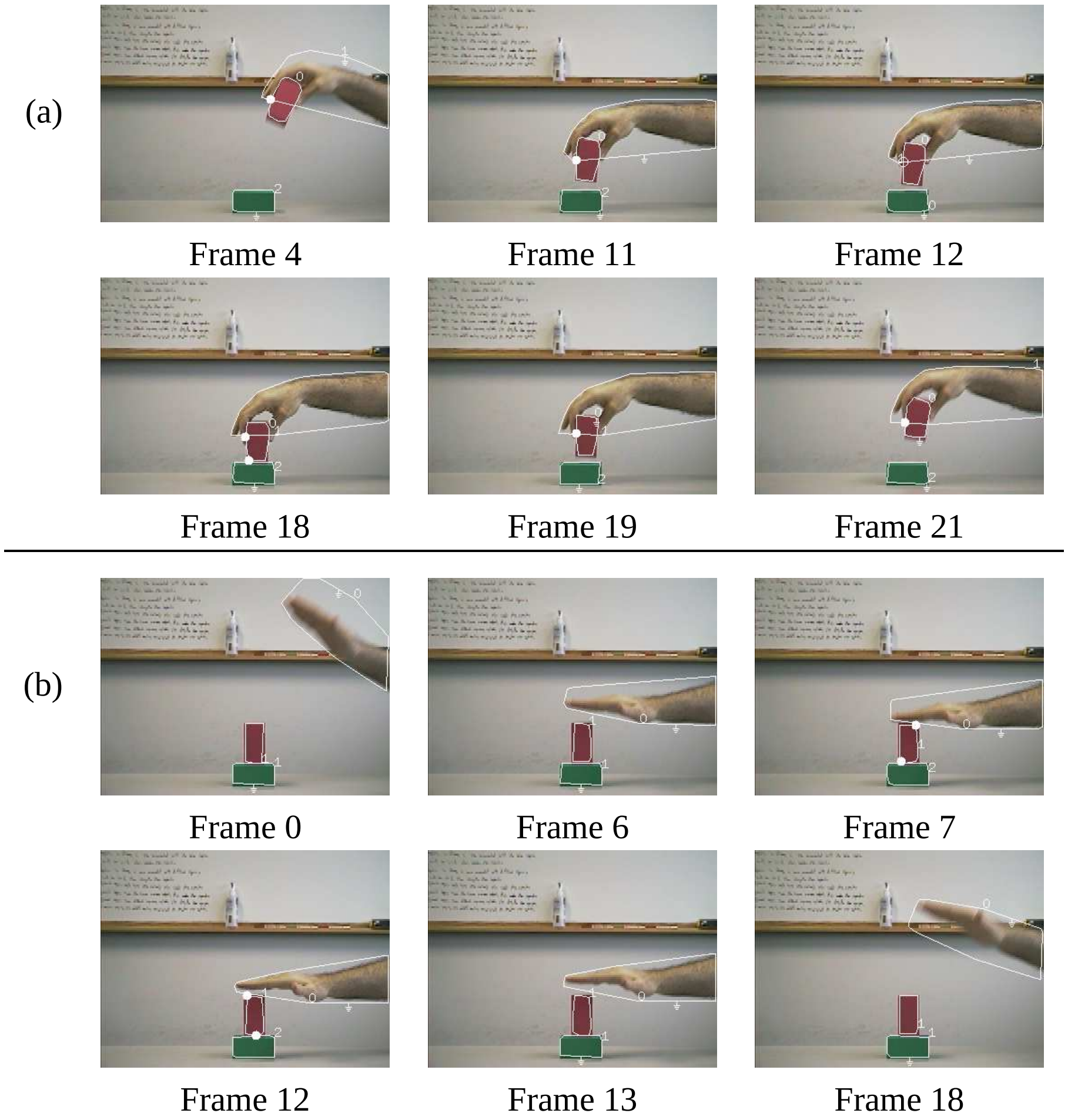

Figure 30: The output of the segmentation-and-tracking and model-reconstruction components applied to the image sequences from Figure 2, image sequences that depict non-events.

```
(PICK-UP MOVING RED GREEN)@{[[0,11],[18,30])}

(SUPPORTED? RED)@{[[0:30])}
(SUPPORTED? GREEN)@{[[11:18])}
(SUPPORTS? MOVING RED)@{[[11:30])}
(SUPPORTS? MOVING GREEN)@{[[11:18])}
(SUPPORTS? RED GREEN)@{[[11:18])}
(SUPPORTS? GREEN RED)@{[[0:11])}
(CONTACTS? RED GREEN)@{[[0:11])}
(ATTACHED? MOVING RED)@{[[11:30])}
(ATTACHED? RED GREEN)@{[[11:18])}
```

Figure 31: The output of the event-classification component applied to the model sequence from Figure 26. Note that the *pick up* event is correctly recognized despite the fact that it was performed from the left instead of from the right.

```
(PICK-UP MOVING RED GREEN)@{[[0,8],[19,30])}

(SUPPORTED? MOVING)@{[[8:19])}
(SUPPORTED? RED)@{[[0:30])}
(SUPPORTED? BLUE)@{[[0:30])}
(SUPPORTS? MOVING RED)@{[[19:30])}
(SUPPORTS? RED MOVING)@{[[8:19])}
(SUPPORTS? GREEN MOVING)@{[[8:19])}
(SUPPORTS? GREEN RED)@{[[0:19])}
(SUPPORTS? YELLOW BLUE)@{[[0:30])}
(CONTACTS? RED GREEN)@{[[0:10]), [[16:19])}
(CONTACTS? BLUE YELLOW)@{[[0:30])}
(ATTACHED? MOVING RED)@{[[8:30])}
(ATTACHED? RED GREEN)@{[[10:16])}
```

Figure 32: The output of the event-classification component applied to the model sequence from Figure 27. Note that the *pick up* event is correctly recognized despite the presence of extraneous objects in the field of view.

```
(PICK-UP MOVING RED GREEN)@{[[52,70],[78,102]),
                            [[0,9],[19,44])}
(PUT-DOWN MOVING RED GREEN)@{[[19,44],[52,70]),
                             [[78,102],[110,117])}

(SUPPORTED? MOVING)@{[[9:18]), [[44:52]),
                     [[70:77]), [[102:110])}
(SUPPORTED? RED)@{[[0:117])}
(SUPPORTS? MOVING RED)@{[[18:44]), [[78:102])}
(SUPPORTS? RED MOVING)@{[[9:18]), [[44:52]),
                        [[70:77]), [[102:110])}
(SUPPORTS? GREEN MOVING)@{[[9:18]), [[44:52]),
                          [[70:77]), [[102:110])}
(SUPPORTS? GREEN RED)@{[[0:19]), [[44:78]), [[102:117])}
(CONTACTS? RED GREEN)@{[[0:9]), [[13:18]), [[46:70]),
                       [[106:117])}
(ATTACHED? MOVING RED)@{[[9:52]), [[70:110])}
(ATTACHED? RED GREEN)@{[[9:13]), [[70:76]), [[104:106])}
```

Figure 33: The output of the event-classification component applied to the model sequence from Figure 28. Note that LEONARD correctly recognizes a *pick up* event, followed by a *put down* event, followed by another *pick up* event, followed by another *put down* event.

```
(PICK-UP MOVING RED GREEN)@{[[0,6],[16,22])}
(PICK-UP MOVING YELLOW BLUE)@{[[0,12],[17,22])}

(SUPPORTED? MOVING)@{[[6:16])}
(SUPPORTED? MOVING)@{[[12:15])}
(SUPPORTED? RED)@{[[0:22])}
(SUPPORTED? YELLOW)@{[[0:22])}
(SUPPORTS? MOVING RED)@{[[16:22])}
(SUPPORTS? MOVING YELLOW)@{[[17:22])}
(SUPPORTS? RED MOVING)@{[[6:16])}
(SUPPORTS? GREEN MOVING)@{[[6:16])}
(SUPPORTS? GREEN RED)@{[[0:16])}
(SUPPORTS? BLUE MOVING)@{[[12:15])}
(SUPPORTS? BLUE YELLOW)@{[[0:17])}
(SUPPORTS? YELLOW MOVING)@{[[12:15])}
(CONTACTS? RED GREEN)@{[[0:15])}
(CONTACTS? BLUE YELLOW)@{[[0:17])}
(ATTACHED? MOVING RED)@{[[6:22])}
(ATTACHED? MOVING YELLOW)@{[[12:22])}
```

Figure 34: The output of the event-classification component applied to the model sequence from Figure 29. Note that the two simultaneous *pick up* events are correctly recognized.

```
(SUPPORTED? RED)@{[[0:19])}
(SUPPORTED? MOVING)@{[[13:31])}
(SUPPORTS? RED MOVING)@{[[13:31])}
(SUPPORTS? MOVING RED)@{[[0:13])}
(SUPPORTS? GREEN RED)@{[[12:19])}
(SUPPORTS? GREEN MOVING)@{[[13:19])}
(ATTACHED? RED MOVING)@{[[0:31])}
(ATTACHED? RED GREEN)@{[[13:19])}
```

(a)

```
(SUPPORTED? RED)@{[[0:25])}
(SUPPORTED? GREEN)@{[[7:13])}
(SUPPORTS? MOVING RED)@{[[7:13])}
(SUPPORTS? MOVING GREEN)@{[[7:13])}
(SUPPORTS? RED GREEN)@{[[7:13])}
(SUPPORTS? GREEN RED)@{[[0:7]), [[13:25])}
(CONTACTS? RED GREEN)@{[[0:7]), [[13:25])}
(ATTACHED? MOVING RED)@{[[7:13])}
(ATTACHED? RED GREEN)@{[[7:13])}
```

(b)

Figure 35: The output of the event-classification component applied to the model sequences from Figure 30. Note that LEONARD correctly recognizes that no events occurred in these sequences.

| | *pick up* | *put down* | *stack* | *unstack* | *move* | *assemble* | *disassemble* |
|---|---|---|---|---|---|---|---|
| *pick up* | 5/5 | | | | | | |
| *put down* | | 5/5 | | | | | |
| *stack* | | 5/5 | 5/5 | | | | |
| *unstack* | 5/5 | | | 4/5 | | | |
| *move* | 4/5 | 5/5 | | | 4/5 | | |
| *assemble* | | 5/10 | 1/5 | | | 1/5 | |
| *disassemble* | 10/10 | | | 5/5 | | | 5/5 |

Table 1: An evaluation of the robustness of LEONARD on a test set of five movies of each of seven event types. The rows represent movies of the indicated event types. The columns represent classifications of the indicated event type. The entries $x/y$ indicate $x$, the number of times that a movie of the indicated event type was classified as the indicated event type, and $y$, the number of times that the movie should have been classified as the indicated event type. Note that *stack* entails *put down*, *unstack* entails *pick up*, *move* entails both a *pick up* and a *put down*, *assemble* entails both a *put down* and a separate *stack*, and *disassemble* entails both a *pick up* and a separate *unstack*. Thus off-diagonal entries are expected in these cases. There were six false negatives and no false positives. Four of the false negatives were for the event type *assemble*. In three of those cases, LEONARD successfully recognized the constituent *put down* subevent but failed to recognize the constituent *stack* subevent as well as the associated *put down* subevent. In one case, LEONARD failed to recognize both the constituent *put down* and *stack* subevents along with the associated *put down* constituent of the *stack* subevent. One of the false negatives was for the event type *move*. In this case, LEONARD successfully recognized the constituent *put down* subevent but failed to recognize the constituent *pick up* subevent. The remaining false negative was for the event type *unstack*. In this case, LEONARD successfully recognized the constituent *pick up* subevent but failed to recognize the aggregate *unstack* event.

ticipants. Let us contrast the number of parameters needed to represent these features in both the motion-profile and force-dynamic approaches. Since, as discussed above, motion-profile features are typically continuous, hidden Markov models with continuous outputs can be used in the motion-profile approach. When the features are independent, such a model requires $O(n)$ parameters per state to specify the output distributions. Even if one uses, say, a multivariate Gaussian to model dependent features, this requires only $O(n^2)$ parameters per state to specify the output distributions in the motion-profile approach. However, force-dynamic features are Boolean. This requires using discrete-output hidden Markov models. Such models output a stream of symbols, not feature vectors. Constructing an appropriate alphabet of output symbols requires considering all possible subsets of features. This requires $O(2^n)$ parameters per state to specify the output distributions in the force-dynamic approach. Thus continuous-output hidden Markov models appear to be better suited to an approach that uses motion-profile features while event logic appears to be better suited to an approach that uses force-dynamic features.

Humans use language for three fundamental purposes: we describe what we see, we ask others to perform actions, and we engage in conversation. The first two require grounding language in perception and action. Only the third involves disembodied use of language. Almost all research in computational linguistics has focused on such disembodied language use. Data-base query processing, information extraction and retrieval, and spoken-language dialog all use language solely to manipulate internal representations. In contrast, the work described in this paper grounds language in perception of the external world. It describes an implemented system, called LEONARD, that uses language to describe events observed in short image sequences.

Why is perceptual grounding of language important and relevant to computational linguistics? Current approaches to lexical semantics suffer from the ‘bold-face syndrome.’ All too often, the meanings of words, like *throw*, are taken to be uninterpreted symbols, like **throw**, or expressions over uninterpreted symbols, like **cause to go** (Leech, 1969; Miller, 1972; Schank, 1973; Jackendoff, 1983, 1990; Pinker, 1989). Since the interpretation of such symbols is left to informal intuition, the correctness of any meaning representation constructed from such uninterpreted symbols cannot be verified. In other words, how is one to know whether **cause to go** is the correct meaning of *throw*? Perceptual grounding offers a way to verify semantic representations. Having an implemented system use a collection of semantic representations to generate appropriate descriptions of observations gives evidence that those semantic representations are correct. This paper takes a small step in this direction. In contrast to prior work, which presents informal semantic representations whose interpretation is left to intuition, it presents perceptually-grounded semantic representations. While the system described in this paper addresses only perceptual grounding of language, the long-term goal of this research is to provide a unified semantic representation that is sufficiently powerful to support all three forms of language use: perception, action, and conversation.

Different parts of speech in language typically describe different aspects of visual percepts. Nouns typically describe objects. Verbs typically describe events. Adjectives typically describe properties. Prepositions typically describe spatial and temporal relations. Grounding language in visual perception will require construction of semantic representations for all of these different parts of speech. It is likely that different parts of speech will require different machinery to represent their lexical semantics. In other words, whatever the ultimate representation of *apple* and *chair* are, they are likely to be based on very different principles than the ultimate representation of *pick up* and *put down*. These, in turn, are likely to be further different from those needed to represent *in*, *on*, *red*, and *big*. Indeed, machine vision research, at least that aspect of machine vision research that focuses on object recognition, can be viewed as an attempt to perceptually ground the lexical semantics of nouns. In contrast, this paper focuses solely on verbs. Accordingly, it develops machinery that is very different from what is typically used in the machine-vision community, machinery that is more reminiscent of that which is used in the knowledge-representation community. On the other hand, unlike typical knowledge-representation work, it grounds that machinery in image processing.

When one proposes a representation, such as **cause to go**, as the meaning of a word, such as *throw*, one must specify three things to effectively specify the meaning of that word. First, one must specify the lexical semantics of the individual primitives, how one determines the truth conditions of items like **cause** and **to go**. Second, one must specify the compositional semantics of the representation, how one combines the truth conditions of primitives like **cause** and **to go** to get the aggregate truth conditions of compound expressions like **cause to go**. Third, one must specify a lexical entry, a map from a word, like *throw*, to a compound expression, like **cause to go**. All three are necessary in order to precisely specify the word meaning.

Prior work in lexical semantics, such as the work of Leech (1969), Miller (1972), Schank (1973), Jackendoff (1983, 1990), and Pinker (1989), is deficient in this regard. It specifies the third component without the first two. In other words, it formulates lexical entries in terms of compound expressions like **cause to go**, without specifying the meanings of the primitives, like **cause** and **to go**, and without specifying how these meanings are combined to form the aggregate meaning of the compound expression. This paper attempts to address that deficiency by specifying all three components. First, the lexical semantics of the event-logic primitives is precisely specified in Figure 9. Second, the compositional semantics of event logic is precisely specified in Section 3. Third, lexical entries for several verbs are precisely specified in Figure 10. These three components together formally specify the meanings of those verbs with a level of precision that is absent in prior work.

While these lexical entries are *precise*, there is no claim that they are *accurate*. Lexical entries are *precise* when their meaning is reduced to an impartial mechanical procedure. Lexical entries are *accurate* when they properly reflect the truth conditions for the words that they define. Even ignoring homonymy and metaphor, words such as *move* and *assemble* clearly have meanings that are much more complex than what is, and even can be, represented with the machinery presented in this paper. But that holds true of prior work as well. The lexical entries given in, for example, Leech (1969), Miller (1972), Schank (1973), Jackendoff (1983, 1990), and Pinker (1989) also do not accurately reflect the truth conditions for the words that they define. The purpose of this paper is not to improve the accuracy of definitions. In fact, the definitions given in prior work might be more accurate, in some ways, than those given here. Rather, its purpose is to improve the precision of definitions. The definitions given in prior work are imprecise and that imprecision makes assessing their accuracy a subjective process: do humans think an informally specified representation matches their intuition. In contrast, precision allows objective assessment of accuracy: does the output of a mechanical procedure applied to sample event occurrences match human judgments of which words characterize those occurrences.

Precision is the key methodological advance of this work. Precise specification of the meaning of lexical semantic representations, by way of perceptual grounding, makes accuracy assessment possible by way of experimental evaluation. Taking this first step of advancing precision and perceptual grounding will hopefully allow us to take future steps towards improving accuracy.

## 7. Related Work

Most prior work uses motion profile, some combination of relative-and absolute linear-and-angular positions, velocities, and accelerations, as the features that drive event classification. That work follows the tradition of linguists and cognitive scientists, such as Leech (1969), Miller (1972), Schank (1973), Jackendoff (1983, 1990), and Pinker (1989), that represent the lexical semantics of verbs via the causal, aspectual, and directional qualities of motion. Some linguists and cognitive scientists, such as Herskovits (1986) and Jackendoff and Landau (1991), have argued that force-dynamic relations (Talmy, 1988), such as support, contact, and attachment, are crucial for representing the lexical semantics of spatial prepositions. For example, in some situations, part of what it means for one object to be *on* another object is for the former to be in contact with, and supported by, the latter. In other situations, something can be on something else by way of attachment, as in *the knob on the door*. Siskind (1992) has argued that change in the state of force-dynamic relations plays a more central role in specifying the lexical semantics of simple spatial motion verbs than motion profile. The particular relative-and-absolute linear-and-angular positions, velocities, and accelerations don't matter when picking something up or putting something down. What matters is a state change in the source of support of the patient. Similarly, what distinguishes putting something down from dropping it is that, in the former, the patient is always supported, while in the latter, the patient undergoes unsupported motion.

The work described in this paper differs from prior work in visual-event perception in a number of respects. Waltz and Boggess (1979), Waltz (1981), Marr and Vaina (1982), and Rubin and Richards (1985) describe unimplemented frameworks that are not based on force dynamics. Thibadeau (1986) describes a system that recognizes when an event occurs but not what event occurs. His system processes simulated video and is not based on force dynamics. Badler (1975), Adler (1977), Tsuji, Morizono, and Kuroda (1977), Okada (1979), Tsuji, Osada, and Yachida (1979, 1980), Abe, Soga, and Tsuji (1981), Abe and Tsuji (1982), Novak and Bulko (1990), and Regier (1992) describe systems that process simulated video and that are not based on force dynamics. Borchardt (1984, 1985) presents event definitions that are based on force-dynamic relations but does not present techniques for recovering those relations automatically from either simulated or real video. Yamoto et al. (1992), Starner (1995), Siskind and Morris (1996), Siskind (1996), Brand (1996, 1997a), Brand, Oliver, and Pentland (1997), and Bobick and Ivanov (1998) present systems that recognize event occurrences from real video using motion profile but not force dynamics. These systems use hidden Markov models rather than event logic as the event-classification engine. Funt (1980) presents a heuristic approach to stability analysis that operates on simulated video but does not perform model reconstruction or event classification. Brand, Birnbaum, and Cooper (1993) and Brand (1997b) present a heuristic approach to stability analysis that operates on real video but do not use stability analysis to perform model reconstruction and event classification. Blum, Griffith, and Neumann (1970) and Fahlman (1974) present stability-analysis algorithms that are based on linear programming but do not use stability analysis to perform model reconstruction or event classification. These stability-analysis algorithms use dynamics rather than kinematics. Siskind (1991, 1992, 1993, 1994, 1995, 1997) presents systems that operate on simulated video and use force dynamics to recognize event occurrences. All of that work, except Siskind (1997), uses heuristic approaches to stability analysis, model reconstruction, and event classification. Siskind (1997) presents an early version of the stability-analysis and event-logic–based event-recognition techniques used in the current system. Mann et al. (1996, 1997) and Mann and Jepson (1998) present a system that does model reconstruction from real video but does not use the recovered force-dynamic relations to perform event classification. That work uses an approach to stability analysis based on dynamics instead of the kinematic approach used in this paper.

There is also a body of prior work that grounds fragments of natural-language semantics in physical relations between objects in graphically represented blocks worlds or for solving physics word problems. Examples of such work include Bobrow (1964), Winograd (1972), and Palmer (1990) as well the ISSAC system (Novak, 1976) and the MECHO project (Bundy, Luger, Palmer, & Welham, 1998; Bundy, Byrd, Luger, Mellish, Milne, & Palmer, 1979; Luger, 1981). While that work does not focus on recognizing events, *per se*, it does relate lexical semantics to physical relations between represented objects.

LEONARD currently does not contain a learning component. It is given a fixed physical theory of the world, implicitly represented in the model-reconstruction procedure, and a fixed collection of event-type descriptions, explicitly formulated as event-logic expressions. One potential area for future work would be to automatically learn a physical theory of the world and/or event-type descriptions. Adding a learning component could potentially produce more robust model-reconstruction and event-classification components than those currently constructed by hand. Techniques such as those presented in Martin and Geffner (2000) and Cumby and Roth (2000) might be useful for this task.

## 8. Conclusion

This paper has presented LEONARD, a comprehensive implemented system for recovering event occurrences from video input. It differs from the prior approach to the same problem in two fundamental ways. First, it uses state changes in the force-dynamic relations between objects, instead of motion profile, as the key descriptive element in defining event types. Second, it uses event logic, instead of hidden Markov models, to perform event classification. One key result of this paper is the formulation of spanning intervals, a novel efficient representation of the infinite sets of intervals that arise when processing liquid and semi-liquid events. A second key result of this paper is the formulation of an efficient procedure, based on spanning intervals, for inferring all occurrences of compound event types from occurrences of primitive event types. The techniques of force-dynamic model reconstruction, spanning intervals, and event-logic inference have been used to successfully recognize seven event types from real video: *pick up*, *put down*, *stack*, *unstack*, *move*, *assemble*, and *disassemble*. Using force dynamics and event logic to perform event recognition offers four key advantages over the prior approach of using motion profile and hidden Markov models. First, it is insensitive to variance in the motion profile of an event occurrence. Second, it is insensitive to the presence of extraneous objects in the field of view. Third, it allows temporal segmentation of sequential and parallel event occurrences. Fourth, it robustly detects the non-occurrence of events as well as their occurrence.

At a more fundamental level, this paper advances a novel methodology: grounding lexical-semantic representations in visual-event perception as a means for assessing the accuracy of such representations. Prior work in lexical-semantic representations has used calculi whose semantics were not precisely specified. Lexical entries formulated in such calculi derived their meaning from intuition and thus could not be empirically tested. By providing a lexical-semantic representation whose semantics is precisely specified via perceptual grounding, this paper opens up the field of lexical semantics to empirical evaluation. The particular representations advanced in this paper are clearly only approximations to the ultimate truth. This follows from the primitive state of our understanding of language and perception. Nonetheless, I hope that this paper offers an advance towards the ultimate truth, both through its novel methodology and the particular details of its mechanisms.